\documentclass{article}

\usepackage{tabularx}
\usepackage{graphicx}
\usepackage{fullpage}
\usepackage{hyperref}
\usepackage{natbib}
\usepackage{delarray}
\usepackage{times}
\usepackage{subfigure}
\usepackage{color}
\usepackage{amssymb}
\usepackage{amsmath}
\usepackage{mathrsfs}
\usepackage{tikz}
\usepackage{xspace}
\usepackage{enumerate}
\usepackage{algorithm}
\usepackage{algorithmic}
\usepackage{multirow}
\usepackage{nameref}
\usepackage{booktabs}
\usepackage{mdframed}
\usepackage{fmtcount}
\usepackage{footnote}
\usepackage{makecell}
\usepackage{colortbl}
\usepackage{thmtools}
\usepackage{thm-restate}
\makesavenoteenv{table}
\makesavenoteenv{tabular}
\makesavenoteenv{algorithm}
\makesavenoteenv{algorithmic}
\makesavenoteenv{figure}

\newcommand{\eat}[1]{}

\newcommand{\head}[1]{\noindent{{\bf #1:}}}

\newcommand{\pr}{{\mathbb P}}
\newcommand{\R}{{\mathbb R}}

\newcommand{\E}{{\mathbb{E}}}

\newcommand{\spe}{{super\_epoch}}

\newcommand{\red}[1]{\textcolor{red}{#1}}

\newcommand{\inner}[2]{\langle #1, #2 \rangle}

\newcommand{\ns}[1]{\| #1 \|^2}

\newcommand{\n}[1]{\| #1 \|}
\newcommand{\hx}{\hat{x}}

\newcommand{\tx}{\widetilde{x}}
\newcommand{\tdo}{\widetilde{O}}

\newcommand{\bx}{\bar{x}}
\newcommand{\hess}{\mathcal{H}}

\newcommand{\mathG}{g_{\mathrm{thres}}}
\newcommand{\mathf}{f_{\mathrm{thres}}}
\newcommand{\mathT}{t_{\mathrm{thres}}}

\definecolor{bgcolor}{rgb}{0.66,0.88,1.00}

\newcommand{\base}{{1+\eta\gamma}}

\newcommand{\Dtop}{{\frac{\delta}{C_1\rho}+r}}

                {\hspace*{\fill}$\Box$\par}
        {\hspace*{\fill}$\Box$\par\vspace{4mm}}
\newenvironment{proofof}[1]{\smallskip\noindent{\bf Proof of #1.}}%
        {\hspace*{\fill}$\Box$\par}

\newtheorem{theorem}{Theorem}
\newtheorem{lemma}{Lemma}

\newtheorem{definition}{Definition}
\newtheorem{assumption}{Assumption}
\newtheorem{proposition}{Proposition}

\hypersetup{
    colorlinks=true,       
    linkcolor=blue,    
    citecolor=blue,        
    filecolor=magenta,     
    urlcolor=blue
}

\begin{document}

\title{SSRGD: Simple Stochastic Recursive Gradient Descent for Escaping Saddle Points}
\author{Zhize Li \\
     IIIS, Tsinghua University \\
    zz-li14@mails.tsinghua.edu.cn}

\date{}
\maketitle

\begin{abstract}
We analyze stochastic gradient algorithms for optimizing nonconvex problems.
In particular, our goal is to find local minima (second-order stationary points) instead of just finding first-order stationary points which may be some bad unstable saddle points.
We show that a simple perturbed version of stochastic recursive gradient descent algorithm (called SSRGD) can find an $(\epsilon,\delta)$-second-order stationary point with $\widetilde{O}(\sqrt{n}/\epsilon^2 + \sqrt{n}/\delta^4 + n/\delta^3)$ stochastic gradient complexity for nonconvex finite-sum problems.
As a by-product, SSRGD finds an $\epsilon$-first-order stationary point with $O(n+\sqrt{n}/\epsilon^2)$ stochastic gradients. These results are almost optimal since \cite{fang2018spider} provided a lower bound $\Omega(\sqrt{n}/\epsilon^2)$ for finding even just an $\epsilon$-first-order stationary point.
We emphasize that SSRGD algorithm for finding second-order stationary points is as simple as for
finding first-order stationary points just by adding a uniform perturbation sometimes, while all other algorithms for finding second-order stationary points with similar gradient complexity need to combine with a negative-curvature search subroutine (e.g., Neon2 \citep{allen2018neon2}).
Moreover, the simple SSRGD algorithm gets a simpler analysis.
Besides, we also extend our results from nonconvex finite-sum problems to nonconvex online (expectation) problems, and prove the corresponding convergence results.
\end{abstract}

\vspace{-1.5mm}
\section{Introduction}\vspace{-1mm}
\label{sec:intro}
Nonconvex optimization is ubiquitous in machine learning applications especially for deep neural networks.
For convex optimization, every local minimum is a global minimum and it can be achieved by any first-order stationary point, i.e., $\nabla f(x)=0$.
However, for nonconvex problems, the point with zero gradient can be a local minimum, a local maximum or a saddle point.
To avoid converging to bad saddle points (including local maxima), we want to find a second-order stationary point, i.e., $\nabla f(x)=0$ and $\nabla^2 f(x) \succeq 0$ (this is a necessary condition for $x$ to be a local minimum). All second-order stationary points indeed are local minima if function $f$ satisfies strict saddle property \citep{ge2015escaping}.
Note that finding the global minimum in nonconvex problems is NP-hard in general.
Also note that it was shown that all local minima are also global minima for some nonconvex problems, e.g., matrix sensing \citep{bhojanapalli2016global}, matrix completion \citep{ge2016matrix}, and some neural networks \citep{ge2017learning}.
Thus, our goal in this paper is to find an approximate second-order stationary point (local minimum) with proved convergence.

There has been extensive research for finding $\epsilon$-first-order stationary point (i.e., $\|\nabla f(x)\|\leq \epsilon$), e.g., GD, SGD and SVRG. See Table \ref{table:finite} for an overview.
Although \cite{xu2018first} and \cite{allen2018neon2} independently proposed reduction algorithms Neon/Neon2 that can be combined with previous $\epsilon$-first-order stationary points finding algorithms to find an $(\epsilon,\delta)$-second-order stationary point (i.e., $\|\nabla f(x)\|\leq \epsilon$ and $\lambda_{\min}(\nabla^2 f(x))\geq -\delta$).
However, algorithms obtained by this reduction are very complicated in practice, and they need to extract negative curvature directions from the Hessian to escape saddle points by using a negative curvature search subroutine: given a point $x$, find an approximate smallest eigenvector of $\nabla^2 f(x)$. This also involves a more complicated analysis.
Note that in practice, standard first-order stationary point finding algorithms can often work (escape bad saddle points) in nonconvex setting without a negative curvature search subroutine.
The reason may be that the saddle points are usually not very stable.
So there is a natural question ``Is there any simple modification to allow first-order stationary point finding algorithms to get a theoretical second-order guarantee?".
For gradient descent (GD), \cite{jin2017escape} showed that a simple perturbation step is enough to escape saddle points for finding a second-order stationary point, and this is necessary \citep{du2017gradient}.
Very recently, \cite{ge2019stable} showed that a simple perturbation step is also enough to find a second-order stationary point for SVRG algorithm \citep{li2018simple}. Moreover, \cite{ge2019stable} also developed a stabilized trick to further improve the dependency of Hessian Lipschitz parameter.

\vspace{-4mm}
\begin{table}[h]
\centering
\caption{Stochastic gradient complexity of optimization algorithms for nonconvex finite-sum problem \eqref{eq:finite}}\label{table:finite}
\begin{tabular}{|c|c|c|c|}
\hline
Algorithm & \makecell{Stochastic gradient\\ complexity} & Guarantee & \makecell{Negative-curvature\\ search subroutine} \\
\hline
GD \citep{nesterov2014introductory} & $O(\frac{n}{\epsilon^2})$ & 1st-order & No \\ \hline
\makecell{SVRG \citep{reddi2016stochastic}, \\
\citep{allen2016variance};\\
  SCSG \citep{lei2017non};\\ SVRG+ \citep{li2018simple}  }  & $O(n+\frac{n^{2/3}}{\epsilon^2})$ & 1st-order & No \\ \hline
\makecell{SNVRG \citep{zhou2018stochastic};\\
SPIDER \citep{fang2018spider};\\
SpiderBoost \citep{wang2018spiderboost};\\
SARAH \citep{pham2019proxsarah}} & $O(n+\frac{n^{1/2}}{\epsilon^2})$  &1st-order & No \\ \hline
\rowcolor{bgcolor}
SSRGD (this paper) & $O(n+\frac{n^{1/2}}{\epsilon^2})$  &1st-order & No \\ \hline
PGD \citep{jin2017escape} & $\tdo(\frac{n}{\epsilon^2} + \frac{n}{\delta^4})$ & 2nd-order & No \\ \hline
\makecell{Neon2+FastCubic/CDHS\\
\citep{agarwal2016finding,carmon2016accelerated}} & $\tdo(\frac{n}{\epsilon^{1.5}}+\frac{n}{\delta^{3}}
+\frac{n^{3/4}}{\epsilon^{1.75}}+\frac{n^{3/4}}{\delta^{3.5}})$ & 2nd-order & Needed \\ \hline Neon2+SVRG \citep{allen2018neon2} & $\tdo(\frac{n^{2/3}}{\epsilon^2} +\frac{n}{\delta^{3}}+\frac{n^{3/4}}{\delta^{3.5}})$ & 2nd-order & Needed \\ \hline
Stabilized SVRG \citep{ge2019stable} & $\tdo(\frac{n^{2/3}}{\epsilon^2} +\frac{n}{\delta^3}+\frac{n^{2/3}}{\delta^4})$ & 2nd-order & No \\ \hline
SNVRG$^+$+Neon2 \citep{zhou2018finding} &$\tdo(\frac{n^{1/2}}{\epsilon^2} +\frac{n}{\delta^3}+\frac{n^{3/4}}{\delta^{3.5}})$ & 2nd-order & Needed\\ \hline
SPIDER-SFO$^+$(+Neon2) \citep{fang2018spider} &$\tdo(\frac{n^{1/2}}{\epsilon^2} +\frac{n^{1/2}}{\epsilon\delta^2}+\frac{1}{\epsilon\delta^3}+\frac{1}{\delta^5})$ & 2nd-order & Needed \\ \hline
\rowcolor{bgcolor}
SSRGD (this paper)& $\tdo(\frac{n^{1/2}}{\epsilon^2} +\frac{n^{1/2}}{\delta^4} + \frac{n}{\delta^3})$ & 2nd-order & No \\ \hline
\end{tabular}
\end{table}

\vspace{-8mm}
\begin{table}[!h]
\centering
\caption{Stochastic gradient complexity of optimization algorithms for nonconvex online (expectation) problem \eqref{eq:online}}\label{table:online}
\begin{tabular}{|c|c|c|c|}
\hline
Algorithm & \makecell{Stochastic gradient\\ complexity} & Guarantee & \makecell{Negative-curvature\\ search subroutine} \\
\hline
SGD \citep{ghadimi2016mini} & $O(\frac{1}{\epsilon^4})$ & 1st-order & No \\ \hline
\makecell{
  SCSG \citep{lei2017non};\\ SVRG+ \citep{li2018simple}}  & $O(\frac{1}{\epsilon^{3.5}})$ & 1st-order & No \\ \hline
\makecell{SNVRG \citep{zhou2018stochastic};\\
SPIDER \citep{fang2018spider};\\
SpiderBoost \citep{wang2018spiderboost};\\
SARAH \citep{pham2019proxsarah}} & $O(\frac{1}{\epsilon^3})$  &1st-order & No \\ \hline
\rowcolor{bgcolor}
SSRGD (this paper) & $O(\frac{1}{\epsilon^3})$  &1st-order & No \\ \hline
Perturbed SGD \citep{ge2015escaping} & poly$(d,\frac{1}{\epsilon},\frac{1}{\delta})$ & 2nd-order & No \\ \hline
CNC-SGD \citep{daneshmand2018escaping}& $\tdo(\frac{1}{\epsilon^{4}} +\frac{1}{\delta^{10}})$ & 2nd-order & No \\ \hline
Neon2+SCSG \citep{allen2018neon2} & $\tdo(\frac{1}{\epsilon^{10/3}} +\frac{1}{\epsilon^2\delta^3}+\frac{1}{\delta^5})$ & 2nd-order & Needed \\ \hline
Neon2+Natasha2 \citep{allen2018natasha} & $\tdo(\frac{1}{\epsilon^{3.25}}+\frac{1}{\epsilon^3\delta}+\frac{1}{\delta^5})$ & 2nd-order & Needed \\ \hline
SNVRG$^+$+Neon2 \citep{zhou2018finding} &$\tdo(\frac{1}{\epsilon^{3}}+\frac{1}{\epsilon^2\delta^3}+\frac{1}{\delta^5})$ & 2nd-order & Needed\\ \hline
SPIDER-SFO$^+$(+Neon2) \citep{fang2018spider} &$\tdo(\frac{1}{\epsilon^{3}}+\frac{1}{\epsilon^2\delta^2}+\frac{1}{\delta^5})$ & 2nd-order & Needed \\ \hline
\rowcolor{bgcolor}
SSRGD (this paper)& $\tdo(\frac{1}{\epsilon^{3}}+\frac{1}{\epsilon^2\delta^3}+\frac{1}{\epsilon\delta^4})$ & 2nd-order & No \\ \hline
\end{tabular}
\end{table}
\vspace{-3mm}
\noindent{{\bf Note:}}
1. Guarantee (see Definition \ref{def:sp}): $\epsilon$-first-order stationary point $\n{\nabla f(x)}\leq \epsilon$; $(\epsilon,\delta)$-second-order stationary point  $\n{\nabla f(x)}\leq \epsilon$ and $\lambda_{\min}(\nabla^2 f(x))\geq -\delta$.

2. In the classical setting where $\delta=O(\sqrt{\epsilon}) $ \citep{nesterov2006cubic,jin2017escape}, our simple SSRGD is always (no matter what $n$ and $\epsilon$ are) not worse than all other algorithms (in both Table \ref{table:finite} and \ref{table:online}) except FastCubic/CDHS (which need to compute Hessian-vector product) and SPIDER-SFO$^+$.
Moreover, our simple SSRGD is not worse than FastCubic/CDHS if $n\geq 1/\epsilon$ and is better than SPIDER-SFO$^+$ if $\delta$ is very small (e.g., $\delta\leq 1/\sqrt{n}$) in Table \ref{table:finite}.
\vspace{-3mm}

\begin{algorithm}[t]
	\caption{Simple Stochastic Recursive Gradient Descent (SSRGD)}
	\label{alg:ssrgd_hl}
	\begin{algorithmic}[1]
		\REQUIRE 
		initial point $x_0$, epoch length $m$, minibatch size $b$, step size $\eta$, perturbation radius $r$, threshold gradient $\mathG$
		\FOR {$s=0,1,2,\ldots$}
			\IF{not currently in a super epoch and $\n{\nabla f(x_{sm})}\leq \mathG$} \label{line:epochbegin}
				\STATE $x_{sm}\leftarrow x_{sm}+\xi,$ where $\xi$ uniformly $\sim \mathbb{B}_0(r)$, start a super epoch \\
                // we use super epoch since we do not want to add the perturbation too often near a saddle point
			\ENDIF
            \STATE $v_{sm}\leftarrow \nabla f(x_{sm})$ \label{line:full}
			\FOR {$k=1, 2, \ldots, m$}
                \STATE $t \leftarrow sm+k$
                \STATE $x_{t} \leftarrow x_{t-1} - \eta v_{t-1}$
    	    	\STATE $v_{t}\leftarrow \frac{1}{b}\sum_{i\in I_b}\big(\nabla f_i(x_{t})-\nabla f_i(x_{t-1})\big) + v_{t-1}$ \quad// $I_b$ are i.i.d. uniform samples with $|I_b|=b$ \label{line:v}
        	   	\STATE \algorithmicif~ meet stop condition
        	   		\algorithmicthen~ stop super epoch  \label{line:cond}
			\ENDFOR \label{line:epochend}
		\ENDFOR
    \end{algorithmic}
\end{algorithm}

\vspace{-7mm}
\subsection{Our Contributions}\vspace{-1mm}
In this paper, we propose a simple SSRGD algorithm (described in Algorithm \ref{alg:ssrgd_hl}) showed that a simple perturbation step is enough to find a second-order stationary point for stochastic recursive gradient descent algorithm. Our results and previous results are summarized in Table \ref{table:finite} and \ref{table:online}.
We would like to highlight the following points:
\begin{itemize}
  \item We improve the result in \citep{ge2019stable} to the almost optimal one (i.e., from $n^{2/3}/\epsilon^2$ to $n^{1/2}/\epsilon^2$) since \cite{fang2018spider} provided a lower bound $\Omega(\sqrt{n}/\epsilon^2)$ for finding even just an $\epsilon$-first-order stationary point.
      Note that for the other two $n^{1/2}$ algorithms (i.e., SNVRG$^+$ and SPIDER-SFO$^+$), they both need the negative curvature search subroutine (e.g. Neon2) thus are more complicated in practice and in analysis compared with their first-order guarantee algorithms (SNVRG and SPIDER), while our SSRGD is as simple as its first-order guarantee algorithm just by adding a uniform perturbation sometimes.
  \item For more general nonconvex online (expectation) problems \eqref{eq:online}, we obtain the first algorithm which is as simple as finding first-order stationary points for finding a second-order stationary point with similar state-of-the-art convergence result.
      See the last column of Table \ref{table:online}.
  \item Our simple SSRGD algorithm gets simpler analysis. Also, the result for finding a first-order stationary point is a by-product from our analysis.
        We also give a clear interpretation to show why our analysis for SSRGD algorithm can improve the original SVRG from $n^{2/3}$ to $n^{1/2}$ in Section \ref{sec:ovfirst}. We believe it is very useful for better understanding these two algorithms.
\end{itemize}

\vspace{-5mm}
\section{Preliminaries}
\label{sec:pre}
\vspace{-2mm}

\noindent{{\bf Notation:}}
Let $[n]$ denote the set $\{1,2,\cdots,n\}$ and $\n{\cdot}$ denote the Eculidean norm for a vector and the spectral norm for a matrix.
Let $\inner{u}{v}$ denote the inner product of two vectors $u$ and $v$.
Let $\lambda_{\min}(A)$ denote the smallest eigenvalue of a symmetric matrix $A$.
Let $\mathbb{B}_x(r)$ denote a Euclidean ball with center $x$ and radius $r$.
We use $O(\cdot)$ to hide the constant and $\tdo(\cdot)$ to hide the polylogarithmic factor.
\vspace{1mm}

In this paper, we consider two types of nonconvex problems. The finite-sum problem has the form
\begin{equation}\label{eq:finite}
 \quad \min_{x\in \R^d}f(x):= \frac{1}{n}\sum_{i=1}^{n}f_i(x),
\end{equation}\vspace{-1mm}
where $f(x)$ and all individual $f_i(x)$ are possibly nonconvex.
This form usually models the empirical risk minimization in machine learning problems.

The online (expectation) problem has the form
\begin{equation}\label{eq:online}
 \quad \min_{x\in \R^d}f(x):= \E_{\zeta\sim D}[F(x,\zeta)],
\end{equation}
where $f(x)$ and $F(x,\zeta)$ are possibly nonconvex.
This form usually models the population risk minimization in machine learning problems.

Now, we make standard smoothness assumptions for these two problems.
\begin{assumption}[Gradient Lipschitz]
\label{asp:1}
\begin{enumerate}
  \item For finite-sum problem \eqref{eq:finite}, each $f_i(x)$ is differentiable and has $L$-Lipschitz continuous gradient, i.e.,
\begin{equation}\label{smoothg1}
  \n{\nabla f_i(x_1)-\nabla f_i(x_2)}\leq L\n{x_1-x_2}, \quad \forall x_1, x_2 \in \R^d.
\end{equation}
  \item For online problem \eqref{eq:online}, $F(x,\zeta)$ is differentiable and has $L$-Lipschitz continuous gradient, i.e.,
\begin{equation}\label{smoothg2}
  \n{\nabla F(x_1,\zeta)-\nabla F(x_2,\zeta)}\leq L\n{x_1-x_2}, \quad \forall x_1, x_2 \in \R^d.
\end{equation}
\end{enumerate}
\end{assumption}

\begin{assumption}[Hessian Lipschitz]
\label{asp:2}
\begin{enumerate}
  \item For finite-sum problem \eqref{eq:finite}, each $f_i(x)$ is twice-differentiable and has $\rho$-Lipschitz continuous Hessian, i.e.,
\begin{equation}\label{smoothh1}
  \n{\nabla^2 f_i(x_1)-\nabla^2 f_i(x_2)}\leq \rho\n{x_1-x_2}, \quad \forall x_1, x_2 \in \R^d.
\end{equation}
  \item For online problem \eqref{eq:online}, $F(x,\zeta)$ is twice-differentiable and has $\rho$-Lipschitz continuous Hessian, i.e.,
\begin{equation}\label{smoothh2}
  \n{\nabla^2 F(x_1,\zeta)-\nabla^2 F(x_2,\zeta)}\leq \rho\n{x_1-x_2}, \quad \forall x_1, x_2 \in \R^d.
\end{equation}
\end{enumerate}
\end{assumption}
These two assumptions are standard for finding first-order stationary points (Assumption \ref{asp:1}) and second-order stationary points (Assumption \ref{asp:1} and \ref{asp:2}) for all algorithms in both Table \ref{table:finite} and \ref{table:online}.

Now we define the approximate first-order stationary points and approximate second-order stationary points.
\begin{definition}\label{def:sp}
$x$ is an $\epsilon$-first-order stationary point for a differentiable function $f$ if
\begin{equation}\label{def:sp1}
  \n{\nabla f(x)}\leq \epsilon.
\end{equation}
$x$ is an $(\epsilon,\delta)$-second-order stationary point for a twice-differentiable function $f$ if
\begin{equation}\label{def:sp2}
  \n{\nabla f(x)}\leq \epsilon  ~~and~~  \lambda_{\min}(\nabla^2 f(x))\geq -\delta.
\end{equation}
\end{definition}
The definition of $(\epsilon,\delta)$-second-order stationary point is the same as \citep{allen2018neon2,daneshmand2018escaping,zhou2018finding,fang2018spider} and it generalizes the classical version where $\delta=\sqrt{\rho\epsilon}$ used in \citep{nesterov2006cubic,jin2017escape,ge2019stable}.

\vspace{-1.5mm}
\section{Simple Stochastic Recursive Gradient Descent}
\label{sec:alg}

In this section, we propose the simple stochastic recursive gradient descent algorithm called SSRGD.
The high-level description (which omits the stop condition details in Line \ref{line:cond}) of this algorithm is in Algorithm \ref{alg:ssrgd_hl} and the full algorithm (containing the stop condition) is described in Algorithm \ref{alg:ssrgd}.
Note that we call each outer loop an \emph{\textbf{epoch}}, i.e., iterations $t$ from $sm$ to $(s+1)m$ for an epoch $s$. We call the iterations between the beginning of perturbation and end of perturbation a \emph{\textbf{super epoch}}.

The SSRGD algorithm is based on the stochastic recursive gradient descent which is introduced in \citep{nguyen2017sarah} for convex optimization.
In particular, \cite{nguyen2017sarah} want to save the storage of past gradients in SAGA \citep{defazio2014saga} by using the recursive gradient.
However, this stochastic recursive gradient descent is widely used in recent work for nonconvex optimization such as SPIDER \citep{fang2018spider}, SpiderBoost \citep{wang2018spiderboost} and some variants of SARAH (e.g., ProxSARAH \citep{pham2019proxsarah}).

Recall that in the well-known SVRG algorithm, \cite{johnson2013accelerating} reused a fixed snapshot full gradient $\nabla f(\tx)$ (which is computed at the beginning of each epoch) in the gradient estimator:
\begin{align}\label{eq:svrg}
  v_t\leftarrow \frac{1}{b}\sum_{i\in I_b}\big(\nabla f_i(x_t)-\nabla f_i(\tx)\big)+\nabla f(\tx),
\end{align}
while the stochastic recursive gradient descent uses a recursive update form (more timely update):
\begin{align}\label{eq:srgd}
  v_{t}\leftarrow \frac{1}{b}\sum_{i\in I_b}\big(\nabla f_i(x_{t})-\nabla f_i(x_{t-1})\big) + v_{t-1}.
\end{align}

\begin{algorithm}[t]
	\caption{Simple Stochastic Recursive Gradient Descent (SSRGD)}
	\label{alg:ssrgd}
	\begin{algorithmic}[1]
		\REQUIRE 
		initial point $x_0$, epoch length $m$, minibatch size $b$, step size $\eta$, perturbation radius $r$, threshold gradient $\mathG$, threshold function value $\mathf$, super epoch length $\mathT$
		\STATE $\spe \leftarrow 0$
		\FOR {$s=0,1,2,\ldots$}
			\IF{$\spe = 0$  and $\n{\nabla f(x_{sm})}\leq \mathG$} \label{line:super}
				\STATE {$\spe \leftarrow 1$}
				\STATE {$\tx\leftarrow x_{sm}, t_{\mathrm{init}}\leftarrow sm$}
				\STATE $x_{sm}\leftarrow \tx +\xi,$ where $\xi$ uniformly $\sim \mathbb{B}_0(r)$ \label{line:init}
			\ENDIF
            \STATE $v_{sm}\leftarrow \nabla f(x_{sm})$ \label{line:up1}
			\FOR {$k=1, 2, \ldots, m$}
                \STATE $t \leftarrow sm+k$
                \STATE $x_{t} \leftarrow x_{t-1} - \eta v_{t-1}$ \label{line:update}
    	    	\STATE $v_{t}\leftarrow \frac{1}{b}\sum_{i\in I_b}\big(\nabla f_i(x_{t})-\nabla f_i(x_{t-1})\big) + v_{t-1}$ \quad// $I_b$ are i.i.d. uniform samples with $|I_b|=b$ \label{line:up2}
        	   	\IF {$\spe=1$ and ($f(\tx)-f(x_t)\geq \mathf$ ~or~  $t-t_{\mathrm{init}}\geq \mathT$)}
        	   		\STATE $\spe \leftarrow 0;$ break
				\ELSIF{$\spe=0$}        	   		
        	   		\STATE {break with probability $\frac{1}{m-k+1}$} \\
        // we use random stop since we want to randomly choose a point as the starting point of the next epoch \label{line:randomstop}
        	   	\ENDIF
			\ENDFOR
			\STATE {$x_{(s+1)m}\leftarrow x_t$} \label{line:randompoint}
		\ENDFOR
    \end{algorithmic}
\end{algorithm}

\section{Convergence Results}
\label{sec:res}

Similar to the perturbed GD \citep{jin2017escape} and perturbed SVRG \citep{ge2019stable}, we add simple perturbations to the stochastic recursive gradient descent algorithm to escape saddle points efficiently.
Besides, we also consider the more general online case.
In the following theorems, we provide the convergence results of SSRGD for finding an $\epsilon$-first-order stationary point and an $(\epsilon,\delta)$-second-order stationary point
for both nonconvex finite-sum problem \eqref{eq:finite} and online problem \eqref{eq:online}.
The proofs are provided in Appendix \ref{app:proof}.
We give an overview of the proofs in next Section \ref{sec:ov}.

\subsection{Nonconvex Finite-sum Problem}

\begin{theorem}\label{thm:f1}
Under Assumption \ref{asp:1} (i.e. \eqref{smoothg1}), let $\Delta f:= f(x_0)-f^*$, where $x_0$ is the initial point and $f^*$ is the optimal value of $f$. By letting step size $\eta \leq \frac{\sqrt{5}-1}{2L}$, epoch length $m=\sqrt{n}$ and minibatch size $b=\sqrt{n}$, SSRGD will find an $\epsilon$-first-order stationary point in expectation using
\begin{equation*}
  O\Big(n+\frac{L\Delta f\sqrt{n}}{\epsilon^2}\Big)
\end{equation*}
stochastic gradients for nonconvex finite-sum problem \eqref{eq:finite}.
\end{theorem}

\begin{theorem}\label{thm:f2}
Under Assumption \ref{asp:1} and \ref{asp:2} (i.e. \eqref{smoothg1} and \eqref{smoothh1}), let $\Delta f:= f(x_0)-f^*$, where $x_0$ is the initial point and $f^*$ is the optimal value of $f$. By letting step size $\eta=\tdo(\frac{1}{L})$, epoch length $m=\sqrt{n}$, minibatch size $b=\sqrt{n}$, perturbation radius $r=\tdo\big(\min(\frac{\delta^3}{\rho^2\epsilon}, \frac{\delta^{3/2}}{\rho\sqrt{L}})\big)$, threshold gradient $\mathG=\epsilon$, threshold function value $\mathf=\tdo(\frac{\delta^3}{\rho^2})$ and super epoch length $\mathT=\tdo(\frac{1}{\eta\delta})$, SSRGD will at least once get to an $(\epsilon,\delta)$-second-order stationary point with high probability using
\begin{equation*}
  \tdo\Big(\frac{L\Delta f\sqrt{n}}{\epsilon^2}
+\frac{L\rho^2\Delta f\sqrt{n}}{\delta^4}
+ \frac{\rho^2\Delta fn}{\delta^3}\Big)
\end{equation*}
stochastic gradients for nonconvex finite-sum problem \eqref{eq:finite}.
\end{theorem}

\subsection{Nonconvex Online (Expectation) Problem}

For nonconvex online problem \eqref{eq:online}, one usually needs the following bounded variance assumption.
For notational convenience, we also
consider this online case as the finite-sum form by letting $\nabla f_i(x) :=\nabla F(x,\zeta_i)$ and thinking
of $n$ as infinity (infinite data samples). 
Although we try to write it as finite-sum form, the convergence analysis of optimization
methods in this online case is a little different from the finite-sum case.

\begin{assumption}[Bounded Variance]
\label{asp:var}
For $\forall x\in \R^d$,~ $\E_{i}[\ns{\nabla f_i(x)-\nabla f(x)}]:=\E_{\zeta_i}[\ns{\nabla F(x,\zeta_i)-\nabla f(x)}]\leq \sigma^2$,
where $\sigma>0$ is a constant.
\end{assumption}
Note that this assumption is standard and necessary for this online case since the full gradients are not available (see e.g., \citep{ghadimi2016mini,lei2017non,li2018simple,zhou2018stochastic,fang2018spider,wang2018spiderboost,pham2019proxsarah}).
Moreover, we need to modify the full gradient computation step at the beginning of each epoch to a large batch stochastic gradient computation step (similar to \citep{lei2017non,li2018simple}), i.e.,
change $v_{sm}\leftarrow \nabla f(x_{sm})$ (Line \ref{line:up1} of Algorithm \ref{alg:ssrgd}) to
\begin{align}\label{eq:batch}
 v_{sm}\leftarrow \frac{1}{B}\sum_{j\in I_B}\nabla f_j(x_{sm}),
\end{align}
where $I_B$ are i.i.d. samples with $|I_B|=B$.
We call $B$ the batch size and $b$ the minibatch size.
Also, we need to change $\n{\nabla f(x_{sm})}\leq \mathG$ (Line \ref{line:super} of Algorithm \ref{alg:ssrgd}) to $\n{v_{sm}}\leq \mathG$.

\begin{theorem}\label{thm:o1}
Under Assumption \ref{asp:1} (i.e. \eqref{smoothg2}) and Assumption \ref{asp:var}, let $\Delta f:= f(x_0)-f^*$, where $x_0$ is the initial point and $f^*$ is the optimal value of $f$. By letting step size $\eta\leq \frac{\sqrt{5}-1}{2L}$, batch size $B=\frac{4\sigma^2}{\epsilon^2}$, minibatch size $b=\sqrt{B}=\frac{\sigma}{\epsilon}$ and epoch length $m=b$, SSRGD will find an $\epsilon$-first-order stationary point in expectation using
\begin{equation*}
  O\Big(\frac{\sigma^2}{\epsilon^2}+\frac{L\Delta f \sigma}{\epsilon^3}\Big)
\end{equation*}
stochastic gradients for nonconvex online problem \eqref{eq:online}.
\end{theorem}

For achieving a high probability result of finding second-order stationary points in this online case (i.e., Theorem \ref{thm:o2}), we need a stronger version of Assumption \ref{asp:var} as in the following Assumption \ref{asp:var2}.
\begin{assumption}[Bounded Variance]
\label{asp:var2}
For $\forall i, x$,~ $\ns{\nabla f_i(x)-\nabla f(x)}:=\ns{\nabla F(x,\zeta_i)-\nabla f(x)}\leq \sigma^2$,
where $\sigma>0$ is a constant.
\end{assumption}
We want to point out that Assumption \ref{asp:var2} can be relaxed such that $\n{\nabla f_i(x)-\nabla f(x)}$ has sub-Gaussian tail, i.e., $\E[\exp(\lambda\n{\nabla f_i(x)-\nabla f(x)})]\leq \exp(\lambda^2\sigma^2/2)$, for $\forall \lambda\in \R$. Then it is sufficient for us to get a high probability bound by using Hoeffding bound on these sub-Gaussian variables.
Note that Assumption \ref{asp:var2} (or the relaxed sub-Gaussian version) is also standard in online case for second-order stationary point finding algorithms (see e.g., \citep{allen2018neon2,zhou2018finding,fang2018spider}).

\begin{theorem}\label{thm:o2}
Under Assumption \ref{asp:1}, \ref{asp:2} (i.e. \eqref{smoothg2} and \eqref{smoothh2}) and Assumption \ref{asp:var2}, let $\Delta f:= f(x_0)-f^*$, where $x_0$ is the initial point and $f^*$ is the optimal value of $f$. By letting step size $\eta=\tdo(\frac{1}{L})$, batch size $B=\tdo(\frac{\sigma^2}{\mathG^2})=\tdo(\frac{\sigma^2}{\epsilon^2})$, minibatch size $b=\sqrt{B}=\tdo(\frac{\sigma}{\epsilon})$, epoch length $m=b$, perturbation radius $r=\tdo\big(\min(\frac{\delta^3}{\rho^2\epsilon}, \frac{\delta^{3/2}}{\rho\sqrt{L}})\big)$, threshold gradient $\mathG=\epsilon\leq \delta^2/\rho$, threshold function value $\mathf=\tdo(\frac{\delta^3}{\rho^2})$ and super epoch length $\mathT=\tdo(\frac{1}{\eta\delta})$, SSRGD will at least once get to an $(\epsilon,\delta)$-second-order stationary point with high probability using
\begin{equation*}
  \tdo\Big(\frac{L\Delta f\sigma}{\epsilon^3}
+\frac{\rho^2\Delta f\sigma^2}{\epsilon^2\delta^3}
+ \frac{L\rho^2\Delta f\sigma}{\epsilon\delta^4}\Big)
\end{equation*}
stochastic gradients for nonconvex online problem \eqref{eq:online}.
\end{theorem}

\section{Overview of the Proofs}
\label{sec:ov}

\subsection{Finding First-order Stationary Points}
\label{sec:ovfirst}
In this section, we first show that why SSRGD algorithm can improve previous SVRG type algorithm (see e.g., \citep{li2018simple,ge2019stable}) from $n^{2/3}/\epsilon^2$ to $n^{1/2}/\epsilon^2$.
Then we give a simple high-level proof for achieving the $n^{1/2}/\epsilon^2$ convergence result (i.e., Theorem \ref{thm:f1}).

\vspace{2mm}
\head{Why it can be improved from $n^{2/3}/\epsilon^2$ to $n^{1/2}/\epsilon^2$}
First, we need a key relation between $f(x_t)$ and $f(x_{t-1})$, where $x_t := x_{t-1}-\eta v_{t-1}$,
\begin{align}\label{eq:keyrelation}
f(x_t) \leq f(x_{t-1})
                - \frac{\eta}{2}\ns{\nabla f(x_{t-1})}
                - \big(\frac{1}{2\eta}- \frac{L}{2}\big)\ns{x_t-x_{t-1}}
                + \frac{\eta}{2}\ns{\nabla f(x_{t-1})-v_{t-1}},
\end{align}
where \eqref{eq:keyrelation} holds since $f$ has $L$-Lipschitz continuous gradient (Assumption \ref{asp:1}). The details for obtaining \eqref{eq:keyrelation} can be found in Appendix \ref{app:prooff} (see \eqref{eq:difff}).

Note that \eqref{eq:keyrelation} is very meaningful and also very important for the proofs. The first term $- \frac{\eta}{2}\ns{\nabla f(x_{t-1})}$ indicates that the function value will decrease a lot if the gradient $\nabla f(x_{t-1})$ is large. The second term $- \big(\frac{1}{2\eta}- \frac{L}{2}\big)\ns{x_t-x_{t-1}}$ indicates that the function value will also decrease a lot if the moving distance $x_t-x_{t-1}$ is large (note that here we require the step size $\eta\leq \frac{1}{L}$).
The additional third term $+ \frac{\eta}{2}\ns{\nabla f(x_{t-1})-v_{t-1}}$ exists since we use $v_{t-1}$ as an estimator of the actual gradient $\nabla f(x_{t-1})$ (i.e., $x_t := x_{t-1}-\eta v_{t-1}$). So it may increase the function value if $v_{t-1}$ is a bad direction in this step.

To get an $\epsilon$-first-order stationary point, we want to cancel the last two terms in \eqref{eq:keyrelation}.
Firstly, we want to bound the last variance term. Recall the variance bound (see Equation (29) in \citep{li2018simple}) for SVRG algorithm, i.e., estimator \eqref{eq:svrg}:
\begin{align}
 \E\big[\ns{\nabla f(x_{t-1})-v_{t-1}}\big]
 &\leq \frac{L^2}{b}\E[\ns{x_{t-1}-\tx}]
       \label{eq:svrg_bound}.
\end{align}
In order to connect the last two terms in \eqref{eq:keyrelation}, we use Young's inequality for the second term $\ns{x_t-x_{t-1}}$, i.e., $-\ns{x_t-x_{t-1}} \leq \frac{1}{\alpha}\ns{x_{t-1}-\tx}-\frac{1}{1+\alpha}\ns{x_t-\tx}$ (for any $\alpha>0$).
By plugging this Young's inequality and \eqref{eq:svrg_bound} into \eqref{eq:keyrelation},
we can cancel the last two terms in \eqref{eq:keyrelation} by summing up \eqref{eq:keyrelation} for each epoch, i.e., for each epoch $s$ (i.e., iterations $sm+1\leq t\leq sm+m$), we have (see Equation (35) in \citep{li2018simple})
\begin{align}\label{eq:keyepoch}
\E[f(x_{(s+1)m})] \leq \E[f(x_{sm})]
               - \frac{\eta}{2}\sum_{j=sm+1}^{sm+m}\E[\ns{\nabla f(x_{j-1})}].
\end{align}
However, due to the Young's inequality, we need to let \red{$b\geq m^2$} to cancel the last two terms in \eqref{eq:keyrelation} for obtaining \eqref{eq:keyepoch}, where $b$ denotes minibatch size and $m$ denotes the epoch length.
According to \eqref{eq:keyepoch}, it is not hard to see that $\hx$ is an $\epsilon$-first-order stationary point in expectation (i.e., $\E[\n{\nabla f(\hx)}]\leq \epsilon$) if $\hx$ is chosen uniformly randomly from $\{x_{t-1}\}_{t\in [T]}$ and the number of iterations $T=Sm=\frac{2(f(x_0)-f^*)}{\eta\epsilon^2}$.
Note that for each iteration we need to compute $b+\frac{n}{m}$ stochastic gradients, where we amortize the full gradient computation of the beginning point of each epoch ($n$ stochastic gradients) into each iteration in its epoch (i.e., $n/m$) for simple presentation.
Thus, the convergence result is $T(b+\frac{n}{m})\ge \frac{n^{2/3}}{\epsilon^2}$ since $b\geq m^2$, where equality holds if $b=m^2=n^{2/3}$. Note that here we ignore the factors of $f(x_0)-f^*$ and $\eta=O(1/L)$.

However, for stochastic recursive gradient descent estimator \eqref{eq:srgd}, we can bound the last variance term in \eqref{eq:keyrelation} as (see Equation \eqref{eq:varfinite} in Appendix \ref{app:prooff}):
\begin{align}
 \E\big[\ns{\nabla f(x_{t-1})-v_{t-1}}\big]
 &\leq \frac{L^2}{b}\sum_{j=sm+1}^{t-1}\E[\ns{x_{j}-x_{j-1}}].
       \label{eq:srgd_bound}
\end{align}
Now, the advantage of \eqref{eq:srgd_bound} compared with \eqref{eq:svrg_bound} is that it is already connected to the second term in \eqref{eq:keyrelation}, i.e., moving distances $\{\ns{x_t-x_{t-1}}\}_t$.
Thus we do not need an additional Young's inequality to transform the second term as before. This makes the function value decrease bound tighter.
Similarly, we plug \eqref{eq:srgd_bound} into \eqref{eq:keyrelation} and sum it up for each epoch to cancel the last two terms in \eqref{eq:keyrelation}, i.e., for each epoch $s$, we have (see Equation \eqref{eq:eta} in Appendix \ref{app:prooff})
\begin{align}\label{eq:keyepoch2}
\E[f(x_{(s+1)m})] \leq \E[f(x_{sm})]
        - \frac{\eta}{2}\sum_{j=sm+1}^{sm+m}\E[\ns{\nabla f(x_{j-1})}].
\end{align}
Compared with \eqref{eq:keyepoch} (which requires $b\geq m^2$), here \eqref{eq:keyepoch2} only requires \red{$b\geq m$} due to the tighter function value decrease bound since it does not involve the additional Young's inequality.

\vspace{2mm}
\head{High-level proof for achieving $n^{1/2}/\epsilon^2$ result}
Now, according to \eqref{eq:keyepoch2}, we can use the same above SVRG arguments to show the $n^{1/2}/\epsilon^2$ convergence result of SSRGD, i.e., $\hx$ is an  $\epsilon$-first-order stationary point in expectation (i.e., $\E[\n{\nabla f(\hx)}]\leq \epsilon$) if $\hx$ is chosen uniformly randomly from $\{x_{t-1}\}_{t\in [T]}$ and the number of iterations $T=Sm=\frac{2(f(x_0)-f^*)}{\eta\epsilon^2}$.
Also, for each iteration, we compute $b+\frac{n}{m}$ stochastic gradients.
The only difference is that now the convergence result is $T(b+\frac{n}{m}) = \red{O(\frac{L \Delta f \sqrt{n}}{\epsilon^2})}$ since $b\geq m$ (rather than $b\geq m^2$), where we let $b=m=n^{1/2}$, $\eta=O(1/L)$ and $\Delta f:=f(x_0)-f^*$.
Moreover, it is optimal since it matches the lower bound $\red{\Omega(\frac{L \Delta f \sqrt{n}}{\epsilon^2})}$ provided by \citep{fang2018spider}.

\subsection{Finding Second-order Stationary Points}
\label{sec:ovsecond}
In this section, we give the high-level proof ideas for finding a second-order stationary point with high probability.
Note that our proof is different from that in \citep{ge2019stable} due to the different estimators \eqref{eq:svrg} and \eqref{eq:srgd}.
\cite{ge2019stable} used the estimator \eqref{eq:svrg} and thus their proof is based on the first-order analysis in \citep{li2018simple}.
Here, our SSRGD uses the estimator \eqref{eq:srgd}.
The difference of the first-order analysis between estimator \eqref{eq:svrg} (\citep{li2018simple}) and estimator \eqref{eq:srgd} (this paper) is already discussed in previous Section \ref{sec:ovfirst}.
For the second-order analysis, since the estimator \eqref{eq:srgd} in our SSRGD is more correlated than \eqref{eq:svrg}, thus we will use martingales to handle it.
Besides, different estimators will incur more differences in the detailed proofs of second-order guarantee analysis than that of first-order guarantee analysis.

We divide the proof into two situations, i.e., \emph{large gradients} and \emph{around saddle points}.
According to \eqref{eq:keyepoch2}, a natural way to prove the convergence result is that the function value will decrease at a \emph{desired rate} with high probability.
Note that the amount for function value decrease is at most $\Delta f:=f(x_0)-f^*$.

\vspace{3mm}
\head{Large gradients} $\n{\nabla f(x)}\geq \mathG$ \\
In this situation, due to the large gradients, it is sufficient to adjust the first-order analysis to show that the function value will decrease a lot in an epoch.
Concretely, we want to show that the function value decrease bound \eqref{eq:keyepoch2} holds with high probability by using Azuma-Hoeffding inequality.
Then, according to \eqref{eq:keyepoch2}, it is not hard to see that the desired rate of function value decrease is $O(\eta\mathG^2)=\tdo(\frac{\epsilon^2}{L})$ per iteration in this situation (recall the parameters $\mathG =\epsilon$ and $\eta=\tdo(1/L)$ in our Theorem \ref{thm:f2}).
Also note that we compute $b+\frac{n}{m}=2\sqrt{n}$ stochastic gradients at each iteration (recall $m=b=\sqrt{n}$ in our Theorem \ref{thm:f2}).
Here we amortize the full gradient computation of the beginning point of each epoch ($n$ stochastic gradients) into each iteration in its epoch (i.e., $n/m$) for simple presentation (we will analyze this more rigorously in the detailed proofs in appendices).
Thus the number of stochastic gradient computation is at most $\tdo(\sqrt{n}\frac{\Delta f}{\epsilon^2/L})=\red{\tdo(\frac{L \Delta f \sqrt{n}}{\epsilon^2})}$ for this large gradients situation.

For the proof, to show the function value decrease bound \eqref{eq:keyepoch2} holds with high probability, we need to show that the bound for variance term ($\ns{v_k-\nabla f(x_k)}$) holds with high probability.
Note that the gradient estimator $v_k$ defined in \eqref{eq:srgd} is correlated with previous $v_{k-1}$. Fortunately, let $y_k:=v_k-\nabla f(x_k)$, then it is not hard to see that $\{y_k\}$ is a martingale vector sequence with respect to a filtration $\{\mathscr{F}_k\}$ such that $\E[y_k|\mathscr{F}_{k-1}]=y_{k-1}$.
Moreover, let $\{z_k\}$ denote the associated martingale difference sequence with respect to the filtration $\{\mathscr{F}_k\}$, i.e., $z_k:=y_k-\E[y_k|\mathscr{F}_{k-1}]=y_k-y_{k-1}$ and
$\E[z_k|\mathscr{F}_{k-1}]=0.$
Thus to bound the variance term $\ns{v_k-\nabla f(x_k)}$ with high probability, it is sufficient to bound the martingale sequence $\{y_k\}$. This can be bounded with high probability by using the martingale Azuma-Hoeffding inequality.
Note that in order to apply Azuma-Hoeffding inequality, we first need to use the Bernstein inequality to bound the associated difference sequence $\{z_k\}$.
In sum, we will get the high probability function value decrease bound by applying these two inequalities (see \eqref{eq:eta2} in Appendix \ref{app:prooff}).

Note that \eqref{eq:eta2} only guarantees function value decrease when the summation of gradients in this epoch is large. However, in order to connect the guarantees between first situation (large gradients) and second situation (around saddle points), we need to show guarantees that are related to the \emph{gradient of the starting point} of each epoch (see Line \ref{line:super} of Algorithm \ref{alg:ssrgd}).
Similar to \citep{ge2019stable}, we achieve this by stopping the epoch at a uniformly random point (see Line \ref{line:randomstop} of Algorithm \ref{alg:ssrgd}).
We use the following lemma to connect these two situations (large gradients and around saddle points):

\begin{lemma}[Connection of Two Situations]\label{lem:first}
For any epoch $s$, let $x_t$ be a point uniformly sampled from this epoch $\{x_{j} \}_{j=sm}^{(s+1)m}$ and choose the step size $\eta \leq \frac{\sqrt{4C'^2+1}-1}{2C'^2L}$ (where $C'=O(\log\frac{dn}{\zeta})=\tdo(1)$) and the minibatch size $b\geq m$.
Then for any $\mathG$, we have two cases:
\begin{enumerate}
\item If at least half of points in this epoch have gradient norm no larger than $\mathG$, then $\|\nabla f(x_t) \| \le \mathG$ holds with probability at least $1/2$;
\item Otherwise, we know $f(x_{sm}) - f(x_t) \ge \frac{\eta m\mathG^2}{8}$ holds with probability at least $1/5.$
\end{enumerate}
Moreover, $f(x_t) \le f(x_{sm})$ holds with high probability no matter which case happens.
\end{lemma}

Note that if Case 2 happens, the function value already decreases a lot in this epoch $s$ (as we already discussed at the beginning of this situation). Otherwise Case 1 happens, we know the starting point of the next epoch $x_{(s+1)m}=x_t$ (i.e., Line \ref{line:randompoint} of Algorithm \ref{alg:ssrgd}), then we know $\n{\nabla f(x_{(s+1)m})}=\n{\nabla f(x_{t})} \leq \mathG$. Then we will start a super epoch (see Line \ref{line:super} of Algorithm \ref{alg:ssrgd}). This corresponds to the following second situation (around saddle points). Note that if $\lambda_{\min}(\nabla^2 f(x_{(s+1)m}))> -\delta$, this point $x_{(s+1)m}$ is already an $(\epsilon,\delta)$-second-order stationary point (recall $\mathG = \epsilon$ in our Theorem \ref{thm:f2}).

\vspace{5mm}
\head{Around saddle points} $\n{\nabla f(\tx)}\leq \mathG$ and $\lambda_{\min}(\nabla^2 f(\tx))\leq -\delta$ at the initial point $\tx$ of a super epoch\\
In this situation, we want to show that the function value will decrease a lot in a \emph{super epoch} (instead of an epoch as in the first situation) with high probability by adding a random perturbation at the initial point $\tx$. To simplify the presentation, we use $x_0:=\tx+\xi$ to denote the starting point of the super epoch after the perturbation, where $\xi$ uniformly $\sim \mathbb{B}_0(r)$ and the perturbation radius is $r$ (see Line \ref{line:init} in Algorithm \ref{alg:ssrgd}).
Following the classical widely used \emph{two-point analysis} developed in \citep{jin2017escape}, we consider two coupled points $x_0$ and $x_0'$ with $w_0:=x_0-x_0'=r_0e_1$, where $r_0$ is a scalar and $e_1$ denotes the smallest eigenvector direction of Hessian $\nabla^2 f(\tx)$. Then we get two coupled sequences $\{x_t\}$ and $\{x_t'\}$ by running SSRGD update steps (Line \ref{line:up1}--\ref{line:up2} of Algorithm \ref{alg:ssrgd}) with the same choice of minibatches (i.e., $I_b$'s in Line \ref{line:up2} of Algorithm \ref{alg:ssrgd}) for a super epoch.
We will show that at least one of these two coupled sequences will decrease the function value a lot (escape the saddle point) with high probability, i.e.,
\begin{align}\label{eq:funcd_main}
 \exists t\leq\mathT, \mathrm{~~such ~that~~} \max\{f(x_0)-f(x_t), f(x_0')-f(x_t')\} \geq 2\mathf.
\end{align}
Similar to the classical argument in \citep{jin2017escape}, according to \eqref{eq:funcd_main}, we know that in the random perturbation ball, the stuck points can only be a short interval in the $e_1$ direction, i.e.,
at least one of two points in the $e_1$ direction will escape the saddle point if their distance is larger than $r_0=\frac{\zeta' r}{\sqrt{d}}$.
Thus, we know that the probability of the starting point $x_0=\tx+\xi$ (where $\xi$ uniformly $\sim \mathbb{B}_0(r)$) located in the stuck region is less than $\zeta'$ (see \eqref{eq:goodx0} in Appendix \ref{app:prooff}).
By a union bound ($x_0$ is not in a stuck region and \eqref{eq:funcd_main} holds), with high probability, we have
\begin{align}\label{eq:escape0_main}
\exists t\leq\mathT~, f(x_0)-f(x_t) \geq 2\mathf.
\end{align}

Note that the initial point of this super epoch is $\tx$ before the perturbation (see Line \ref{line:init} of Algorithm \ref{alg:ssrgd}), thus we also need to show that the perturbation step $x_0=\tx+\xi$ (where $\xi$ uniformly $\sim \mathbb{B}_0(r)$) does not increase the function value a lot, i.e.,
\begin{align}
f(x_0)&\leq f(\tx) +\inner{\nabla f(\tx)}{x_0-\tx}
                    + \frac{L}{2}\ns{x_0-\tx}  \notag \\
        &\leq f(\tx) +\mathG \cdot r +\frac{L}{2}r^2 \notag\\
        &= f(\tx) +\mathf, \label{eq:perturbless_main}
\end{align}
where the second inequality holds since the initial point $\tx$ satisfying $\n{\nabla f(\tx)}\leq \mathG$ and the perturbation radius is $r$, and the last equality holds by letting the perturbation radius $r$ small enough.
By combining \eqref{eq:escape0_main} and \eqref{eq:perturbless_main}, we obtain with high probability
\begin{align}
f(\tx)-f(x_t)=f(\tx)-f(x_0)+f(x_0)-f(x_t) \geq -\mathf+2\mathf=\mathf.
\end{align}

Now, we can obtain the desired rate of function value decrease in this situation is $\frac{\mathf}{\mathT}=\tdo(\frac{\delta^3/\rho^2}{1/(\eta\delta)})
=\tdo(\frac{\delta^4}{L\rho^2})$ per iteration (recall the parameters $\mathf=\tdo(\delta^3/\rho^2)$, $\mathT=\tdo(1/(\eta\delta))$ and $\eta=\tdo(1/L)$ in our Theorem \ref{thm:f2}).
Same as before, we compute $b+\frac{n}{m}=2\sqrt{n}$ stochastic gradients at each iteration (recall $m=b=\sqrt{n}$ in our Theorem \ref{thm:f2}).
Thus the number of stochastic gradient computation is at most $\tdo(\sqrt{n}\frac{\Delta f}{\delta^4/(L\rho^2)})=\red{\tdo(\frac{L \rho^2\Delta f \sqrt{n}}{\delta^4})}$ for this around saddle points situation.

Now, the remaining thing is to prove \eqref{eq:funcd_main}.
It can be proved by contradiction. Assume the contrary, $f(x_0)-f(x_t)<2\mathf$ and $f(x_0')-f(x_t')<2\mathf$.
First, we show that if function value does not decrease a lot, then all iteration points are not far from the starting point with high probability.

\begin{lemma}[Localization]\label{lem:local}
Let $\{x_t\}$ denote the sequence by running SSRGD update steps (Line \ref{line:up1}--\ref{line:up2} of Algorithm \ref{alg:ssrgd}) from $x_0$.
Moreover, let the step size $\eta\leq \frac{1}{2C'L}$ and minibatch size $b\geq m$, with probability $1-\zeta$, we have
\begin{align}\label{eq:local_main}
\forall t,~~ \n{x_t-x_0}\leq \sqrt{\frac{4t(f(x_0)-f(x_t))}{C'L}},
\end{align}
where $C'=O(\log\frac{dt}{\zeta})=\tdo(1)$.
\end{lemma}

Then we show that the stuck region is relatively small in the random perturbation ball, i.e., at least one of $x_t$ and $x_t'$ will go far away from their starting point $x_0$ and $x_0'$ with high probability.
\begin{lemma}[Small Stuck Region]\label{lem:smallstuck}
If the initial point $\tx$ satisfies $-\gamma:=\lambda_{\min}(\nabla^2 f(\tx))\leq -\delta$,
then let $\{x_t\}$ and $\{x_t'\}$ be two coupled sequences by running SSRGD update steps (Line \ref{line:up1}--\ref{line:up2} of Algorithm \ref{alg:ssrgd}) with the same choice of minibatches (i.e., $I_b$'s in Line \ref{line:up2}) from $x_0$ and $x_0'$ with $w_0:=x_0-x_0'=r_0e_1$, where $x_0\in\mathbb{B}_{\tx}(r)$, $x_0'\in\mathbb{B}_{\tx}(r)$ , $r_0=\frac{\zeta' r}{\sqrt{d}}$ and $e_1$ denotes the smallest eigenvector direction of Hessian $\nabla^2 f(\tx)$.
Moreover, let the super epoch length $\mathT=\frac{2\log(\frac{8\delta\sqrt{d}}{C_1\rho\zeta' r})}{\eta\delta}=\tdo(\frac{1}{\eta\delta})$, the step size $\eta\leq \min\big(\frac{1}{8\log(\frac{8\delta\sqrt{d}}{C_1\rho\zeta' r})L}, \frac{1}{4C_2L\log \mathT}\big)=\tdo(\frac{1}{L})$, minibatch size $b\geq m$ and
the perturbation radius $r\leq \frac{\delta}{C_1 \rho}$, then with probability $1-\zeta$, we have
\begin{align}\label{eq:stuck_main}
\exists T\leq \mathT,~~ \max\{\n{x_T-x_0}, \n{x_T'-x_0'}\}\geq \frac{\delta}{C_1\rho},
\end{align}
where $C_1\geq \frac{20C_2}{\eta L}$ and $C_2=O(\log\frac{d \mathT}{\zeta})=\tdo(1)$.
\end{lemma}

Based on these two lemmas, we are ready to show that \eqref{eq:funcd_main} holds with high probability. Without loss of generality, we assume $\n{x_T-x_0} \geq \frac{\delta}{C_1\rho}$ in \eqref{eq:stuck_main} (note that \eqref{eq:local_main} holds for both $\{x_t\}$ and $\{x_t'\}$), then by plugging it into \eqref{eq:local_main} to obtain
\begin{align}
 \sqrt{\frac{4T(f(x_0)-f(x_T))}{C'L}}&\geq \frac{\delta}{C_1\rho} \notag\\
 f(x_0)-f(x_T) &\geq \frac{C'L\delta^2}{4C_1^2\rho^2T} \notag\\
    &\geq \frac{\eta C'L\delta^3}{8C_1^2\rho^2\log(\frac{8\delta\sqrt{d}}{C_1\rho\zeta' r})} \notag\\
    &=\frac{\delta^3}{C_1'\rho^2} \notag\\
    &=2\mathf,\notag
\end{align}
where the last inequality is due to $T\leq \mathT$ and the first equality holds by letting $C_1'=\frac{8C_1^2\log(\frac{8\delta\sqrt{d}}{C_1\rho\zeta' r})}{\eta C'L} =\tdo(1)$ (recall the parameters $\mathf=\tdo(\delta^3/\rho^2)$ and $\eta=\tdo(1/L)$ in our Theorem \ref{thm:f2}).
Now, the high-level proof for this situation is finished.

In sum, the number of stochastic gradient computation is at most
$\tdo(\frac{L \Delta f \sqrt{n}}{\epsilon^2})$ for the large gradients situation
and is at most
$\tdo(\frac{L \rho^2\Delta f \sqrt{n}}{\delta^4})$
for the around saddle points situation.
Moreover, for the classical version where $\delta=\sqrt{\rho\epsilon}$ \citep{nesterov2006cubic,jin2017escape}, then $\tdo(\frac{L \rho^2\Delta f \sqrt{n}}{\delta^4})=\tdo(\frac{L \Delta f \sqrt{n}}{\epsilon^2})$, i.e., both situations get the same stochastic gradient complexity. This also matches the convergence result for finding first-order stationary points (see our Theorem \ref{thm:f1}) if we ignore the logarithmic factor.
More importantly, it also almost matches the lower bound $\Omega(\frac{L \Delta f \sqrt{n}}{\epsilon^2})$ provided by \citep{fang2018spider} for finding even just an $\epsilon$-first-order stationary point.

Finally, we point out that there is an extra term $\frac{\rho^2\Delta fn}{\delta^3}$ in Theorem \ref{thm:f2} beyond these two terms obtained from the above two situations.
The reason is that we amortize the full gradient computation of the beginning point of each epoch ($n$ stochastic gradients) into each iteration in its epoch (i.e., $n/m$) for simple presentation. We will analyze this more rigorously in the appendices, which incurs the term $\frac{\rho^2\Delta fn}{\delta^3}$.
For the more general online problem \eqref{eq:online}, the high-level proofs are almost the same as the finite-sum problem \eqref{eq:finite}.
The difference is that we need to use more concentration bounds in the detailed proofs since the full gradients are not available in online case.

\section{Conclusion}
In this paper, we focus on developing simple algorithms that have theoretical second-order guarantee for nonconvex finite-sum problems and more general nonconvex online problems.
Concretely,
we propose a simple perturbed version of stochastic recursive gradient descent algorithm (called SSRGD), which is as simple as its first-order stationary point finding algorithm (just by adding a random perturbation sometimes) and thus can be simply applied in practice for escaping saddle points (finding local minima).
Moreover, the theoretical convergence results of SSRGD for finding second-order stationary points (local minima) almost match the theoretical results for finding first-order stationary points and these results are near-optimal as they almost match the lower bound.

\subsection*{Acknowledgments}
The author would like to thank Rong Ge since the author learned a lot under his genuine guidance during the visit at Duke.

\newpage
\bibliographystyle{plainnat}
\bibliography{draft}

\begin{thebibliography}{33}
\providecommand{\natexlab}[1]{#1}
\providecommand{\url}[1]{\texttt{#1}}
\expandafter\ifx\csname urlstyle\endcsname\relax
  \providecommand{\doi}[1]{doi: #1}\else
  \providecommand{\doi}{doi: \begingroup \urlstyle{rm}\Url}\fi

\bibitem[Agarwal et~al.(2016)Agarwal, Allen-Zhu, Bullins, Hazan, and
  Ma]{agarwal2016finding}
Naman Agarwal, Zeyuan Allen-Zhu, Brian Bullins, Elad Hazan, and Tengyu Ma.
\newblock Finding approximate local minima for nonconvex optimization in linear
  time.
\newblock \emph{arXiv preprint arXiv:1611.01146}, 2016.

\bibitem[Allen-Zhu(2018)]{allen2018natasha}
Zeyuan Allen-Zhu.
\newblock Natasha 2: Faster non-convex optimization than sgd.
\newblock In \emph{Advances in Neural Information Processing Systems}, pages
  2680--2691, 2018.

\bibitem[Allen-Zhu and Hazan(2016)]{allen2016variance}
Zeyuan Allen-Zhu and Elad Hazan.
\newblock Variance reduction for faster non-convex optimization.
\newblock In \emph{International Conference on Machine Learning}, pages
  699--707, 2016.

\bibitem[Allen-Zhu and Li(2018)]{allen2018neon2}
Zeyuan Allen-Zhu and Yuanzhi Li.
\newblock Neon2: Finding local minima via first-order oracles.
\newblock In \emph{Advances in Neural Information Processing Systems}, pages
  3720--3730, 2018.

\bibitem[Bhojanapalli et~al.(2016)Bhojanapalli, Neyshabur, and
  Srebro]{bhojanapalli2016global}
Srinadh Bhojanapalli, Behnam Neyshabur, and Nati Srebro.
\newblock Global optimality of local search for low rank matrix recovery.
\newblock In \emph{Advances in Neural Information Processing Systems}, pages
  3873--3881, 2016.

\bibitem[Carmon et~al.(2016)Carmon, Duchi, Hinder, and
  Sidford]{carmon2016accelerated}
Yair Carmon, John~C Duchi, Oliver Hinder, and Aaron Sidford.
\newblock Accelerated methods for non-convex optimization.
\newblock \emph{arXiv preprint arXiv:1611.00756}, 2016.

\bibitem[Chung and Lu(2006)]{chung2006concentration}
Fan Chung and Linyuan Lu.
\newblock Concentration inequalities and martingale inequalities: a survey.
\newblock \emph{Internet Mathematics}, 3\penalty0 (1):\penalty0 79--127, 2006.

\bibitem[Daneshmand et~al.(2018)Daneshmand, Kohler, Lucchi, and
  Hofmann]{daneshmand2018escaping}
Hadi Daneshmand, Jonas Kohler, Aurelien Lucchi, and Thomas Hofmann.
\newblock Escaping saddles with stochastic gradients.
\newblock \emph{arXiv preprint arXiv:1803.05999}, 2018.

\bibitem[Defazio et~al.(2014)Defazio, Bach, and
  Lacoste-Julien]{defazio2014saga}
Aaron Defazio, Francis Bach, and Simon Lacoste-Julien.
\newblock Saga: A fast incremental gradient method with support for
  non-strongly convex composite objectives.
\newblock In \emph{Advances in Neural Information Processing Systems}, pages
  1646--1654, 2014.

\bibitem[Du et~al.(2017)Du, Jin, Lee, Jordan, Singh, and
  Poczos]{du2017gradient}
Simon~S Du, Chi Jin, Jason~D Lee, Michael~I Jordan, Aarti Singh, and Barnabas
  Poczos.
\newblock Gradient descent can take exponential time to escape saddle points.
\newblock In \emph{Advances in Neural Information Processing Systems}, pages
  1067--1077, 2017.

\bibitem[Fang et~al.(2018)Fang, Li, Lin, and Zhang]{fang2018spider}
Cong Fang, Chris~Junchi Li, Zhouchen Lin, and Tong Zhang.
\newblock Spider: Near-optimal non-convex optimization via stochastic
  path-integrated differential estimator.
\newblock In \emph{Advances in Neural Information Processing Systems}, pages
  687--697, 2018.

\bibitem[Ge et~al.(2015)Ge, Huang, Jin, and Yuan]{ge2015escaping}
Rong Ge, Furong Huang, Chi Jin, and Yang Yuan.
\newblock Escaping from saddle points --- online stochastic gradient for tensor
  decomposition.
\newblock In \emph{Conference on Learning Theory}, pages 797--842, 2015.

\bibitem[Ge et~al.(2016)Ge, Lee, and Ma]{ge2016matrix}
Rong Ge, Jason~D Lee, and Tengyu Ma.
\newblock Matrix completion has no spurious local minimum.
\newblock In \emph{Advances in Neural Information Processing Systems}, pages
  2973--2981, 2016.

\bibitem[Ge et~al.(2017)Ge, Lee, and Ma]{ge2017learning}
Rong Ge, Jason~D Lee, and Tengyu Ma.
\newblock Learning one-hidden-layer neural networks with landscape design.
\newblock \emph{arXiv preprint arXiv:1711.00501}, 2017.

\bibitem[Ge et~al.(2019)Ge, Li, Wang, and Wang]{ge2019stable}
Rong Ge, Zhize Li, Weiyao Wang, and Xiang Wang.
\newblock Stabilized svrg: Simple variance reduction for nonconvex
  optimization.
\newblock In \emph{Conference on Learning Theory}, 2019.

\bibitem[Ghadimi et~al.(2016)Ghadimi, Lan, and Zhang]{ghadimi2016mini}
Saeed Ghadimi, Guanghui Lan, and Hongchao Zhang.
\newblock Mini-batch stochastic approximation methods for nonconvex stochastic
  composite optimization.
\newblock \emph{Mathematical Programming}, 155\penalty0 (1-2):\penalty0
  267--305, 2016.

\bibitem[Hoeffding(1963)]{hoeffding1963probability}
Wassily Hoeffding.
\newblock Probability inequalities for sums of bounded random variables.
\newblock \emph{Journal of the American Statistical Association}, 58\penalty0
  (301):\penalty0 13--30, 1963.

\bibitem[Jin et~al.(2017)Jin, Ge, Netrapalli, Kakade, and
  Jordan]{jin2017escape}
Chi Jin, Rong Ge, Praneeth Netrapalli, Sham~M Kakade, and Michael~I Jordan.
\newblock How to escape saddle points efficiently.
\newblock In \emph{Proceedings of the 34th International Conference on Machine
  Learning-Volume 70}, pages 1724--1732. JMLR. org, 2017.

\bibitem[Johnson and Zhang(2013)]{johnson2013accelerating}
Rie Johnson and Tong Zhang.
\newblock Accelerating stochastic gradient descent using predictive variance
  reduction.
\newblock In \emph{Advances in neural information processing systems}, pages
  315--323, 2013.

\bibitem[Lei et~al.(2017)Lei, Ju, Chen, and Jordan]{lei2017non}
Lihua Lei, Cheng Ju, Jianbo Chen, and Michael~I Jordan.
\newblock Non-convex finite-sum optimization via scsg methods.
\newblock In \emph{Advances in Neural Information Processing Systems}, pages
  2345--2355, 2017.

\bibitem[Li and Li(2018)]{li2018simple}
Zhize Li and Jian Li.
\newblock A simple proximal stochastic gradient method for nonsmooth nonconvex
  optimization.
\newblock In \emph{Advances in Neural Information Processing Systems}, pages
  5569--5579, 2018.

\bibitem[Nesterov(2004)]{nesterov2014introductory}
Yurii Nesterov.
\newblock \emph{Introductory Lectures on Convex Optimization: A Basic Course}.
\newblock Kluwer, 2004.

\bibitem[Nesterov and Polyak(2006)]{nesterov2006cubic}
Yurii Nesterov and Boris~T Polyak.
\newblock Cubic regularization of newton method and its global performance.
\newblock \emph{Mathematical Programming}, 108\penalty0 (1):\penalty0 177--205,
  2006.

\bibitem[Nguyen et~al.(2017)Nguyen, Liu, Scheinberg, and
  Tak{\'a}{\v{c}}]{nguyen2017sarah}
Lam~M Nguyen, Jie Liu, Katya Scheinberg, and Martin Tak{\'a}{\v{c}}.
\newblock Sarah: A novel method for machine learning problems using stochastic
  recursive gradient.
\newblock In \emph{Proceedings of the 34th International Conference on Machine
  Learning-Volume 70}, pages 2613--2621. JMLR. org, 2017.

\bibitem[Pham et~al.(2019)Pham, Nguyen, Phan, and Tran-Dinh]{pham2019proxsarah}
Nhan~H Pham, Lam~M Nguyen, Dzung~T Phan, and Quoc Tran-Dinh.
\newblock Proxsarah: An efficient algorithmic framework for stochastic
  composite nonconvex optimization.
\newblock \emph{arXiv preprint arXiv:1902.05679}, 2019.

\bibitem[Reddi et~al.(2016)Reddi, Hefny, Sra, P{\'o}czos, and
  Smola]{reddi2016stochastic}
Sashank~J Reddi, Ahmed Hefny, Suvrit Sra, Barnab{\'a}s P{\'o}czos, and Alex
  Smola.
\newblock Stochastic variance reduction for nonconvex optimization.
\newblock In \emph{International conference on machine learning}, pages
  314--323, 2016.

\bibitem[Tao and Vu(2015)]{tao2015random}
Terence Tao and Van Vu.
\newblock Random matrices: Universality of local spectral statistics of
  non-hermitian matrices.
\newblock \emph{The Annals of Probability}, 43\penalty0 (2):\penalty0 782--874,
  2015.

\bibitem[Tropp(2011)]{tropp2011user}
Joel~A Tropp.
\newblock User-friendly tail bounds for matrix martingales.
\newblock Technical report, CALIFORNIA INST OF TECH PASADENA, 2011.

\bibitem[Tropp(2012)]{tropp2012user}
Joel~A Tropp.
\newblock User-friendly tail bounds for sums of random matrices.
\newblock \emph{Foundations of computational mathematics}, 12\penalty0
  (4):\penalty0 389--434, 2012.

\bibitem[Wang et~al.(2018)Wang, Ji, Zhou, Liang, and
  Tarokh]{wang2018spiderboost}
Zhe Wang, Kaiyi Ji, Yi~Zhou, Yingbin Liang, and Vahid Tarokh.
\newblock Spiderboost: A class of faster variance-reduced algorithms for
  nonconvex optimization.
\newblock \emph{arXiv preprint arXiv:1810.10690}, 2018.

\bibitem[Xu et~al.(2018)Xu, Rong, and Yang]{xu2018first}
Yi~Xu, Jing Rong, and Tianbao Yang.
\newblock First-order stochastic algorithms for escaping from saddle points in
  almost linear time.
\newblock In \emph{Advances in Neural Information Processing Systems}, pages
  5535--5545, 2018.

\bibitem[Zhou et~al.(2018{\natexlab{a}})Zhou, Xu, and Gu]{zhou2018finding}
Dongruo Zhou, Pan Xu, and Quanquan Gu.
\newblock Finding local minima via stochastic nested variance reduction.
\newblock \emph{arXiv preprint arXiv:1806.08782}, 2018{\natexlab{a}}.

\bibitem[Zhou et~al.(2018{\natexlab{b}})Zhou, Xu, and Gu]{zhou2018stochastic}
Dongruo Zhou, Pan Xu, and Quanquan Gu.
\newblock Stochastic nested variance reduction for nonconvex optimization.
\newblock \emph{arXiv preprint arXiv:1806.07811}, 2018{\natexlab{b}}.

\end{thebibliography}

\newpage
\appendix

\section{Tools}
In this appendix, we recall some classical concentration bounds for matrices and vectors.
\begin{proposition} [Bernstein Inequality \citep{tropp2012user}]\label{prop:bernstein_original}
Consider a finite sequence $\{Z_k\}$ of independent, random matrices with dimension $d_1\times d_2$. Assume that each random matrix satisfies
\begin{align*}
\E[Z_k]=0  ~~and~~ \n{Z_k}\leq R ~~almost ~surely.
\end{align*}
Define
$$\sigma^2:= \max\Big\{\big\|\sum_k \E[Z_k Z_k^T]\big\|, \big\| \sum_k \E[Z_k^T Z_k]\big\| \Big\}.$$
Then, for all $t\geq 0$,
$$\pr\Big\{\big\|\sum_k Z_k\big\|\geq t \Big\}\leq (d_1+d_2) \exp\Big(\frac{-t^2/2}{\sigma^2+Rt/3}\Big).$$
\end{proposition}
In our proof, we only need its special case vector version as follows, where $z_k=v_k-\E[v_k]$.
\begin{proposition} [Bernstein Inequality \citep{tropp2012user}]\label{prop:bernstein}
Consider a finite sequence $\{v_k\}$ of independent, random vectors with dimension $d$. Assume that each random matrix satisfies
\begin{align*}
\n{v_k-\E[v_k]}\leq R ~~almost ~surely.
\end{align*}
Define
$$\sigma^2 := \sum_{k}\E\ns{v_k-\E[v_k]}.$$
Then, for all $t\geq 0$,
$$\pr\Big\{\big\|\sum_k (v_k-\E[v_k])\big\|\geq t \Big\}\leq (d+1) \exp\Big(\frac{-t^2/2}{\sigma^2+Rt/3}\Big).$$
\end{proposition}

Moreover, we also need the martingale concentration bounds, i.e., Azuma-Hoffding inequality.
Now, we will only write the vector version not repeat the more general matrix version.
\begin{proposition} [Azuma-Hoeffding Inequality \citep{hoeffding1963probability,tropp2011user}]\label{prop:azuma}
Consider a martingale vector sequence $\{y_k\}$ with dimension $d$, and let $\{z_k\}$ denote the associated martingale difference sequence with respect to a filtration $\{\mathscr{F}_k\}$, i.e., $z_k:=y_k-\E[y_k|\mathscr{F}_{k-1}]=y_k-y_{k-1}$ and $\E[z_k|\mathscr{F}_{k-1}]=0$.
Suppose that $\{z_k\}$ satisfies
\begin{align}
\n{z_k}=\n{y_k-y_{k-1}} \leq c_k ~~almost ~surely. \label{eq:diffbound}
\end{align}
Then, for all $t\geq 0$,
$$\pr\Big\{\|y_k-y_0\|\geq t \Big\}\leq (d+1) \exp\Big(\frac{-t^2}{8\sum_{i=1}^k c_i^2}\Big).$$
\end{proposition}

However, the assumption that $\n{z_k} \leq c_k$ in \eqref{eq:diffbound} with probability one sometime fails.
Fortunately, the Azuma-Hoffding inequality also holds with a slackness if $\n{z_k} \leq c_k$ with high probability.

\begin{proposition} [Azuma-Hoeffding Inequality with High Probability \citep{chung2006concentration,tao2015random}]\label{prop:azumahigh}
Con\-sider a martingale vector sequence $\{y_k\}$ with dimension $d$, and let $\{z_k\}$ denote the associated martingale difference sequence with respect to a filtration $\{\mathscr{F}_k\}$, i.e., $z_k:=y_k-\E[y_k|\mathscr{F}_{k-1}]=y_k-y_{k-1}$ and $\E[z_k|\mathscr{F}_{k-1}]=0$.
Suppose that $\{z_k\}$ satisfies
\begin{align*}
\n{z_k}=\n{y_k-y_{k-1}} \leq c_k ~~with~ high~ probability~ 1-\zeta_k.
\end{align*}
Then, for all $t\geq 0$,
$$\pr\Big\{\|y_k-y_0\|\geq t \Big\}\leq (d+1) \exp\Big(\frac{-t^2}{8\sum_{i=1}^k c_i^2}\Big)+\sum_{i=1}^k\zeta_k.$$
\end{proposition}

\section{Missing Proofs}
\label{app:proof}

In this appendix, we provide the detailed proofs for Theorem \ref{thm:f1}--\ref{thm:o2}.
\subsection{Proofs for Finite-sum Problem}
\label{app:prooff}
In this section, we provide the detailed proofs for nonconvex finite-sum problem \eqref{eq:finite} (i.e., Theorem \ref{thm:f1}--\ref{thm:f2}).

First, we obtain the relation between $f(x_t)$ and $f(x_{t-1})$ as follows similar to \citep{li2018simple,ge2019stable}, where we let $x_t := x_{t-1}-\eta v_{t-1}$ and $\bx_t := x_{t-1}-\eta \nabla f(x_{t-1})$,
\begin{align}
f(x_t) \leq& f(x_{t-1})
                    + \inner{\nabla f(x_{t-1})}{x_t-x_{t-1}}
                    + \frac{L}{2}\ns{x_t-x_{t-1}} \label{eq:lp} \\
         =&f(x_{t-1})
                    + \inner{\nabla f(x_{t-1})-v_{t-1}}{x_t-x_{t-1}}
                    + \inner{v_{t-1}}{x_t-x_{t-1}}
                    + \frac{L}{2}\ns{x_t-x_{t-1}} \notag \\
         =&f(x_{t-1})
                    + \inner{\nabla f(x_{t-1})-v_{t-1}}{-\eta v_{t-1}}
                    - \big(\frac{1}{\eta}- \frac{L}{2}\big)\ns{x_t-x_{t-1}} \notag \\
         =&f(x_{t-1})
                    + \eta\ns{\nabla f(x_{t-1})-v_{t-1}}
                    - \eta\inner{\nabla f(x_{t-1})-v_{t-1}}{\nabla f(x_{t-1})}
                    - \big(\frac{1}{\eta}- \frac{L}{2}\big)\ns{x_t-x_{t-1}} \notag \\
         =&f(x_{t-1})
                    + \eta\ns{\nabla f(x_{t-1})-v_{t-1}}
                    - \frac{1}{\eta}\inner{x_t-\bx_t}{x_{t-1}-\bx_t}
                    - \big(\frac{1}{\eta}- \frac{L}{2}\big)\ns{x_t-x_{t-1}} \notag \\
         =&f(x_{t-1})
                    + \eta\ns{\nabla f(x_{t-1})-v_{t-1}}
                    - \big(\frac{1}{\eta}- \frac{L}{2}\big)\ns{x_t-x_{t-1}} \notag\\
            &\qquad\qquad\qquad -\frac{1}{2\eta}\big(\ns{x_t-\bx_t}+\ns{x_{t-1}-\bx_t}
                            -\ns{x_t-x_{t-1}}\big) \notag \\
         =&f(x_{t-1})
                    + \frac{\eta}{2}\ns{\nabla f(x_{t-1})-v_{t-1}}
                    - \frac{\eta}{2}\ns{\nabla f(x_{t-1})}
                    - \big(\frac{1}{2\eta}- \frac{L}{2}\big)\ns{x_t-x_{t-1}}, \label{eq:difff}
\end{align}
where (\ref{eq:lp}) holds since $f$ has $L$-Lipschitz continuous gradient (Assumption \ref{asp:1}).
Now, we bound the variance term as follows, where we take expectations with the history:
\begin{align}
 &\E[\ns{v_{t-1}-\nabla f(x_{t-1})}] \notag\\
 &= \E\Big[\Big\|\frac{1}{b}\sum_{i\in I_b}\big(\nabla f_i(x_{t-1})-\nabla f_i(x_{t-2})\big)
            + v_{t-2}-\nabla f(x_{t-1})\Big\|^2\Big] \notag\\
 &= \E\Big[\Big\|\frac{1}{b}
     \sum_{i\in I_b}\Big(\big(\nabla f_i(x_{t-1})-\nabla f_i(x_{t-2})\big)
            -\big(\nabla f(x_{t-1})-\nabla f(x_{t-2})\big)\Big)
            +v_{t-2}-\nabla f(x_{t-2}) \Big\|^2\Big] \notag \\
 &= \E\Big[\Big\|\frac{1}{b}
     \sum_{i\in I_b}\Big(\big(\nabla f_i(x_{t-1})-\nabla f_i(x_{t-2})\big)
            -\big(\nabla f(x_{t-1})-\nabla f(x_{t-2})\big)\Big)\Big\|^2\Big]
            +\E[\| v_{t-2}-\nabla f(x_{t-2}) \|^2] \label{eq:ijind}\\
 &= \frac{1}{b^2}\E\Big[\sum_{i\in I_b}\Big\|
     \big(\nabla f_i(x_{t-1})-\nabla f_i(x_{t-2})\big)
            -\big(\nabla f(x_{t-1})-\nabla f(x_{t-2})\big)\Big\|^2\Big]
            +\E[\| v_{t-2}-\nabla f(x_{t-2}) \|^2] \label{eq:v1}\\
 &\leq \frac{1}{b^2}
         \E\Big[\sum_{i\in I_b}\Big\|\nabla f_i(x_{t-1})-\nabla f_i(x_{t-2})\Big\|^2\Big]
         +\E[\| v_{t-2}-\nabla f(x_{t-2}) \|^2] \label{eq:v2}\\
 &\leq \frac{L^2}{b}\E[\ns{x_{t-1}-x_{t-2}}]
        +\E[\| v_{t-2}-\nabla f(x_{t-2}) \|^2] \label{eq:useasp},
\end{align}
where \eqref{eq:ijind} and \eqref{eq:v1} use the law of total expectation and $\E[\ns{x_1+x_2+\cdots+x_k}]=\sum_{i=1}^k\E[\ns{x_i}]$ if $x_1, x_2, \ldots, x_k$ are independent and of mean zero, \eqref{eq:v2} uses the fact $\E[\ns{x-\E x}]\leq \E[\ns{x}]$, and \eqref{eq:useasp} holds due to the gradient Lipschitz Assumption \ref{asp:1}.

Note that for $\E[\| v_{t-2}-\nabla f(x_{t-2}) \|^2]$ in \eqref{eq:useasp}, we can reuse the same computation above. Thus we can sum up \eqref{eq:useasp} from the beginning of this epoch $sm$ to the point $t-1$,
\begin{align}
 \E[\ns{v_{t-1}-\nabla f(x_{t-1})}] &\leq \frac{L^2}{b}\sum_{j=sm+1}^{t-1}\E[\ns{x_{j}-x_{j-1}}]
        +\E[\| v_{sm}-\nabla f(x_{sm}) \|^2] \label{eq:var} \\
        &\leq \frac{L^2}{b}\sum_{j=sm+1}^{t-1}\E[\ns{x_{j}-x_{j-1}}] \label{eq:varfinite},
\end{align}
where \eqref{eq:varfinite} holds since we compute the full gradient at the beginning point of this epoch, i.e., $v_{sm}=\nabla f(x_{sm})$ (see Line \ref{line:full} of Algorithm \ref{alg:ssrgd_hl}).
Now, we take expectations for \eqref{eq:difff} and then sum it up from the beginning of this epoch $s$, i.e., iterations from $sm$ to $t$, by plugging the variance \eqref{eq:varfinite} into them to get:
\begin{align}
\E[f(x_{t})] &\leq \E[f(x_{sm})] - \frac{\eta}{2}\sum_{j=sm+1}^{t}\E[\ns{\nabla f(x_{j-1})}]
        - \big(\frac{1}{2\eta}- \frac{L}{2}\big)\sum_{j=sm+1}^{t}\E[\ns{x_j-x_{j-1}}] \notag\\
        &\qquad + \frac{\eta L^2}{2b}\sum_{k=sm+1}^{t-1}\sum_{j=sm+1}^{k}\E[\ns{x_{j}-x_{j-1}}] \notag\\
    &\leq \E[f(x_{sm})] - \frac{\eta}{2}\sum_{j=sm+1}^{t}\E[\ns{\nabla f(x_{j-1})}]
        - \big(\frac{1}{2\eta}- \frac{L}{2}\big)\sum_{j=sm+1}^{t}\E[\ns{x_j-x_{j-1}}] \notag\\
        &\qquad + \frac{\eta L^2(t-1-sm)}{2b}\sum_{j=sm+1}^{t}\E[\ns{x_{j}-x_{j-1}}] \notag\\
    &\leq \E[f(x_{sm})] - \frac{\eta}{2}\sum_{j=sm+1}^{t}\E[\ns{\nabla f(x_{j-1})}]
        - \big(\frac{1}{2\eta}- \frac{L}{2}\big)\sum_{j=sm+1}^{t}\E[\ns{x_j-x_{j-1}}] \notag\\
        &\qquad + \frac{\eta L^2}{2}\sum_{j=sm+1}^{t}\E[\ns{x_{j}-x_{j-1}}] \label{eq:bm}\\
    &\leq \E[f(x_{sm})] - \frac{\eta}{2}\sum_{j=sm+1}^{t}\E[\ns{\nabla f(x_{j-1})}] \label{eq:eta},
\end{align}
where \eqref{eq:bm} holds if the minibatch size $b\geq m$ (note that here $t\leq (s+1)m$), and
\eqref{eq:eta} holds if the step size $\eta\leq \frac{\sqrt{5}-1}{2L}$.

\begin{proofof}{Theorem \ref{thm:f1}}
Let $b=m=\sqrt{n}$ and step size $\eta\leq \frac{\sqrt{5}-1}{2L}$, then \eqref{eq:eta} holds.
Now, the proof is directly obtained by summing up \eqref{eq:eta} for all epochs $0\leq s\le S$ as follows:
\begin{align}
\E[f(x_{T})]
    &\leq \E[f(x_0)] - \frac{\eta}{2}\sum_{j=1}^{T}\E[\ns{\nabla f(x_{j-1})}] \notag\\
\E[\n{\nabla f(\hx)}]\leq  \sqrt{\E[\ns{\nabla f(\hx)}]} &\leq \sqrt{\frac{2(f(x_0)-f^*)}{\eta T}} = \epsilon \label{eq:finitefinal_main},
\end{align}
where \eqref{eq:finitefinal_main} holds by choosing $\hx$ uniformly from $\{x_{t-1}\}_{t\in[T]}$
and letting $Sm\leq T=\frac{2(f(x_0)-f^*)}{\eta\epsilon^2}=O(\frac{L(f(x_0)-f^*)}{\epsilon^2})$.
Note that the total number of computation of stochastic gradients equals to
\begin{align*}
  Sn+Smb\leq  \Big\lceil\frac{T}{m}\Big\rceil n + Tb \leq \Big(\frac{T}{\sqrt{n}}+1\Big)n+T\sqrt{n}
  =n+2T\sqrt{n}=O\Big(n+\frac{L(f(x_0)-f^*)\sqrt{n}}{\epsilon^2}\Big).
\end{align*}
\end{proofof}

\subsubsection{Proof of Theorem \ref{thm:f2}}
\label{app:prooff2}
For proving the second-order guarantee, we divide the proof into two situations.
The first situation (\textbf{large gradients}) is almost the same as the above arguments for first-order guarantee, where the function value will decrease a lot since the gradients are large (see \eqref{eq:eta}).
For the second situation (\textbf{around saddle points}), we will show that the function value can also decrease a lot by adding a random perturbation. The reason is that saddle points are usually unstable and the stuck region is relatively small in a random perturbation ball.

\vspace{1mm}
\noindent{{\bf Large Gradients}: }
First, we need a high probability bound for the variance term instead of the expectation one \eqref{eq:varfinite}. Then we use it to get a high probability bound of \eqref{eq:eta} for function value decrease.
Recall that $v_k=\frac{1}{b}\sum_{i\in I_b}\big(\nabla f_i(x_{k})-\nabla f_i(x_{k-1})\big) + v_{k-1}$ (see Line \ref{line:v} of Algorithm \ref{alg:ssrgd_hl}), we let $y_k:=v_k-\nabla f(x_k)$ and $z_k:=y_k-y_{k-1}$.
It is not hard to verify that $\{y_k\}$ is a martingale sequence and $\{z_k\}$ is the associated martingale difference sequence.
In order to apply the Azuma-Hoeffding inequalities to get a high probability bound, we first need to bound the difference sequence $\{z_k\}$.
We use the Bernstein inequality to bound the differences as follows.
\begin{align}
  z_k=y_k-y_{k-1}&= v_k-\nabla f(x_k) - (v_{k-1}-\nabla f(x_{k-1})) \notag\\
  &=\frac{1}{b}\sum_{i\in I_b}\big(\nabla f_i(x_{k})-\nabla f_i(x_{k-1})\big) + v_{k-1}
        -\nabla f(x_k) - (v_{k-1}-\nabla f(x_{k-1})) \notag \\
  &=\frac{1}{b}\sum_{i\in I_b}\Big(\nabla f_i(x_{k})-\nabla f_i(x_{k-1})
        -(\nabla f(x_k) -\nabla f(x_{k-1}))\Big). \label{eq:zk}
\end{align}
We define $u_i:=\nabla f_i(x_{k})-\nabla f_i(x_{k-1})-(\nabla f(x_k) -\nabla f(x_{k-1}))$, and then we have
\begin{align}
\|u_i\|=\|\nabla f_i(x_{k})-\nabla f_i(x_{k-1})-(\nabla f(x_k) -\nabla f(x_{k-1}))\|\leq 2\|x_{k}-x_{k-1}\|, \label{eq:b1}
\end{align}
where the last inequality holds due to the gradient Lipschitz Assumption \ref{asp:1}.
Then, consider the variance term $\sigma^2$
\begin{align}
  \sigma^2&=\sum_{i\in I_b}\E[\|u_i\|^2] \notag\\
          &=\sum_{i\in I_b}\E[\ns{\nabla f_i(x_{k})-\nabla f_i(x_{k-1})-(\nabla f(x_k) -\nabla f(x_{k-1}))}] \notag\\
          &\leq \sum_{i\in I_b}\E[\ns{\nabla f_i(x_{k})-\nabla f_i(x_{k-1})}] \notag\\
          &\leq bL^2\ns{x_{k}-x_{k-1}}, \label{eq:b2}
\end{align}
where the first inequality uses the fact $\E[\ns{x-\E x}]\leq \E[\ns{x}]$, and the last inequality uses the gradient Lipschitz Assumption \ref{asp:1}.
According to \eqref{eq:b1} and \eqref{eq:b2}, we can bound the difference $z_k$ by Bernstein inequality (Proposition \ref{prop:bernstein}) as
\begin{align*}
\pr\Big\{\big\|z_k\big\|\geq \frac{t}{b} \Big\} &\leq (d+1) \exp\Big(\frac{-t^2/2}{\sigma^2+Rt/3}\Big) \notag \\
    & = (d+1) \exp\Big(\frac{-t^2/2}{bL^2\ns{x_{k}-x_{k-1}}+ 2\|x_{k}-x_{k-1}\|t/3}\Big)
    \notag \\
    & = \zeta_k,
\end{align*}
where the last equality holds by letting $t=CL\sqrt{b}\n{x_{k}-x_{k-1}}$, where $C=O(\log\frac{d}{\zeta_k})=\tdo(1)$.
Now, we have a high probability bound for the difference sequence $\{z_k\}$, i.e.,
\begin{align}
    \|z_k\| \leq  \frac{CL\n{x_{k}-x_{k-1}}}{\sqrt{b}} \quad \mathrm{~with~probability~} 1-\zeta_k.
\end{align}

Now, we are ready to get a high probability bound for our original variance term \eqref{eq:varfinite} by using the martingale Azuma-Hoeffding inequality.
Consider in a specifical epoch $s$, i.e, iterations $t$ from $sm+1$ to current $sm+k$, where $k$ is less than $m$ (note that we only need to consider the current epoch since each epoch we start with $y=0$), we use a union bound for the difference sequence $\{z_t\}$ by letting $\zeta_k = \zeta/m$ such that
\begin{align}
\|z_t\|\leq c_t=\frac{CL\n{x_{t}-x_{t-1}}}{\sqrt{b}} \mathrm{~~for~ all~} sm+1 \leq t\leq sm+k \mathrm{~~with~ probability~~} 1-\zeta.
\end{align}
Then according to Azuma-Hoeffding inequality (Proposition \ref{prop:azumahigh}) and noting that $\zeta_k = \zeta/m$, we have
\begin{align*}
\pr\Big\{\big\|y_{sm+k}-y_{sm}\big\|\geq \beta \Big\} &\leq (d+1) \exp\Big(\frac{-\beta^2}{8\sum_{t=sm+1}^{sm+k} c_t^2}\Big)+\zeta \notag \\
    & = 2\zeta,
\end{align*}
where the last equality holds by letting $\beta=\sqrt{8\sum_{t=sm+1}^{sm+k} c_t^2\log\frac{d}{\zeta}}
=\frac{C'L\sqrt{\sum_{t=sm+1}^{sm+k}\ns{x_{t}-x_{t-1}}}}{\sqrt{b}}$, where $C'=O(C\sqrt{\log\frac{d}{\zeta}})=\tdo(1)$.
Recall that $y_k:=v_k-\nabla f(x_k)$ and at the beginning point of this epoch $y_{sm}=0$ due to $v_{sm}=\nabla f(x_{sm})$ (see Line \ref{line:full} of Algorithm \ref{alg:ssrgd_hl}), thus we have
\begin{align}\label{eq:highvar}
\n{v_{t-1}-\nabla f(x_{t-1})}=\n{y_{t-1}} \leq \frac{C'L\sqrt{\sum_{j=sm+1}^{t-1}\ns{x_{j}-x_{j-1}}}}{\sqrt{b}}
\end{align}
with probability $1-2\zeta$, where $t$ belongs to $[sm+1,(s+1)m]$.

Now, we use this high probability version \eqref{eq:highvar} instead of the expectation one \eqref{eq:varfinite} to obtain the high probability bound for function value decrease (see \eqref{eq:eta}).
We sum up \eqref{eq:difff} from the beginning of this epoch $s$, i.e., iterations from $sm$ to $t$, by plugging \eqref{eq:highvar} into them to get:
\begin{align}
f(x_{t}) &\leq f(x_{sm}) - \frac{\eta}{2}\sum_{j=sm+1}^{t}\ns{\nabla f(x_{j-1})}
        - \big(\frac{1}{2\eta}- \frac{L}{2}\big)\sum_{j=sm+1}^{t}\ns{x_j-x_{j-1}} \notag\\
        &\qquad + \frac{\eta}{2}\sum_{k=sm+1}^{t-1}\frac{C'^2L^2\sum_{j=sm+1}^{k}\ns{x_{j}-x_{j-1}}}{b} \notag\\
    &\leq f(x_{sm}) - \frac{\eta}{2}\sum_{j=sm+1}^{t}\ns{\nabla f(x_{j-1})}
        - \big(\frac{1}{2\eta}- \frac{L}{2}\big)\sum_{j=sm+1}^{t}\ns{x_j-x_{j-1}} \notag\\
        &\qquad + \frac{\eta C'^2L^2}{2b}\sum_{k=sm+1}^{t-1}\sum_{j=sm+1}^{k}\ns{x_{j}-x_{j-1}} \notag\\
    &\leq f(x_{sm}) - \frac{\eta}{2}\sum_{j=sm+1}^{t}\ns{\nabla f(x_{j-1})}
        - \big(\frac{1}{2\eta}- \frac{L}{2}\big)\sum_{j=sm+1}^{t}\ns{x_j-x_{j-1}} \notag\\
        &\qquad + \frac{\eta C'^2L^2(t-1-sm)}{2b}\sum_{j=sm+1}^{t}\ns{x_{j}-x_{j-1}} \notag\\
    &\leq f(x_{sm}) - \frac{\eta}{2}\sum_{j=sm+1}^{t}\ns{\nabla f(x_{j-1})}
        - \big(\frac{1}{2\eta}- \frac{L}{2}-\frac{\eta C'^2L^2}{2}\big)\sum_{j=sm+1}^{t}\ns{x_j-x_{j-1}}  \label{eq:bm2}\\
    &\leq f(x_{sm}) - \frac{\eta}{2}\sum_{j=sm+1}^{t}\ns{\nabla f(x_{j-1})} \label{eq:eta2},
\end{align}
where \eqref{eq:bm2} holds if the minibatch size $b\geq m$ (note that here $t\leq (s+1)m$), and
\eqref{eq:eta2} holds if the step size $\eta\leq \frac{\sqrt{4C'^2+1}-1}{2C'^2L}$.

Note that \eqref{eq:eta2} only guarantees function value decrease when the summation of gradients in this epoch is large. However, in order to connect the guarantees between first situation (large gradients) and second situation (around saddle points), we need to show guarantees that are related to the \emph{gradient of the starting point} of each epoch (see Line \ref{line:super} of Algorithm \ref{alg:ssrgd}).
Similar to \citep{ge2019stable}, we achieve this by stopping the epoch at a uniformly random point (see Line \ref{line:randomstop} of Algorithm \ref{alg:ssrgd}).

Now we recall Lemma \ref{lem:first} to connect these two situations (large gradients and around saddle points):

\begingroup
\def\thelemma{\ref{lem:first}}
\begin{lemma}[Connection of Two Situations]
\label{lem:first1}
For any epoch $s$, let $x_t$ be a point uniformly sampled from this epoch $\{x_{j} \}_{j=sm}^{(s+1)m}$ and choose the step size $\eta \leq \frac{\sqrt{4C'^2+1}-1}{2C'^2L}$ (where $C'=O(\log\frac{dn}{\zeta})=\tdo(1)$) and the minibatch size $b\geq m$.
Then for any $\mathG$, we have two cases:
\begin{enumerate}
\item If at least half of points in this epoch have gradient norm no larger than $\mathG$, then $\|\nabla f(x_t) \| \le \mathG$ holds with probability at least $1/2$;
\item Otherwise, we know $f(x_{sm}) - f(x_t) \ge \frac{\eta m\mathG^2}{8}$ holds with probability at least $1/5.$
\end{enumerate}
Moreover, $f(x_t) \le f(x_{sm})$ holds with high probability no matter which case happens.
\end{lemma}
\addtocounter{lemma}{-1}
\endgroup

\begin{proofof}{Lemma~\ref{lem:first}}
There are two cases in this epoch:
\begin{enumerate}
\item If at least half of points of in this epoch $\{x_{j} \}_{j=sm}^{(s+1)m}$ have gradient norm no larger than $\mathG$, then it is easy to see that a uniformly sampled point $x_t$ has gradient norm $\n{\nabla f(x_t)}\leq \mathG$ with probability at least $1/2.$
\item Otherwise, at least half of points have gradient norm larger than $\mathG$. Then, as long as the sampled point $x_t$ falls into the last quarter of $\{x_{j} \}_{j=sm}^{(s+1)m}$, we know $\sum_{j=sm+1}^{t}\ns{\nabla f(x_{j-1})}\geq \frac{m\mathG^2}{4}$. This holds with probability at least $1/4$ since $x_t$ is uniformly sampled. Then combining with \eqref{eq:eta2}, i.e.,  $f(x_{sm}) - f(x_t) \geq \frac{\eta}{2}\sum_{j=sm+1}^{t}\ns{\nabla f(x_{j-1})}$, we obtain the function value decrease $f(x_{sm}) - f(x_t) \ge \frac{\eta m\mathG^2}{8}$. Note that
    \eqref{eq:eta2} holds with high probability if we choose the minibatch size $b\geq m$ and the step size $\eta\leq \frac{\sqrt{4C'^2+1}-1}{2C'^2L}$.
    By a union bound, the function value decrease $f(x_{sm}) - f(x_t) \ge \frac{\eta m\mathG^2}{8}$ with probability at least $1/5$.
\end{enumerate}
Again according to \eqref{eq:eta2}, $f(x_t)\leq f(x_{sm})$ always holds with high probability.
\end{proofof}

Note that if Case 2 happens, the function value already decreases a lot in this epoch $s$ (corresponding to the first situation large gradients). Otherwise Case 1 happens, we know the starting point of the next epoch $x_{(s+1)m}=x_t$ (i.e., Line \ref{line:randompoint} of Algorithm \ref{alg:ssrgd}), then we know $\n{\nabla f(x_{(s+1)m})}=\n{\nabla f(x_{t})} \leq \mathG$. Then we will start a super epoch (corresponding to the second situation around saddle points). Note that if $\lambda_{\min}(\nabla^2 f(x_{(s+1)m}))> -\delta$, this point $x_{(s+1)m}$ is already an $(\epsilon,\delta)$-second-order stationary point (recall that $\mathG = \epsilon$ in our Theorem \ref{thm:f2}).

\vspace{3mm}
\noindent{{\bf Around Saddle Points} $\n{\nabla f(\tx)} \leq \mathG$ and $\lambda_{\min}(\nabla^2 f(\tx))\leq -\delta$: }
In this situation, we will show that the function value decreases a lot in a \emph{super epoch} (instead of an epoch as in the first situation) with high probability by adding a random perturbation at the initial point $\tx$. To simplify the presentation, we use $x_0:=\tx+\xi$ to denote the starting point of the super epoch after the perturbation, where $\xi$ uniformly $\sim \mathbb{B}_0(r)$ and the perturbation radius is $r$ (see Line \ref{line:init} in Algorithm \ref{alg:ssrgd}).
Following the classical widely used \emph{two-point analysis} developed in \citep{jin2017escape}, we consider two coupled points $x_0$ and $x_0'$ with $w_0:=x_0-x_0'=r_0e_1$, where $r_0$ is a scalar and $e_1$ denotes the smallest eigenvector direction of Hessian $\hess := \nabla^2 f(\tx)$. Then we get two coupled sequences $\{x_t\}$ and $\{x_t'\}$ by running SSRGD update steps (Line \ref{line:up1}--\ref{line:up2} of Algorithm \ref{alg:ssrgd}) with the same choice of minibatches (i.e., $I_b$'s in Line \ref{line:up2} of Algorithm \ref{alg:ssrgd}) for a super epoch.
We will show that at least one of these two coupled sequences will decrease the function value a lot (escape the saddle point), i.e.,
\begin{align}\label{eq:funcd}
 \exists t\leq\mathT, \mathrm{~~such ~that~~} \max\{f(x_0)-f(x_t), f(x_0')-f(x_t')\} \geq 2\mathf.
\end{align}
We will prove \eqref{eq:funcd} by contradiction. Assume the contrary, $f(x_0)-f(x_t)<2\mathf$ and $f(x_0')-f(x_t')<2\mathf$.
First, we show that if function value does not decrease a lot, then all iteration points are not far from the starting point with high probability.
Then we will show that the stuck region is relatively small in the random perturbation ball, i.e., at least one of $x_t$ and $x_t'$ will go far away from their starting point $x_0$ and $x_0'$ with high probability. Thus there is a contradiction.
We recall these two lemmas here and their proofs are deferred to the end of this section.
\begingroup
\def\thelemma{\ref{lem:local}}
\begin{lemma}[Localization]
Let $\{x_t\}$ denote the sequence by running SSRGD update steps (Line \ref{line:up1}--\ref{line:up2} of Algorithm \ref{alg:ssrgd}) from $x_0$.
Moreover, let the step size $\eta\leq \frac{1}{2C'L}$ and minibatch size $b\geq m$, with probability $1-\zeta$, we have
\begin{align}\label{eq:local}
\forall t,~~ \n{x_t-x_0}\leq \sqrt{\frac{4t(f(x_0)-f(x_t))}{C'L}},
\end{align}
where $C'=O(\log\frac{dt}{\zeta})=\tdo(1)$.
\end{lemma}
\addtocounter{lemma}{-1}
\endgroup

\begingroup
\def\thelemma{\ref{lem:smallstuck}}
\begin{lemma}[Small Stuck Region]
If the initial point $\tx$ satisfies $-\gamma:=\lambda_{\min}(\nabla^2 f(\tx))\leq -\delta$,
then let $\{x_t\}$ and $\{x_t'\}$ be two coupled sequences by running SSRGD update steps (Line \ref{line:up1}--\ref{line:up2} of Algorithm \ref{alg:ssrgd}) with the same choice of minibatches (i.e., $I_b$'s in Line \ref{line:up2}) from $x_0$ and $x_0'$ with $w_0:=x_0-x_0'=r_0e_1$, where $x_0\in\mathbb{B}_{\tx}(r)$, $x_0'\in\mathbb{B}_{\tx}(r)$ , $r_0=\frac{\zeta' r}{\sqrt{d}}$ and $e_1$ denotes the smallest eigenvector direction of Hessian $\nabla^2 f(\tx)$.
Moreover, let the super epoch length $\mathT=\frac{2\log(\frac{8\delta\sqrt{d}}{C_1\rho\zeta' r})}{\eta\delta}=\tdo(\frac{1}{\eta\delta})$, the step size $\eta\leq \min\big(\frac{1}{8\log(\frac{8\delta\sqrt{d}}{C_1\rho\zeta' r})L}, \frac{1}{4C_2L\log \mathT}\big)=\tdo(\frac{1}{L})$, minibatch size $b\geq m$ and
the perturbation radius $r\leq \frac{\delta}{C_1 \rho}$, then with probability $1-\zeta$, we have
\begin{align}\label{eq:stuck}
\exists T\leq \mathT,~~ \max\{\n{x_T-x_0}, \n{x_T'-x_0'}\}\geq \frac{\delta}{C_1\rho},
\end{align}
where $C_1\geq \frac{20C_2}{\eta L}$ and $C_2=O(\log\frac{d \mathT}{\zeta})=\tdo(1)$.
\end{lemma}
\addtocounter{lemma}{-1}
\endgroup

Based on these two lemmas, we are ready to show that \eqref{eq:funcd} holds with high probability. Without loss of generality, we assume $\n{x_T-x_0} \geq \frac{\delta}{C_1\rho}$ in \eqref{eq:stuck} (note that \eqref{eq:local} holds for both $\{x_t\}$ and $\{x_t'\}$), then plugging it into \eqref{eq:local} to obtain
\begin{align}
 \sqrt{\frac{4T(f(x_0)-f(x_T))}{C'L}}&\geq \frac{\delta}{C_1\rho} \notag\\
 f(x_0)-f(x_T) &\geq \frac{C'L\delta^2}{4C_1^2\rho^2T} \notag\\
            &\geq \frac{\eta C'L\delta^3}{8C_1^2\rho^2\log(\frac{8\delta\sqrt{d}}{C_1\rho\zeta' r})} \notag\\
            &=\frac{\delta^3}{C_1'\rho^2} \label{eq:largeeta}\\
            &=2\mathf,\notag
\end{align}
where the last inequality is due to $T\leq \mathT$ and \eqref{eq:largeeta} holds by letting
$C_1'=\frac{8C_1^2\log(\frac{8\delta\sqrt{d}}{C_1\rho\zeta' r})}{\eta C'L}$.
Thus, we already prove that at least one of sequences $\{x_t\}$ and $\{x_t'\}$
escapes the saddle point with high probability, i.e.,
\begin{align}
\exists T\leq \mathT~~, \max\{f(x_0)-f(x_T), f(x_0')-f(x_T')\} \geq 2\mathf,
\end{align}
if their starting points $x_0$ and $x_0'$ satisfying $w_0:=x_0-x_0'=r_0e_1$, where $r_0= \frac{\zeta' r}{\sqrt{d}}$ and $e_1$ denotes the smallest eigenvector direction of Hessian $\hess := \nabla^2 f(\tx)$.
Similar to the classical argument in \citep{jin2017escape}, we know that in the random perturbation ball, the stuck points can only be a short interval in the $e_1$ direction, i.e.,
at least one of two points in the $e_1$ direction will escape the saddle point if their distance is larger than $r_0=\frac{\zeta' r}{\sqrt{d}}$.
Thus, we know that the probability of the starting point $x_0=\tx+\xi$ (where $\xi$ uniformly $\sim \mathbb{B}_0(r)$) located in the stuck region is less than
\begin{align}\label{eq:goodx0}
  \frac{r_0V_{d-1}(r)}{V_d(r)}=
  \frac{r_0\Gamma(\frac{d}{2}+1)}{\sqrt{\pi}r\Gamma(\frac{d}{2}+\frac{1}{2})}
  \leq \frac{r_0}{\sqrt{\pi}r}\big(\frac{d}{2}+1\big)^{1/2}
  \leq \frac{r_0\sqrt{d}}{r}=\zeta',
\end{align}
where $V_d(r)$ denotes the volume of a Euclidean ball with radius $r$ in $d$ dimension,
and the first inequality holds due to Gautschi's inequality.
By a union bound for \eqref{eq:goodx0} and \eqref{eq:largeeta} (holds with high probability if $x_0$ is not in a stuck region), we know
\begin{align}\label{eq:escape0}
f(x_0)-f(x_T) \geq 2\mathf=\frac{\delta^3}{C_1'\rho^2}
\end{align}
with high probability.
Note that the initial point of this super epoch is $\tx$ before the perturbation (see Line \ref{line:init} of Algorithm \ref{alg:ssrgd}), thus we need to show that the perturbation step $x_0=\tx+\xi$ (where $\xi$ uniformly $\sim \mathbb{B}_0(r)$) does not increase the function value a lot, i.e.,
\begin{align}
f(x_0)&\leq f(\tx) +\inner{\nabla f(\tx)}{x_0-\tx}
                    + \frac{L}{2}\ns{x_0-\tx}  \notag \\
        &\leq f(\tx) +\n{\nabla f(\tx)}\n{x_0-\tx}
                    + \frac{L}{2}\ns{x_0-\tx}  \notag\\
        &\leq f(\tx) +\mathG \cdot r +\frac{L}{2}r^2 \notag\\
        &\leq f(\tx) + \frac{\delta^3}{2C_1'\rho^2} \notag\\
        &= f(\tx) +\mathf, \label{eq:perturbless}
\end{align}
where the last inequality holds by letting the perturbation radius $r\leq \min\{\frac{\delta^3}{4C_1'\rho^2\mathG}, \sqrt{\frac{\delta^3}{2C_1'\rho^2L}}\}$.

Now we combine with \eqref{eq:escape0} and \eqref{eq:perturbless} to obtain with high probability
\begin{align}\label{eq:escapehigh}
f(\tx)-f(x_T)=f(\tx)-f(x_0)+f(x_0)-f(x_T) \geq -\mathf+2\mathf=\frac{\delta^3}{2C_1'\rho^2}.
\end{align}

Thus we have finished the proof for the second situation (around saddle points), i.e., we show that the function value decrease a lot ($\mathf=\frac{\delta^3}{2C_1'\rho^2}$) in a \emph{super epoch} (recall that $T\leq \mathT=\frac{2\log(\frac{8\delta\sqrt{d}}{C_1\rho\zeta' r})}{\eta\delta}$) by adding a random perturbation $\xi \sim \mathbb{B}_0(r)$ at the initial point $\tx$.

\vspace{3mm}
\noindent{{\bf Combing these two situations (large gradients and around saddle points) to prove Theorem \ref{thm:f2}:}}
First, we recall Theorem \ref{thm:f2} here since we want to recall the parameter setting.
\begingroup
\def\thetheorem{\ref{thm:f2}}
\begin{theorem}
Under Assumption \ref{asp:1} and \ref{asp:2} (i.e. \eqref{smoothg1} and \eqref{smoothh1}), let $\Delta f:= f(x_0)-f^*$, where $x_0$ is the initial point and $f^*$ is the optimal value of $f$. By letting step size $\eta=\tdo(\frac{1}{L})$, epoch length $m=\sqrt{n}$, minibatch size $b=\sqrt{n}$, perturbation radius $r=\tdo\big(\min(\frac{\delta^3}{\rho^2\epsilon}, \frac{\delta^{3/2}}{\rho\sqrt{L}})\big)$, threshold gradient $\mathG=\epsilon$, threshold function value $\mathf=\tdo(\frac{\delta^3}{\rho^2})$ and super epoch length $\mathT=\tdo(\frac{1}{\eta\delta})$, SSRGD will at least once get to an $(\epsilon,\delta)$-second-order stationary point with high probability using
\begin{equation*}
  \tdo\Big(\frac{L\Delta f\sqrt{n}}{\epsilon^2}
+\frac{L\rho^2\Delta f\sqrt{n}}{\delta^4}
+ \frac{\rho^2\Delta fn}{\delta^3}\Big)
\end{equation*}
stochastic gradients for nonconvex finite-sum problem \eqref{eq:finite}.
\end{theorem}
\addtocounter{theorem}{-1}
\endgroup
\begin{proofof}{Theorem \ref{thm:f2}}
Now, we prove this theorem by distinguishing the epochs into three types as follows:
\begin{enumerate}
  \item \emph{Type-1 useful epoch}: If at least half of points in this epoch have gradient norm larger than $\mathG$ (Case 2 of Lemma \ref{lem:first1});
  \item \emph{Wasted epoch}: If at least half of points in this epoch have gradient norm no larger than $\mathG$ and the starting point of the next epoch has gradient norm larger than $\mathG$ (it means that this epoch does not guarantee decreasing the function value a lot as the large gradients situation, also it cannot connect to the second super epoch situation since the starting point of the next epoch has gradient norm larger than $\mathG$);
  \item \emph{Type-2 useful super epoch}: If at least half of points in this epoch have gradient norm no larger than $\mathG$ and the starting point of the next epoch (here we denote this point as $\tx$) has gradient norm no larger than $\mathG$ (i.e., $\n{\nabla f(\tx)}\leq \mathG$) (Case 1 of Lemma \ref{lem:first}), according to Line \ref{line:super} of Algorithm \ref{alg:ssrgd}, we will start a super epoch. So here we denote this epoch along with its following super epoch as a type-2 useful super epoch.
\end{enumerate}
First, it is easy to see that the probability of a wasted epoch happened is less than $1/2$ due to the random stop (see Case 1 of Lemma \ref{lem:first} and Line \ref{line:randomstop} of Algorithm \ref{alg:ssrgd}) and different wasted epoch are independent.
Thus, with high probability, there are at most $\tdo(1)$ wasted epochs happened before a type-1 useful epoch or type-2 useful super epoch.
Now, we use $N_1$ and $N_2$ to denote the number of type-1 useful epochs and type-2 useful super epochs that the algorithm is needed. Recall that $\Delta f:= f(x_0)-f^*$, where $x_0$ is the initial point and $f^*$ is the optimal value of $f$. Also recall that the function value always does not increase with high probability (see Lemma \ref{lem:first}).

For type-1 useful epoch, according to Case 2 of Lemma \ref{lem:first}, we know that the function value decreases at least $\frac{\eta m\mathG^2}{8}$ with probability at least $1/5$.
Using a standard concentration, we know that with high probability $N_1$ type-1 useful epochs will decrease the function value at least $\frac{\eta m\mathG^2N_1}{80}$, note that the function value can decrease at most $\Delta f$.
So $\frac{\eta m\mathG^2N_1}{80}\leq \Delta f$, we get $N_1\leq \frac{80\Delta f}{\eta m\mathG^2}$.

For type-2 useful super epoch, first we know that the starting point of the super epoch $\tx$ has gradient norm $\n{\nabla f(\tx)}\leq \mathG$. Now if $\lambda_{\min}(\nabla^2 f(\tx))\geq -\delta$, then $\tx$ is already a $(\epsilon,\delta)$-second-order stationary point. Otherwise,
$\n{\nabla f(\tx)}\leq \mathG$ and $\lambda_{\min}(\nabla^2 f(\tx)) \leq -\delta$, this is exactly our second situation (around saddle points).
According to \eqref{eq:escapehigh}, we know that the the function value decrease ($f(\tx)-f(x_T)$) is at least $\mathf=\frac{\delta^3}{2C_1'\rho^2}$ with high probability.
Similar to type-1 useful epoch, we know $N_2\leq \frac{C_1''\rho^2\Delta f}{\delta^3}$ by a union bound (so we change $C_1'$ to $C_1''$, anyway we also have $C_1''=\tdo(1)$).

Now, we are ready to compute the convergence results to finish the proof for Theorem \ref{thm:f2}.
\begin{align}
&N_1(\tdo(1)n+n+mb) +N_2(\tdo(1)n+\big\lceil\frac{\mathT}{m}\big\rceil n+\mathT b) \notag\\
&\leq \tdo\Big(\frac{\Delta fn}{\eta m\mathG^2}+\frac{\rho^2\Delta f}{\delta^3}(n+\frac{\sqrt{n}}{\eta \delta})\Big) \notag\\
& \leq \tdo\Big(\frac{L\Delta f\sqrt{n}}{\epsilon^2}
+\frac{L\rho^2\Delta f\sqrt{n}}{\delta^4}
+ \frac{\rho^2\Delta fn}{\delta^3}\Big)
\end{align}
\end{proofof}

Now, the only remaining thing is to prove Lemma \ref{lem:local} and \ref{lem:smallstuck}.
We provide these two proofs as follows.

\begingroup
\def\thelemma{\ref{lem:local}}
\begin{lemma}[Localization]
Let $\{x_t\}$ denote the sequence by running SSRGD update steps (Line \ref{line:up1}--\ref{line:up2} of Algorithm \ref{alg:ssrgd}) from $x_0$.
Moreover, let the step size $\eta\leq \frac{1}{2C'L}$ and minibatch size $b\geq m$, with probability $1-\zeta$, we have
\begin{align*}
\forall t,~~ \n{x_t-x_0}\leq \sqrt{\frac{4t(f(x_0)-f(x_t))}{C'L}},
\end{align*}
where $C'=O(\log\frac{dt}{\zeta})=\tdo(1)$.
\end{lemma}
\addtocounter{lemma}{-1}
\endgroup

\begin{proofof}{Lemma~\ref{lem:local}}
First, we assume the variance bound \eqref{eq:highvar} holds for all $0\leq j\leq t-1$ (this is true with high probability using a union bound by letting $C'=O(\log\frac{dt}{\zeta})$).
Then, according to \eqref{eq:bm2}, we know for any $\tau \leq t$ in some epoch $s$
\begin{align}
f(x_{\tau})
    &\leq f(x_{sm}) - \frac{\eta}{2}\sum_{j=sm+1}^{\tau}\ns{\nabla f(x_{j-1})}
        - \big(\frac{1}{2\eta}- \frac{L}{2}-\frac{\eta C'^2L^2}{2}\big)\sum_{j=sm+1}^{\tau}\ns{x_j-x_{j-1}}  \notag\\
    &\leq f(x_{sm}) - \big(\frac{1}{2\eta}- \frac{L}{2}-\frac{\eta C'^2L^2}{2}\big)\sum_{j=sm+1}^{\tau}\ns{x_j-x_{j-1}} \notag\\
    &\leq f(x_{sm}) - \frac{C'L}{4}\sum_{j=sm+1}^{\tau}\ns{x_j-x_{j-1}}, \label{eq:dist}
\end{align}
where the last inequality holds since the step size $\eta\leq \frac{1}{2C'L}$ and assuming $C'\geq 1$.
Now, we sum up \eqref{eq:dist} for all epochs before iteration $t$,
\begin{align*}
f(x_{t})
    &\leq f(x_{0}) - \frac{C'L}{4}\sum_{j=1}^{t}\ns{x_j-x_{j-1}}.
\end{align*}
Then, the proof is finished as
\begin{align*}
   \n{x_t-x_0}\leq \sum_{j=1}^t\n{x_j-x_{j-1}}\leq \sqrt{t\sum_{j=1}^{t}\ns{x_j-x_{j-1}}}
    \leq \sqrt{\frac{4t(f(x_0)-f(x_t))}{C'L}}.
\end{align*}
\end{proofof}

\begingroup
\def\thelemma{\ref{lem:smallstuck}}
\begin{lemma}[Small Stuck Region]
If the initial point $\tx$ satisfies $-\gamma:=\lambda_{\min}(\nabla^2 f(\tx))\leq -\delta$,
then let $\{x_t\}$ and $\{x_t'\}$ be two coupled sequences by running SSRGD update steps (Line \ref{line:up1}--\ref{line:up2} of Algorithm \ref{alg:ssrgd}) with the same choice of minibatches (i.e., $I_b$'s in Line \ref{line:up2}) from $x_0$ and $x_0'$ with $w_0:=x_0-x_0'=r_0e_1$, where $x_0\in\mathbb{B}_{\tx}(r)$, $x_0'\in\mathbb{B}_{\tx}(r)$ , $r_0=\frac{\zeta' r}{\sqrt{d}}$ and $e_1$ denotes the smallest eigenvector direction of Hessian $\nabla^2 f(\tx)$.
Moreover, let the super epoch length $\mathT=\frac{2\log(\frac{8\delta\sqrt{d}}{C_1\rho\zeta' r})}{\eta\delta}=\tdo(\frac{1}{\eta\delta})$, the step size $\eta\leq \min\big(\frac{1}{8\log(\frac{8\delta\sqrt{d}}{C_1\rho\zeta' r})L}, \frac{1}{4C_2L\log \mathT}\big)=\tdo(\frac{1}{L})$, minibatch size $b\geq m$ and
the perturbation radius $r\leq \frac{\delta}{C_1 \rho}$, then with probability $1-\zeta$, we have
\begin{align*}
\exists T\leq \mathT,~~ \max\{\n{x_T-x_0}, \n{x_T'-x_0'}\}\geq \frac{\delta}{C_1\rho},
\end{align*}
where $C_1\geq \frac{20C_2}{\eta L}$ and $C_2=O(\log\frac{d \mathT}{\zeta})=\tdo(1)$.
\end{lemma}
\addtocounter{lemma}{-1}
\endgroup

\begin{proofof}{Lemma~\ref{lem:smallstuck}}
We prove this lemma by contradiction. Assume the contrary,
\begin{align}\label{eq:distbound}
\forall t\leq \mathT~~, \n{x_t-x_0} \leq \frac{\delta}{C_1\rho} \mathrm{~~and~~} \n{x_t'-x_0'} \leq \frac{\delta}{C_1\rho}
\end{align}
We will show that the distance between these two coupled sequences $w_t:=x_t-x_t'$ will grow exponentially since they have a gap in the $e_1$ direction at the beginning, i.e., $w_0:=x_0-x_0'=r_0e_1$, where $r_0=\frac{\zeta' r}{\sqrt{d}}$ and $e_1$ denotes the smallest eigenvector direction of Hessian $\hess := \nabla^2 f(\tx)$.
However, $\n{w_t}=\n{x_t-x_t'}\leq \n{x_t-x_0}+\n{x_0-\tx}+\n{x_t'-x_0'}+\n{x_0'-\tx}\leq 2r+2\frac{\delta}{C_1\rho}$ according to \eqref{eq:distbound} and the perturbation radius $r$.
It is not hard to see that the exponential increase will break this upper bound, thus we get a contradiction.

In the following, we prove the exponential increase of $w_t$ by induction.
First, we need the expression of $w_t$ (recall that $x_t=x_{t-1}-\eta v_{t-1}$ (see Line \ref{line:update} of Algorithm \ref{alg:ssrgd})):
\begin{align}
w_t&=w_{t-1}-\eta(v_{t-1}-v_{t-1}') \notag\\
&=w_{t-1}-\eta\big(\nabla f(x_{t-1})- \nabla f(x_{t-1}')
        +v_{t-1}-\nabla f(x_{t-1})-v_{t-1}'+\nabla f(x_{t-1}')\big) \notag\\
&=w_{t-1}-\eta\Big(\int_0^1\nabla^2 f(x_{t-1}' + \theta(x_{t-1}-x_{t-1}'))d\theta(x_{t-1}-x_{t-1}')
        +v_{t-1}-\nabla f(x_{t-1})-v_{t-1}'+\nabla f(x_{t-1}')\Big) \notag\\
&=w_{t-1}-\eta\Big((\hess +  \Delta_{t-1})w_{t-1}
        +v_{t-1}-\nabla f(x_{t-1})-v_{t-1}'+\nabla f(x_{t-1}')\Big) \notag\\
&=(I-\eta \hess)w_{t-1}-\eta(\Delta_{t-1}w_{t-1}+y_{t-1}) \notag\\
&=(I-\eta \hess)^{t}w_0-\eta\sum_{\tau=0}^{t-1}(I-\eta \hess)^{t-1-\tau}(\Delta_\tau w_\tau+y_\tau) \label{eq:expw}
\end{align}
where $\Delta_{\tau} :=\int_0^1(\nabla^2 f(x_\tau'+\theta(x_\tau-x_\tau'))-\hess)d\theta$
and $y_\tau :=v_{\tau}-\nabla f(x_{\tau})-v_{\tau}'+\nabla f(x_{\tau}')$.
Note that the first term of \eqref{eq:expw} is in the $e_1$ direction and is exponential with respect to $t$, i.e., $(1+\eta\gamma)^t r_0 e_1$, where $-\gamma:=\lambda_{\min}(\hess)=\lambda_{\min}(\nabla^2 f(\tx))\leq -\delta$.
To prove the exponential increase of $w_t$, it is sufficient to show that the first term of \eqref{eq:expw} will dominate the second term.
We inductively prove the following two bounds
\begin{enumerate}
  \item $\frac{1}{2}(\base)^t r_0\leq\n{w_t}\leq\frac{3}{2}(\base)^t r_0$
  \item $\n{y_t}\leq \eta\gamma L(\base)^t r_0$
\end{enumerate}
First, check the base case $t=0$, $\n{w_0}=\n{r_0 e_1}=r_0$ and
$\n{y_0}= \n{v_{0}-\nabla f(x_0)-v_{0}'+\nabla f(x_{0}')}=\n{\nabla f(x_0)-\nabla f(x_0)-\nabla f(x_{0}')+\nabla f(x_{0}')}=0$. Assume they hold for all $\tau\leq t-1$, we now prove they hold for $t$ one by one.
For Bound 1, it is enough to show the second term of \eqref{eq:expw} is dominated by half of the first term.
\begin{align}
\n{\eta\sum_{\tau=0}^{t-1}(I-\eta \hess)^{t-1-\tau}(\Delta_\tau w_\tau)}
&\leq \eta\sum_{\tau=0}^{t-1}(\base)^{t-1-\tau}\n{\Delta_\tau}\n{w_\tau} \notag\\
&\leq \frac{3}{2}\eta(\base)^{t-1}r_0\sum_{\tau=0}^{t-1}\n{\Delta_\tau} \label{eq:0}\\
&\leq \frac{3}{2}\eta(\base)^{t-1}r_0\sum_{\tau=0}^{t-1}\rho D_\tau^x \label{eq:1}\\
&\leq \frac{3}{2}\eta(\base)^{t-1}r_0t\rho \big(\Dtop\big)\label{eq:2}\\
&\leq \frac{3}{C_1}\eta\delta t(\base)^{t-1}r_0\label{eq:3}\\
&\leq \frac{6\log(\frac{8\delta\sqrt{d}}{C_1\rho\zeta' r})}{C_1}(\base)^{t-1}r_0 \label{eq:4}\\
&\leq \frac{1}{4}(\base)^{t}r_0, \label{eq:5}
\end{align}
where \eqref{eq:0} uses the induction for $w_\tau$ with $\tau\leq t-1$,
\eqref{eq:1} uses the definition $D_\tau^x:=\max\{\n{x_\tau-\tx},\n{x_\tau'-\tx}\}$, \eqref{eq:2} follows from $\n{x_t-\tx}\leq\n{x_t-x_0}+\n{x_0-\tx}=\Dtop$ due to  \eqref{eq:distbound} and the perturbation radius $r$,
\eqref{eq:3} holds by letting the perturbation radius $r\leq \frac{\delta}{C_1\rho}$,
\eqref{eq:4} holds since $t\leq\mathT=\frac{2\log(\frac{8\delta\sqrt{d}}{C_1\rho\zeta' r})}{\eta\delta}$,
and \eqref{eq:5} holds by letting $C_1\geq 24\log(\frac{8\delta\sqrt{d}}{\rho\zeta' r})$.

\begin{align}
\n{\eta\sum_{\tau=0}^{t-1}(I-\eta \hess)^{t-1-\tau}y_\tau}
&\leq \eta\sum_{\tau=0}^{t-1}(\base)^{t-1-\tau}\n{y_\tau} \notag\\
&\leq \eta\sum_{\tau=0}^{t-1}(\base)^{t-1-\tau} \eta\gamma L(\base)^\tau r_0 \label{eq:10}\\
&=\eta \eta\gamma Lt(\base)^{t-1} r_0 \notag\\
&\leq \eta \eta\gamma L \frac{2\log(\frac{8\delta\sqrt{d}}{C_1\rho\zeta' r})}{\eta\delta} (\base)^{t-1} r_0 \label{eq:11}\\
&\leq 2\eta \log(\frac{8\delta\sqrt{d}}{C_1\rho\zeta' r})L(\base)^{t-1} r_0 \label{eq:12}\\
&\leq \frac{1}{4}(\base)^{t}r_0, \label{eq:13}
\end{align}
where \eqref{eq:10} uses the induction for $y_\tau$ with $\tau\leq t-1$,
\eqref{eq:11} holds since $t\leq\mathT=\frac{2\log(\frac{8\delta\sqrt{d}}{C_1\rho\zeta' r})}{\eta\delta}$,
\eqref{eq:12} holds $\gamma\geq \delta$ (recall $-\gamma:=\lambda_{\min}(\hess)=\lambda_{\min}(\nabla^2 f(\tx))\leq -\delta$),
and \eqref{eq:13} holds by letting $\eta\leq \frac{1}{8 \log(\frac{8\delta\sqrt{d}}{C_1\rho\zeta' r})L}$.

Combining \eqref{eq:5} and \eqref{eq:13}, we proved the second term of \eqref{eq:expw} is dominated by half of the first term.
Note that the first term of \eqref{eq:expw} is $\n{(I-\eta \hess)^{t}w_0}=(1+\eta\gamma)^t r_0$. Thus, we have
\begin{align}\label{eq:wt}
\frac{1}{2}(\base)^t r_0\leq\n{w_t}\leq\frac{3}{2}(\base)^t r_0
\end{align}

Now, the remaining thing is to prove the second bound $\n{y_t}\leq \eta\gamma L(\base)^t r_0$. First, we write the concrete expression of $y_t$:
\begin{align}
y_t &= v_{t}-\nabla f(x_{t})-v_{t}'+\nabla f(x_{t}') \notag \\
    &= \frac{1}{b}\sum_{i\in I_b}\big(\nabla f_i(x_{t})-\nabla f_i(x_{t-1})\big)
        + v_{t-1}-\nabla f(x_{t})\notag\\
        &\qquad
        -\frac{1}{b}\sum_{i\in I_b}\big(\nabla f_i(x_{t}')-\nabla f_i(x_{t-1}')\big)
        - v_{t-1}'+\nabla f(x_{t}')  \label{eq:30} \\
    &= \frac{1}{b}\sum_{i\in I_b}\big(\nabla f_i(x_{t})-\nabla f_i(x_{t-1})\big)
        + \nabla f(x_{t-1})-\nabla f(x_{t})  \notag\\
        &\qquad -\frac{1}{b}\sum_{i\in I_b}\big(\nabla f_i(x_{t}')-\nabla f_i(x_{t-1}')\big)
        - \nabla f(x_{t-1}') +\nabla f(x_{t}') \notag\\
        &\qquad
        +v_{t-1}- \nabla f(x_{t-1})
        -v_{t-1}' + \nabla f(x_{t-1}') \notag\\
    &= \frac{1}{b}\sum_{i\in I_b}\big(\nabla f_i(x_{t})-\nabla f_i(x_{t}')
        -\nabla f_i(x_{t-1})+\nabla f_i(x_{t-1}')\big)\notag\\
        &\qquad
        -\big(\nabla f(x_{t})-\nabla f(x_{t}') - \nabla f(x_{t-1}) + \nabla f(x_{t-1}')\big)
        + y_{t-1}, \notag
\end{align}
where \eqref{eq:30} is due to the definition of the estimator $v_t$ (see Line \ref{line:up2} of Algorithm \ref{alg:ssrgd}).
We further define the difference $z_t:=y_t-y_{t-1}$.
It is not hard to verify that $\{y_t\}$ is a martingale sequence and $\{z_t\}$ is the associated martingale difference sequence.
We will apply the Azuma-Hoeffding inequalities to get an upper bound for $\n{y_t}$
and then we prove $\n{y_t}\leq \eta\gamma L(\base)^t r_0$ based on that upper bound.
In order to apply the Azuma-Hoeffding inequalities for martingale sequence $\n{y_t}$, we first need to bound the difference sequence $\{z_t\}$.
We use the Bernstein inequality to bound the differences as follows.
\begin{align}
  z_t=y_t-y_{t-1}& = \frac{1}{b}\sum_{i\in I_b}\big(\nabla f_i(x_{t})-\nabla f_i(x_{t}')
        -\nabla f_i(x_{t-1})+\nabla f_i(x_{t-1}')\big)\notag\\
        &\qquad
        -\big(\nabla f(x_{t})-\nabla f(x_{t}')
        - \nabla f(x_{t-1}) + \nabla f(x_{t-1}')\big)\notag\\
&=\frac{1}{b}\sum_{i\in I_b}\Big(\big(\nabla f_i(x_{t})-\nabla f_i(x_{t}')\big)
        -\big(\nabla f_i(x_{t-1})-\nabla f_i(x_{t-1}')\big)\notag\\
        &\qquad\qquad\quad
        -\big(\nabla f(x_{t})-\nabla f(x_{t}') \big)
        + \big(\nabla f(x_{t-1}) - \nabla f(x_{t-1}')\big)\Big). \label{eq:zk2}
\end{align}
We define $u_i:=\big(\nabla f_i(x_{t})-\nabla f_i(x_{t}')\big)
        -\big(\nabla f_i(x_{t-1})-\nabla f_i(x_{t-1}')\big)
        -\big(\nabla f(x_{t})-\nabla f(x_{t}') \big)
        + \big(\nabla f(x_{t-1}) - \nabla f(x_{t-1}')\big)$, and then we have
\begin{align}
\|u_i\|&=\|\big(\nabla f_i(x_{t})-\nabla f_i(x_{t}')\big)
        -\big(\nabla f_i(x_{t-1})-\nabla f_i(x_{t-1}')\big)
        -\big(\nabla f(x_{t})-\nabla f(x_{t}') \big)
        + \big(\nabla f(x_{t-1}) - \nabla f(x_{t-1}')\big)\| \notag\\
   &\leq \Big\|\int_0^1\nabla^2 f_i(x_{t}'
                      + \theta(x_{t}-x_{t}'))d\theta(x_{t}-x_{t}')
                    -\int_0^1\nabla^2 f_i(x_{t-1}'
                      + \theta(x_{t-1}-x_{t-1}'))d\theta(x_{t-1}-x_{t-1}')\notag\\
        &\qquad
                -\int_0^1\nabla^2 f(x_{t}'
                    +\theta(x_{t}-x_{t}'))d\theta(x_{t}-x_{t}')
                +\int_0^1\nabla^2 f(x_{t-1}'
                    +\theta(x_{t-1}-x_{t-1}'))d\theta(x_{t-1}-x_{t-1}')\Big\| \notag\\
   &=\n{\hess_iw_t+\Delta_{t}^i w_t - (\hess_iw_{t-1}+\Delta_{t-1}^i w_{t-1})
                -(\hess w_t+\Delta_{t} w_t)+(\hess w_{t-1}+\Delta_{t-1} w_{t-1})}    \label{eq:40} \\
    &\leq \n{(\hess_i-\hess)(w_t-w_{t-1})}
            +\n{(\Delta_{t}^i -\Delta_{t}) w_t-(\Delta_{t-1}^i-\Delta_{t-1}) w_{t-1}} \notag\\
   &\leq 2L\|w_t-w_{t-1}\|+2\rho D_t^x\n{w_t}+2\rho D_{t-1}^x\n{w_{t-1}}, \label{eq:b12}
\end{align}
where \eqref{eq:40} holds since we define $\Delta_{t}:=\int_0^1(\nabla^2 f(x_t'+\theta(x_t-x_t'))-\hess)d\theta$ and $\Delta_{t}^i:=\int_0^1(\nabla^2 f_i(x_t'+\theta(x_t-x_t'))-\hess_i)d\theta$,
and the last inequality holds due to the gradient Lipschitz Assumption \ref{asp:1} and Hessian Lipschitz Assumption \ref{asp:2} (recall $D_t^x:=\max\{\n{x_t-\tx},\n{x_t'-\tx}\}$).
Then, consider the variance term $\sigma^2$
\begin{align}
  \sigma^2&=\sum_{i\in I_b}\E[\|u_i\|^2] \notag\\
          &\leq\sum_{i\in I_b}\E[\ns{\big(\nabla f_i(x_{t})-\nabla f_i(x_{t}')\big)
                    -\big(\nabla f_i(x_{t-1})-\nabla f_i(x_{t-1}')\big)}] \notag\\
          &=\sum_{i\in I_b}\E[\ns{\hess_iw_t+\Delta_{t}^i w_t
                                - (\hess_iw_{t-1}+\Delta_{t-1}^i w_{t-1})}] \notag\\
          &\leq b(L\|w_t-w_{t-1}\|+\rho D_t^x\n{w_t}+\rho D_{t-1}^x\n{w_{t-1}})^2, \label{eq:b22}
\end{align}
where the first inequality uses the fact $\E[\ns{x-\E x}]\leq \E[\ns{x}]$, and the last inequality uses the gradient Lipschitz Assumption \ref{asp:1} and Hessian Lipschitz Assumption \ref{asp:2}.
According to \eqref{eq:b12} and \eqref{eq:b22}, we can bound the difference $z_k$ by Bernstein inequality (Proposition \ref{prop:bernstein}) as (where $R=2L\|w_t-w_{t-1}\|+2\rho D_t^x\n{w_t}+2\rho D_{t-1}^x\n{w_{t-1}}$ and $\sigma^2=b(L\|w_t-w_{t-1}\|+\rho D_t^x\n{w_t}+\rho D_{t-1}^x\n{w_{t-1}})^2$)
\begin{align*}
\pr\Big\{\big\|z_t\big\|\geq \frac{\alpha}{b} \Big\} &\leq (d+1) \exp\Big(\frac{-\alpha^2/2}{\sigma^2+R\alpha/3}\Big)= \zeta_k,\\
\end{align*}
where the last equality holds by letting $\alpha=C_4\sqrt{b}(L\|w_t-w_{t-1}\|+\rho D_t^x\n{w_t}+\rho D_{t-1}^x\n{w_{t-1}})$, where $C_4=O(\log\frac{d}{\zeta_k})=\tdo(1)$.

Now, we have a high probability bound for the difference sequence $\{z_k\}$, i.e.,
\begin{align}
    \|z_k\| \leq c_k= \frac{C_4(L\|w_t-w_{t-1}\|+\rho D_t^x\n{w_t}+\rho D_{t-1}^x\n{w_{t-1}})}{\sqrt{b}} \quad \mathrm{~with~probability~} 1-\zeta_k.
\end{align}

Now, we are ready to get an upper bound for $y_t$ by using the martingale Azuma-Hoeffding inequality.
Note that we only need to consider the current epoch that contains the iteration $t$ since each epoch we start with $y=0$.
Let $s$ denote the current epoch, i.e, iterations from $sm+1$ to current $t$, where $t$ is no larger than $(s+1)m$.
According to Azuma-Hoeffding inequality (Proposition \ref{prop:azumahigh}) and letting $\zeta_k = \zeta/m$, we have
\begin{align*}
\pr\Big\{\big\|y_{t}-y_{sm}\big\|\geq \beta \Big\} &\leq (d+1) \exp\Big(\frac{-\beta^2}{8\sum_{k=sm+1}^{t} c_k^2}\Big)+\zeta \notag \\
    & = 2\zeta,
\end{align*}
where the last equality is due to $\beta=\sqrt{8\sum_{k=sm+1}^{t} c_k^2\log\frac{d}{\zeta}}
=\frac{C_3\sqrt{\sum_{k=sm+1}^{t}(L\|w_t-w_{t-1}\|+\rho D_t^x\n{w_t}+\rho D_{t-1}^x\n{w_{t-1}})^2}}{\sqrt{b}}$, where $C_3=O(C_4\sqrt{\log\frac{d}{\zeta}})=\tdo(1)$.
Recall that $y_k:=v_{k}-\nabla f(x_{k})-v_{k}'+\nabla f(x_{k}')$ and at the beginning point of this epoch $y_{sm}=0$ due to $v_{sm}=\nabla f(x_{sm})$ and $v_{sm}'=\nabla f(x_{sm}')$ (see Line \ref{line:full} of Algorithm \ref{alg:ssrgd_hl}), thus we have
\begin{align}\label{eq:highvar2x}
\n{y_{t}}=\n{y_t-y_{sm}} \leq \frac{C_3\sqrt{\sum_{k=sm+1}^{t}(L\|w_t-w_{t-1}\|+\rho D_t^x\n{w_t}+\rho D_{t-1}^x\n{w_{t-1}})^2}}{\sqrt{b}}
\end{align}
with probability $1-2\zeta$, where $t$ belongs to $[sm+1,(s+1)m]$.
Note that we can further relax the parameter $C_3$ in \eqref{eq:highvar2x} to $C_2=O(\log\frac{d \mathT}{\zeta})$ (see \eqref{eq:highvar2})
for making sure the above arguments hold with probability $1-\zeta$ for all $t\leq \mathT$ by using a union bound for $\zeta_t$'s:
\begin{align}\label{eq:highvar2}
\n{y_{t}}=\n{y_t-y_{sm}} \leq \frac{C_2\sqrt{\sum_{k=sm+1}^{t}(L\|w_t-w_{t-1}\|+\rho D_t^x\n{w_t}+\rho D_{t-1}^x\n{w_{t-1}})^2}}{\sqrt{b}}.
\end{align}

Now, we will show how to bound the right-hand-side of \eqref{eq:highvar2} to finish the proof, i.e., prove the remaining second bound $\n{y_t}\leq \eta\gamma L(\base)^t r_0$.

First, we show that the last two terms in the right-hand-side of \eqref{eq:highvar2} can be bounded as
\begin{align}
\rho D_t^x\n{w_t}+\rho D_{t-1}^x\n{w_{t-1}} &\leq
\rho\big(\Dtop\big)\frac{3}{2}(\base)^t r_0 + \rho\big(\Dtop\big)\frac{3}{2}(\base)^{t-1} r_0 \notag\\
& \leq 3\rho\big(\Dtop\big)(\base)^t r_0 \notag\\
& \leq \frac{6\delta}{C_1}(\base)^t r_0, \label{eq:50}
\end{align}
where the first inequality follows from the induction of $\n{w_{t-1}}\leq\frac{3}{2}(\base)^{t-1} r_0$ and the already proved
$\n{w_t}\leq\frac{3}{2}(\base)^t r_0$ in \eqref{eq:wt},
and the last inequality holds by letting the perturbation radius $r\leq \frac{\delta}{C_1 \rho}$.

Now, we show that the first term of right-hand-side of \eqref{eq:highvar2} can be bounded as
\begin{align}
L\|w_t-w_{t-1}\| &=
L\big\|-\eta \hess(I-\eta \hess)^{t-1}w_0
        -\eta\sum_{\tau=0}^{t-2}\eta \hess(I-\eta \hess)^{t-2-\tau}(\Delta_\tau w_\tau+y_\tau)
        +\eta(\Delta_{t-1} w_{t-1}+y_{t-1}) \big\|
\notag\\
& \leq L\eta \gamma(\base)^{t-1}r_0
        + L\big\|\eta\sum_{\tau=0}^{t-2}
                \eta \hess(I-\eta \hess)^{t-2-\tau}(\Delta_\tau w_\tau+y_\tau) \big\|
        + L\|\eta(\Delta_{t-1} w_{t-1}+y_{t-1})\| \notag\\
& \leq L\eta \gamma(\base)^{t-1}r_0
        + L\eta\big\|\sum_{\tau=0}^{t-2}
                \eta \hess(I-\eta \hess)^{t-2-\tau}\big\|
                \max_{0\leq k\leq t-2}\n{\Delta_k w_k+y_k} \notag\\
        &\qquad
        + L\eta\rho\big(\Dtop\big)\| w_{t-1}\|+L\eta\|y_{t-1}\| \label{eq:60}\\
& \leq L\eta \gamma(\base)^{t-1}r_0
        + L\eta\sum_{\tau=0}^{t-2}\frac{1}{t-1-\tau}
                \max_{0\leq k\leq t-2}\n{\Delta_k w_k+y_k} \notag\\
        &\qquad
        + L\eta\rho\big(\Dtop\big)\| w_{t-1}\|+L\eta\|y_{t-1}\| \label{eq:61}\\
& \leq L\eta \gamma(\base)^{t-1}r_0
        + L\eta\log t
                \max_{0\leq k\leq t-2}\n{\Delta_k w_k+y_k} \notag\\
        &\qquad
        + L\eta\rho\big(\Dtop\big)\| w_{t-1}\|+L\eta\|y_{t-1}\| \notag\\
& \leq L\eta \gamma(\base)^{t-1}r_0
        + L\eta\log t
                \max_{0\leq k\leq t-2}\n{\Delta_k w_k+y_k} \notag\\
        &\qquad
        + L\eta\rho\big(\Dtop\big)\frac{3}{2}(\base)^{t-1} r_0
        +L\eta \eta\gamma L(\base)^{t-1} r_0 \label{eq:62}\\
& \leq L\eta \gamma(\base)^{t-1}r_0
        + L\eta\log t
                \Big(\rho\big(\Dtop\big)\frac{3}{2}(\base)^{t-2} r_0
                    +\eta\gamma L(\base)^{t-2} r_0\Big) \notag\\
        &\qquad
        + L\eta\rho\big(\Dtop\big)\frac{3}{2}(\base)^{t-1} r_0
        +L\eta \eta\gamma L(\base)^{t-1} r_0 \label{eq:63}\\
& \leq L\eta \gamma(\base)^{t-1}r_0
        + L\eta\log t
                \Big(\frac{3\delta}{C_1}(\base)^{t-2} r_0
                    +\eta\gamma L(\base)^{t-2} r_0\Big) \notag\\
        &\qquad
        +\frac{3 L\eta\delta}{C_1}(\base)^{t-1} r_0
        +L\eta \eta\gamma L(\base)^{t-1} r_0 \label{eq:64}\\
& \leq  \Big(\frac{4}{C_1}\log t +2L\eta\log t\Big)\eta\gamma L(\base)^t r_0, \label{eq:65}
\end{align}
where the first equality follows from \eqref{eq:expw},
\eqref{eq:60} holds from the following \eqref{eq:66},
\begin{align}\label{eq:66}
  \n{\Delta_{t}}\leq \rho D_t^x \leq \rho\big(\Dtop\big),
\end{align}
where \eqref{eq:66} holds due to Hessian Lipschitz Assumption \ref{asp:2}, \eqref{eq:distbound} and the perturbation radius $r$ (recall that $\Delta_{t}:=\int_0^1(\nabla^2 f(x_t'+\theta(x_t-x_t'))-\hess)d\theta$, $\hess:=\nabla^2 f(\tx)$ and $D_t^x:=\max\{\n{x_t-\tx},\n{x_t'-\tx}\}$),
\eqref{eq:61} holds due to $\n{\eta \hess(I-\eta \hess)^{t}}\leq \frac{1}{t+1}$,
\eqref{eq:62} holds by plugging the induction $\n{w_{t-1}}\leq\frac{3}{2}(\base)^{t-1} r_0$
and $\n{y_{t-1}}\leq \eta\gamma L(\base)^{t-1} r_0$,
\eqref{eq:63} follows from \eqref{eq:66}, the induction
$\n{w_{k}}\leq\frac{3}{2}(\base)^{k} r_0$ and
$\n{y_{k}}\leq \eta\gamma L(\base)^{k} r_0$ (hold for all $k\leq t-1$),
\eqref{eq:64} holds by letting the perturbation radius $r\leq \frac{\delta}{C_1 \rho}$,
and the last inequality holds due to $\gamma\geq \delta$ (recall $-\gamma:=\lambda_{\min}(\hess)=\lambda_{\min}(\nabla^2 f(\tx))\leq -\delta$).

By plugging \eqref{eq:50} and \eqref{eq:65} into \eqref{eq:highvar2}, we have
\begin{align}
\n{y_{t}}
&\leq C_2\left(\frac{6\delta}{C_1}(\base)^t r_0+\Big(\frac{4}{C_1}\log t+2L\eta\log t\Big)\eta\gamma L(\base)^t r_0\right) \notag\\
&\leq C_2\Big(\frac{6}{C_1\eta L}+\frac{4}{C_1}\log t+2L\eta\log t\Big)\eta\gamma L(\base)^t r_0 \notag\\
&\leq \eta\gamma L(\base)^t r_0, \label{eq:last}
\end{align}
where the second inequality holds due to $\gamma\geq \delta$,
and the last inequality holds by letting
$C_1\geq \frac{20C_2}{\eta L}$
and $\eta \leq \frac{1}{4C_2 L\log t}$.
Recall that $C_2=O(\log\frac{d \mathT}{\zeta})$ is enough to let the arguments
in this proof hold with probability $1-\zeta$ for all $t\leq \mathT$.

From \eqref{eq:wt} and \eqref{eq:last}, we know that the two induction bounds hold for $t$.
We recall the first induction bound here:
\begin{enumerate}
  \item $\frac{1}{2}(\base)^t r_0\leq\n{w_t}\leq\frac{3}{2}(\base)^t r_0$
\end{enumerate}
Thus, we know that $\n{w_t} \geq \frac{1}{2}(\base)^t r_0=\frac{1}{2}(\base)^t\frac{\zeta' r}{\sqrt{d}}$. However, $\n{w_t}:=\n{x_t-x_t'}\leq \n{x_t-x_0}+\n{x_0-\tx}+\n{x_t'-x_0'}+\n{x_0'-\tx}\leq 2r+2\frac{\delta}{C_1\rho}\leq \frac{4\delta}{C_1\rho}$ according to \eqref{eq:distbound} and the perturbation radius $r$.
The last inequality is due to the perturbation radius $r\leq \frac{\delta}{C_1 \rho}$ (we already used this condition in the previous arguments).
This will give a contradiction for \eqref{eq:distbound} if $\frac{1}{2}(\base)^t\frac{\zeta' r}{\sqrt{d}}\geq \frac{4\delta}{C_1\rho}$ and it will happen if $t\geq \frac{2\log(\frac{8\delta\sqrt{d}}{C_1\rho\zeta' r})}{\eta\delta}$.

So the proof of this lemma is finished by contradiction if we let $\mathT:=\frac{2\log(\frac{8\delta\sqrt{d}}{C_1\rho\zeta' r})}{\eta\delta}$, i.e., we have
\begin{align*}
\exists T\leq \mathT,~~ \max\{\n{x_T-x_0}, \n{x_T'-x_0'}\}\geq \frac{\delta}{C_1\rho}.
\end{align*}
\end{proofof}

\newpage
\subsection{Proofs for Online Problem}
\label{app:proofo}
In this section, we provide the detailed proofs for online problem \eqref{eq:online} (i.e., Theorem \ref{thm:o1}--\ref{thm:o2}).
We will reuse some parts of our previous proofs for finite-sum problem  \eqref{eq:finite} in previous Section \ref{app:prooff}.

First, we recall the previous key relation \eqref{eq:difff} between $f(x_t)$ and $f(x_{t-1})$ as follows (recall $x_t := x_{t-1}-\eta v_{t-1}$):
\begin{align}
f(x_t) \leq& f(x_{t-1})
                    + \frac{\eta}{2}\ns{\nabla f(x_{t-1})-v_{t-1}}
                    - \frac{\eta}{2}\ns{\nabla f(x_{t-1})}
                    - \big(\frac{1}{2\eta}- \frac{L}{2}\big)\ns{x_t-x_{t-1}}. \label{eq:difff1}
\end{align}

Next, we recall the previous bound \eqref{eq:useasp} for the variance term:
\begin{align}
 \E[\ns{v_{t-1}-\nabla f(x_{t-1})}] \leq \frac{L^2}{b}\E[\ns{x_{t-1}-x_{t-2}}]
        +\E[\| v_{t-2}-\nabla f(x_{t-2}) \|^2] \label{eq:useasp1}.
\end{align}
Now, the following bound for the variance term will be different from the previous finite-sum case.
Similar to \eqref{eq:var}, we sum up
\eqref{eq:useasp1} from the beginning of this epoch $sm$ to the point $t-1$,
\begin{align}
 \E[\ns{v_{t-1}-\nabla f(x_{t-1})}] &\leq \frac{L^2}{b}\sum_{j=sm+1}^{t-1}\E[\ns{x_{j}-x_{j-1}}]
        +\E[\| v_{sm}-\nabla f(x_{sm}) \|^2] \label{eq:var1} \\
        &= \frac{L^2}{b}\sum_{j=sm+1}^{t-1}\E[\ns{x_{j}-x_{j-1}}]
              +\E\Big[\Big\|\frac{1}{B}\sum_{j\in I_B}\nabla f_j(x_{sm})
                        -\nabla f(x_{sm})\Big\|^2\Big] \label{eq:usechange}\\
        &\leq \frac{L^2}{b}\sum_{j=sm+1}^{t-1}\E[\ns{x_{j}-x_{j-1}}]
                +\frac{\sigma^2}{B}
\label{eq:varonline},
\end{align}
where \eqref{eq:var1} is the same as \eqref{eq:var},
\eqref{eq:usechange} uses the modification \eqref{eq:batch} (i.e., $v_{sm}= \frac{1}{B}\sum_{j\in I_B}\nabla f_j(x_{sm})$ instead of the full gradient computation $v_{sm}= \nabla f(x_{sm})$ in the finite-sum case),
and the last inequality \eqref{eq:varonline} follows from the bounded variance Assumption \ref{asp:var}.

Now, we take expectations for \eqref{eq:difff1} and then sum it up from the beginning of this epoch $s$, i.e., iterations from $sm$ to $t$, by plugging the variance \eqref{eq:varonline} into them to get:
\begin{align}
\E[f(x_{t})] &\leq \E[f(x_{sm})] - \frac{\eta}{2}\sum_{j=sm+1}^{t}\E[\ns{\nabla f(x_{j-1})}]
        - \big(\frac{1}{2\eta}- \frac{L}{2}\big)\sum_{j=sm+1}^{t}\E[\ns{x_j-x_{j-1}}] \notag\\
        &\qquad + \frac{\eta L^2}{2b}\sum_{k=sm+1}^{t-1}\sum_{j=sm+1}^{k}\E[\ns{x_{j}-x_{j-1}}]
                + \frac{\eta}{2}\sum_{j=sm+1}^{t}\frac{\sigma^2}{B} \notag\\
    &\leq \E[f(x_{sm})] - \frac{\eta}{2}\sum_{j=sm+1}^{t}\E[\ns{\nabla f(x_{j-1})}]
        - \big(\frac{1}{2\eta}- \frac{L}{2}\big)\sum_{j=sm+1}^{t}\E[\ns{x_j-x_{j-1}}] \notag\\
        &\qquad + \frac{\eta L^2(t-1-sm)}{2b}\sum_{j=sm+1}^{t}\E[\ns{x_{j}-x_{j-1}}]
                + \frac{(t-sm)\eta\sigma^2}{2B}\notag\\
    &\leq \E[f(x_{sm})] - \frac{\eta}{2}\sum_{j=sm+1}^{t}\E[\ns{\nabla f(x_{j-1})}]
        - \big(\frac{1}{2\eta}- \frac{L}{2}\big)\sum_{j=sm+1}^{t}\E[\ns{x_j-x_{j-1}}] \notag\\
        &\qquad + \frac{\eta L^2}{2}\sum_{j=sm+1}^{t}\E[\ns{x_{j}-x_{j-1}}] + \frac{(t-sm)\eta\sigma^2}{2B} \label{eq:bm1}\\
    &\leq \E[f(x_{sm})] - \frac{\eta}{2}\sum_{j=sm+1}^{t}\E[\ns{\nabla f(x_{j-1})}] + \frac{(t-sm)\eta\sigma^2}{2B} \label{eq:eta1},
\end{align}
where \eqref{eq:bm1} holds if the minibatch size $b\geq m$ (note that here $t\leq (s+1)m$),
\eqref{eq:eta1} holds if the step size $\eta\leq \frac{\sqrt{5}-1}{2L}$.

\begin{proofof}{Theorem \ref{thm:o1}}
Let $b=m=\frac{2\sigma}{\epsilon}$ and step size $\eta\leq \frac{\sqrt{5}-1}{2L}$, then \eqref{eq:eta1} holds.
Now, the proof is directly obtained by summing up \eqref{eq:eta1} for all epochs $0\leq s\le S$ as follows:
\begin{align}
\E[f(x_{T})]
    &\leq \E[f(x_0)] - \frac{\eta}{2}\sum_{j=1}^{T}\E[\ns{\nabla f(x_{j-1})}] +\frac{T\eta\sigma^2}{2B} \notag\\
\E[\n{\nabla f(\hx)}]\leq  \sqrt{\E[\ns{\nabla f(\hx)}]} &\leq \sqrt{\frac{2(f(x_0)-f^*)}{\eta T}+\frac{\sigma^2}{B}} =\frac{\epsilon}{2}+\frac{\epsilon}{2}= \epsilon \label{eq:finitefinal},
\end{align}
where \eqref{eq:finitefinal} holds by choosing $\hx$ uniformly from $\{x_{t-1}\}_{t\in[T]}$
and letting
$Sm\leq T=\frac{8(f(x_0)-f^*)}{\eta\epsilon^2}=O(\frac{L(f(x_0)-f^*)}{\epsilon^2})$ and $B=\frac{4\sigma^2}{\epsilon^2}$.
Note that the total number of computation of stochastic gradients equals to
\begin{align*}
  SB+Smb\leq \Big\lceil\frac{T}{m}\Big\rceil B + Tb \leq \Big(\frac{T}{2\sigma/\epsilon}+1\Big)\frac{4\sigma^2}{\epsilon^2}+T\frac{2\sigma}{\epsilon}
  =\frac{4\sigma^2}{\epsilon^2}+2T\frac{2\sigma}{\epsilon}
=O\Big(\frac{\sigma^2}{\epsilon^2}+\frac{L(f(x_0)-f^*)\sigma}{\epsilon^3}\Big).
\end{align*}
\end{proofof}

\subsubsection{Proof of Theorem \ref{thm:o2}}

Similar to the proof of Theorem \ref{thm:f2},
for proving the second-order guarantee, we will divide the proof into two situations.
The first situation (\textbf{large gradients}) is also almost the same as the above arguments for first-order guarantee, where the function value will decrease a lot since the gradients are large (see \eqref{eq:eta1}).
For the second situation (\textbf{around saddle points}), we will show that the function value can also decrease a lot by adding a random perturbation. The reason is that saddle points are usually unstable and the stuck region is relatively small in a random perturbation ball.

\vspace{1mm}
\noindent{{\bf Large Gradients}: }
First, we need a high probability bound for the variance term instead of the expectation one \eqref{eq:varonline}. Then we use it to get a high probability bound of \eqref{eq:eta1} for function value decrease.
Note that in this online case, $v_{sm}= \frac{1}{B}\sum_{j\in I_B}\nabla f_j(x_{sm})$ at the beginning of each epoch (see \eqref{eq:batch}) instead of $v_{sm}= \nabla f(x_{sm})$ in the previous finite-sum case.
Thus we first need a high probability bound for $\n{v_{sm}-\nabla f(x_{sm})}$.
According to Assumption \ref{asp:var2}, we have
\begin{align*}
  \n{\nabla f_j(x)-\nabla f(x)} &\leq \sigma, \notag\\
  \sum_{j\in I_B}\ns{\nabla f_j(x)-\nabla f(x)} &\leq B\sigma^2.
\end{align*}
By applying Bernstein inequality (Proposition \ref{prop:bernstein}), we get the high probability bound for $\n{v_{sm}-\nabla f(x_{sm})}$ as follows:
\begin{align*}
\pr\Big\{\big\|v_{sm}-\nabla f(x_{sm})\big\|\geq \frac{t}{B} \Big\} &\leq  (d+1) \exp\Big(\frac{-t^2/2}{B\sigma^2+ \sigma t/3}\Big)
    \notag  = \zeta,
\end{align*}
where the last equality holds by letting $t=C\sqrt{B}\sigma$, where $C=O(\log\frac{d}{\zeta})=\tdo(1)$.
Now, we have a high probability bound for $\n{v_{sm}-\nabla f(x_{sm})}$, i.e.,
\begin{align}\label{eq:varhigh_start}
   \big\|v_{sm}-\nabla f(x_{sm})\big\| \leq  \frac{C\sigma}{\sqrt{B}} \quad \mathrm{~with~probability~} 1-\zeta.
\end{align}

Now we will try to obtain a high probability bound for the variance term of other points beyond the starting points. Recall that $v_k=\frac{1}{b}\sum_{i\in I_b}\big(\nabla f_i(x_{k})-\nabla f_i(x_{k-1})\big) + v_{k-1}$ (see Line \ref{line:v} of Algorithm \ref{alg:ssrgd_hl}), we let $y_k:=v_k-\nabla f(x_k)$ and $z_k:=y_k-y_{k-1}$.
It is not hard to verify that $\{y_k\}$ is a martingale sequence and $\{z_k\}$ is the associated martingale difference sequence.
In order to apply the Azuma-Hoeffding inequalities to get a high probability bound, we first need to bound the difference sequence $\{z_k\}$.
We use the Bernstein inequality to bound the differences as follows.
\begin{align}
  z_k=y_k-y_{k-1}&= v_k-\nabla f(x_k) - (v_{k-1}-\nabla f(x_{k-1})) \notag\\
  &=\frac{1}{b}\sum_{i\in I_b}\big(\nabla f_i(x_{k})-\nabla f_i(x_{k-1})\big) + v_{k-1}
        -\nabla f(x_k) - (v_{k-1}-\nabla f(x_{k-1})) \notag \\
  &=\frac{1}{b}\sum_{i\in I_b}\Big(\nabla f_i(x_{k})-\nabla f_i(x_{k-1})
        -(\nabla f(x_k) -\nabla f(x_{k-1}))\Big). \label{eq:zk1}
\end{align}
We define $u_i:=\nabla f_i(x_{k})-\nabla f_i(x_{k-1})-(\nabla f(x_k) -\nabla f(x_{k-1}))$, and then we have
\begin{align}
\|u_i\|=\|\nabla f_i(x_{k})-\nabla f_i(x_{k-1})-(\nabla f(x_k) -\nabla f(x_{k-1}))\|\leq 2\|x_{k}-x_{k-1}\|, \label{eq:b11}
\end{align}
where the last inequality holds due to the gradient Lipschitz Assumption \ref{asp:1}.
Then, consider the variance term
\begin{align}
  &\sum_{i\in I_b}\E[\|u_i\|^2] \notag\\
          &=\sum_{i\in I_b}\E[\ns{\nabla f_i(x_{k})-\nabla f_i(x_{k-1})-(\nabla f(x_k) -\nabla f(x_{k-1}))}] \notag\\
          &\leq \sum_{i\in I_b}\E[\ns{\nabla f_i(x_{k})-\nabla f_i(x_{k-1})}] \notag\\
          &\leq bL^2\ns{x_{k}-x_{k-1}}, \label{eq:b21}
\end{align}
where the first inequality uses the fact $\E[\ns{x-\E x}]\leq \E[\ns{x}]$, and the last inequality uses the gradient Lipschitz Assumption \ref{asp:1}.
According to \eqref{eq:b11} and \eqref{eq:b21}, we can bound the difference $z_k$ by Bernstein inequality (Proposition \ref{prop:bernstein}) as
\begin{align*}
\pr\Big\{\big\|z_k\big\|\geq \frac{t}{b} \Big\} &\leq (d+1) \exp\Big(\frac{-t^2/2}{\sigma^2+Rt/3}\Big) \notag \\
    & = (d+1) \exp\Big(\frac{-t^2/2}{bL^2\ns{x_{k}-x_{k-1}}+ 2\|x_{k}-x_{k-1}\|t/3}\Big)
    \notag \\
    & = \zeta_k,
\end{align*}
where the last equality holds by letting $t=CL\sqrt{b}\n{x_{k}-x_{k-1}}$, where $C=O(\log\frac{d}{\zeta_k})=\tdo(1)$.
Now, we have a high probability bound for the difference sequence $\{z_k\}$, i.e.,
\begin{align}
    \|z_k\| \leq c_k= \frac{CL\n{x_{k}-x_{k-1}}}{\sqrt{b}} \quad \mathrm{~with~probability~} 1-\zeta_k.
\end{align}

Now, we are ready to get a high probability bound for our original variance term \eqref{eq:varonline} by using the martingale Azuma-Hoeffding inequality.
Consider in a specifical epoch $s$, i.e, iterations $t$ from $sm+1$ to current $sm+k$, where $k$ is less than $m$.
According to Azuma-Hoeffding inequality (Proposition \ref{prop:azumahigh}) and letting $\zeta_k = \zeta/m$, we have
\begin{align*}
\pr\Big\{\big\|y_{sm+k}-y_{sm}\big\|\geq \beta \Big\} &\leq (d+1) \exp\Big(\frac{-\beta^2}{8\sum_{t=sm+1}^{sm+k} c_t^2}\Big)+\zeta \notag \\
    & = 2\zeta,
\end{align*}
where the last equality holds by letting $\beta=\sqrt{8\sum_{t=sm+1}^{sm+k} c_t^2\log\frac{d}{\zeta}}
=\frac{C'L\sqrt{\sum_{t=sm+1}^{sm+k}\ns{x_{t}-x_{t-1}}}}{\sqrt{b}}$, where $C'=O(C\sqrt{\log\frac{d}{\zeta}})=\tdo(1)$.
Recall that $y_k:=v_k-\nabla f(x_k)$ and at the beginning point of this epoch $\n{y_{sm}}=\n{v_{sm}-\nabla f(x_{sm})}\leq C\sigma/\sqrt{B}$ with probability $1-\zeta$, where $C=O(\log\frac{d}{\zeta})=\tdo(1)$ (see \eqref{eq:varhigh_start}).
Combining with \eqref{eq:varhigh_start} and using a union bound, we have
\begin{align}\label{eq:highvar1}
\n{v_{t-1}-\nabla f(x_{t-1})}=\n{y_{t-1}} \leq \beta +\n{y_{sm}} \leq \frac{C'L\sqrt{\sum_{j=sm+1}^{t-1}\ns{x_{j}-x_{j-1}}}}{\sqrt{b}} +\frac{C\sigma}{\sqrt{B}}
\end{align}
with probability $1-3\zeta$, where $t$ belongs to $[sm+1,(s+1)m]$.

Now, we use this high probability version \eqref{eq:highvar1} instead of the expectation one \eqref{eq:varonline} to obtain the high probability bound for function value decrease (see \eqref{eq:eta1}).
We sum up \eqref{eq:difff1} from the beginning of this epoch $s$, i.e., iterations from $sm$ to $t$, by plugging \eqref{eq:highvar1} into them to get:
\begin{align}
f(x_{t}) &\leq f(x_{sm}) - \frac{\eta}{2}\sum_{j=sm+1}^{t}\ns{\nabla f(x_{j-1})}
        - \big(\frac{1}{2\eta}- \frac{L}{2}\big)\sum_{j=sm+1}^{t}\ns{x_j-x_{j-1}} \notag\\
        &\qquad + \frac{\eta}{2}\sum_{k=sm+1}^{t-1}\frac{2C'^2L^2\sum_{j=sm+1}^{k}\ns{x_{j}-x_{j-1}}}{b}
    +\frac{\eta}{2}\sum_{j=sm+1}^{t}\frac{2C^2\sigma^2}{B} \notag\\
    &\leq f(x_{sm}) - \frac{\eta}{2}\sum_{j=sm+1}^{t}\ns{\nabla f(x_{j-1})}
        - \big(\frac{1}{2\eta}- \frac{L}{2}\big)\sum_{j=sm+1}^{t}\ns{x_j-x_{j-1}} \notag\\
        &\qquad + \frac{\eta C'^2L^2}{b}\sum_{k=sm+1}^{t-1}\sum_{j=sm+1}^{k}\ns{x_{j}-x_{j-1}}
   + \frac{(t-sm)\eta C^2\sigma^2}{B} \notag\\
    &\leq f(x_{sm}) - \frac{\eta}{2}\sum_{j=sm+1}^{t}\ns{\nabla f(x_{j-1})}
        - \big(\frac{1}{2\eta}- \frac{L}{2}\big)\sum_{j=sm+1}^{t}\ns{x_j-x_{j-1}} \notag\\
        &\qquad + \frac{\eta C'^2L^2(t-1-sm)}{b}\sum_{j=sm+1}^{t}\ns{x_{j}-x_{j-1}}
+ \frac{(t-sm)\eta C^2\sigma^2}{B} \notag\\
    &\leq f(x_{sm}) - \frac{\eta}{2}\sum_{j=sm+1}^{t}\ns{\nabla f(x_{j-1})}
        - \big(\frac{1}{2\eta}- \frac{L}{2}-\eta C'^2L^2\big)\sum_{j=sm+1}^{t}\ns{x_j-x_{j-1}} \notag\\
        &\qquad + \frac{(t-sm)\eta C^2\sigma^2}{B} \label{eq:bm21}\\
    &\leq f(x_{sm}) - \frac{\eta}{2}\sum_{j=sm+1}^{t}\ns{\nabla f(x_{j-1})} + \frac{(t-sm)\eta C^2\sigma^2}{B} \label{eq:eta21},
\end{align}
where \eqref{eq:bm21} holds if the minibatch size $b\geq m$ (note that here $t\leq (s+1)m$), and
\eqref{eq:eta21} holds if the step size $\eta\leq \frac{\sqrt{8C'^2+1}-1}{4C'^2L}$.

Similar to the previous finite-sum case, \eqref{eq:eta21} only guarantees function value decrease when the summation of gradients in this epoch is large. However, in order to connect the guarantees between first situation (large gradients) and second situation (around saddle points), we need to show guarantees that are related to the \emph{gradient of the starting point} of each epoch (see Line \ref{line:super} of Algorithm \ref{alg:ssrgd}).
As we discussed in previous Section \ref{app:prooff2}, we achieve this by stopping the epoch at a uniformly random point (see Line \ref{line:randomstop} of Algorithm \ref{alg:ssrgd}).

We want to point out that the second situation will have a little difference due to \eqref{eq:batch}, i.e., the full gradient of the starting point is not available (see Line \ref{line:super} of Algorithm \ref{alg:ssrgd}).
Thus some modifications are needed for previous Lemma \ref{lem:first}, we use the following lemma to connect these two situations (large gradients and around saddle points):

\begin{lemma}[Connection of Two Situations]
\label{lem:firstonline}
For any epoch $s$, let $x_t$ be a point uniformly sampled from this epoch $\{x_{j} \}_{j=sm}^{(s+1)m}$ and choose the step size $\eta \leq \frac{\sqrt{8C'^2+1}-1}{4C'^2L}$ (where $C'=O(\log\frac{dm}{\zeta})=\tdo(1)$) and the minibatch size $b\geq m$.
Then for any $\mathG$, by letting batch size $B\geq \frac{256C^2\sigma^2}{\mathG^2}$ (where $C=O(\log\frac{d}{\zeta})=\tdo(1)$), we have two cases:
\begin{enumerate}
\item If at least half of points in this epoch have gradient norm no larger than $\frac{\mathG}{2}$, then $\n{\nabla f(x_{(s+1)m})}\leq \frac{\mathG}{2}$ and $\|v_{(s+1)m} \| \le \mathG$ hold with probability at least $1/3$;
\item Otherwise, we know $f(x_{sm}) - f(x_t) \ge \frac{7\eta m\mathG^2}{256}$ holds with probability at least $1/5.$
\end{enumerate}
Moreover, $f(x_t) \le f(x_{sm}) + \frac{(t-sm)\eta C^2\sigma^2}{B}$ holds with high probability no matter which case happens.
\end{lemma}

\begin{proofof}{Lemma~\ref{lem:firstonline}}
There are two cases in this epoch:
\begin{enumerate}
\item If at least half of points of in this epoch $\{x_{j} \}_{j=sm}^{(s+1)m}$ have gradient norm no larger than $\frac{\mathG}{2}$, then it is easy to see that a uniformly sampled point $x_t$ has gradient norm $\n{\nabla f(x_t)}\leq \frac{\mathG}{2}$ with probability at least $1/2.$ Moreover, note that the starting point of the next epoch $x_{(s+1)m}=x_t$ (i.e., Line \ref{line:randompoint} of Algorithm \ref{alg:ssrgd}), thus we have $\n{\nabla f(x_{(s+1)m})}\leq \frac{\mathG}{2}$ with probability $1/2$. According to \eqref{eq:varhigh_start}, we have
   $\|v_{(s+1)m}-\nabla f(x_{(s+1)m})\| \leq  \frac{C\sigma}{\sqrt{B}}$ with probability $1-\zeta$, where $C=O(\log\frac{d}{\zeta})=\tdo(1)$.
By a union bound, with probability at least $1/3$, we have
$$\|v_{(s+1)m}\|\leq \frac{C\sigma}{\sqrt{B}} + \frac{\mathG}{2}\leq \frac{\mathG}{16}+\frac{\mathG}{2} \leq \mathG.$$
\item Otherwise, at least half of points have gradient norm larger than $\frac{\mathG}{2}$. Then, as long as the sampled point $x_t$ falls into the last quarter of $\{x_{j} \}_{j=sm}^{(s+1)m}$, we know $\sum_{j=sm+1}^{t}\ns{\nabla f(x_{j-1})}\geq \frac{m\mathG^2}{16}$. This holds with probability at least $1/4$ since $x_t$ is uniformly sampled. Then by combining with \eqref{eq:eta21}, we obtain the function value decrease
 $$f(x_{sm}) - f(x_t) \geq \frac{\eta}{2}\sum_{j=sm+1}^{t}\ns{\nabla f(x_{j-1})}-\frac{(t-sm)\eta C^2\sigma^2}{B} \ge \frac{\eta m\mathG^2}{32}-\frac{\eta m\mathG^2}{256}=\frac{7\eta m\mathG^2}{256},$$ where the last inequality is due to $B\geq \frac{256C^2\sigma^2}{\mathG^2}$.
Note that
    \eqref{eq:eta21} holds with high probability if we choose the minibatch size $b\geq m$ and the step size $\eta\leq \frac{\sqrt{8C'^2+1}-1}{4C'^2L}$.
    By a union bound, the function value decrease $f(x_{sm}) - f(x_t) \ge \frac{7\eta m\mathG^2}{256}$ with probability at least $1/5$.
\end{enumerate}
Again according to \eqref{eq:eta21}, $f(x_t)\leq f(x_{sm})+\frac{(t-sm)\eta C^2\sigma^2}{B}$ always holds with high probability.
\end{proofof}

Note that if Case 2 happens, the function value already decreases a lot in this epoch $s$ (corresponding to the first situation large gradients). Otherwise Case 1 happens, we know the starting point of the next epoch $x_{(s+1)m}=x_t$ (i.e., Line \ref{line:randompoint} of Algorithm \ref{alg:ssrgd}), then we know $\n{\nabla f(x_{(s+1)m})}\leq \frac{\mathG}{2}$ and $\|v_{(s+1)m} \| \le \mathG$. Then we will start a super epoch (corresponding to the second situation around saddle points). Note that if $\lambda_{\min}(\nabla^2 f(x_{(s+1)m}))> -\delta$, this point $x_{(s+1)m}$ is already an $(\epsilon,\delta)$-second-order stationary point (recall that $\mathG \leq \epsilon$ in our Theorem \ref{thm:o2}).

\vspace{3mm}
\noindent{{\bf Around Saddle Points} $\|v_{(s+1)m} \| \le \mathG$ and $\lambda_{\min}(\nabla^2 f(x_{(s+1)m})) \leq -\delta$: }
In this situation, we will show that the function value decreases a lot in a \emph{super epoch} (instead of an epoch as in the first situation) with high probability by adding a random perturbation at the initial point $\tx=x_{(s+1)m}$. To simplify the presentation, we use $x_0:=\tx+\xi$ to denote the starting point of the super epoch after the perturbation, where $\xi$ uniformly $\sim \mathbb{B}_0(r)$ and the perturbation radius is $r$ (see Line \ref{line:init} in Algorithm \ref{alg:ssrgd}).
Following the classical widely used \emph{two-point analysis} developed in \citep{jin2017escape}, we consider two coupled points $x_0$ and $x_0'$ with $w_0:=x_0-x_0'=r_0e_1$, where $r_0$ is a scalar and $e_1$ denotes the smallest eigenvector direction of Hessian $\hess := \nabla^2 f(\tx)$. Then we get two coupled sequences $\{x_t\}$ and $\{x_t'\}$ by running SSRGD update steps (Line \ref{line:up1}--\ref{line:up2} of Algorithm \ref{alg:ssrgd}) with the same choice of batches and minibatches (i.e., $I_B$'s (see \eqref{eq:batch} and Line \ref{line:up1}) and $I_b$'s (see Line \ref{line:up2}))for a super epoch.
We will show that at least one of these two coupled sequences will decrease the function value a lot (escape the saddle point), i.e.,
\begin{align}\label{eq:funcd1}
 \exists t\leq\mathT, \mathrm{~~such ~that~~} \max\{f(x_0)-f(x_t), f(x_0')-f(x_t')\} \geq 2\mathf.
\end{align}
We will prove \eqref{eq:funcd1} by contradiction. Assume the contrary, $f(x_0)-f(x_t)<2\mathf$ and $f(x_0')-f(x_t')<2\mathf$.
First, we show that if function value does not decrease a lot, then all iteration points are not far from the starting point with high probability.
Then we will show that the stuck region is relatively small in the random perturbation ball, i.e., at least one of $x_t$ and $x_t'$ will go far away from their starting point $x_0$ and $x_0'$ with high probability. Thus there is a contradiction.
Similar to Lemma \ref{lem:local} and Lemma \ref{lem:smallstuck}, we
need the following two lemmas. Their proofs are deferred to the end of this section.

\begin{lemma}[Localization]
\label{lem:localonline}
Let $\{x_t\}$ denote the sequence by running SSRGD update steps (Line \ref{line:up1}--\ref{line:up2} of Algorithm \ref{alg:ssrgd}) from $x_0$.
Moreover, let the step size $\eta\leq \frac{1}{4C'L}$ and minibatch size $b\geq m$, with probability $1-\zeta$, we have
\begin{align}\label{eq:localonline}
\forall t,~~ \n{x_t-x_0}\leq \sqrt{\frac{4t(f(x_0)-f(x_t))}{5C'L} +\frac{4t^2\eta C^2\sigma^2}{5C'LB}},
\end{align}
where $C'=O(\log\frac{dt}{\zeta})=\tdo(1)$ and $C=O(\log\frac{dt}{\zeta m})=\tdo(1)$.
\end{lemma}

\begin{lemma}[Small Stuck Region]
\label{lem:smallstuckonline}
If the initial point $\tx$ satisfies $-\gamma:=\lambda_{\min}(\nabla^2 f(\tx))\leq -\delta$,
then let $\{x_t\}$ and $\{x_t'\}$ be two coupled sequences by running SSRGD update steps (Line \ref{line:up1}--\ref{line:up2} of Algorithm \ref{alg:ssrgd}) with the same choice of batches and minibatches (i.e., $I_B$'s (see \eqref{eq:batch} and Line \ref{line:up1}) and $I_b$'s (see Line \ref{line:up2})) from $x_0$ and $x_0'$ with $w_0:=x_0-x_0'=r_0e_1$, where $x_0\in\mathbb{B}_{\tx}(r)$, $x_0'\in\mathbb{B}_{\tx}(r)$ , $r_0=\frac{\zeta' r}{\sqrt{d}}$ and $e_1$ denotes the smallest eigenvector direction of Hessian $\nabla^2 f(\tx)$.
Moreover, let the super epoch length $\mathT=\frac{2\log(\frac{8\delta\sqrt{d}}{C_1\rho\zeta' r})}{\eta\delta}=\tdo(\frac{1}{\eta\delta})$, the step size $\eta\leq \min\big(\frac{1}{16\log(\frac{8\delta\sqrt{d}}{C_1\rho\zeta' r})L}, \frac{1}{8C_2L\log \mathT}\big)=\tdo(\frac{1}{L})$, minibatch size $b\geq m$, batch size $B=\tdo(\frac{\sigma^2}{\mathG^2})$ and
the perturbation radius $r\leq \frac{\delta}{C_1 \rho}$, then with probability $1-\zeta$, we have
\begin{align}\label{eq:stuckonline}
\exists T\leq \mathT,~~ \max\{\n{x_T-x_0}, \n{x_T'-x_0'}\}\geq \frac{\delta}{C_1\rho},
\end{align}
where $C_1\geq \frac{20C_2}{\eta L}$, $C_2=O(\log\frac{d \mathT}{\zeta})=\tdo(1)$ and $C_2'=O(\log\frac{d \mathT}{\zeta m})=\tdo(1)$.
\end{lemma}

Based on these two lemmas, we are ready to show that \eqref{eq:funcd1} holds with high probability. Without loss of generality, we assume $\n{x_T-x_0} \geq \frac{\delta}{C_1\rho}$ in \eqref{eq:stuckonline} (note that \eqref{eq:localonline} holds for both $\{x_t\}$ and $\{x_t'\}$), then plugging it into \eqref{eq:localonline} to obtain
\begin{align}
 \sqrt{\frac{4T(f(x_0)-f(x_T))}{5C'L} +\frac{4T^2\eta C^2\sigma^2}{5C'LB}} &\geq \frac{\delta}{C_1\rho} \notag\\
 f(x_0)-f(x_T) &\geq \frac{5C'L\delta^2}{4C_1^2\rho^2T}
            -\frac{T\eta C^2\sigma^2}{B}\notag\\
            &\geq \frac{5\eta C'L\delta^3}{8C_1^2\rho^2\log(\frac{8\delta\sqrt{d}}{C_1\rho\zeta' r})}
 - \frac{2C^2\sigma^2\log(\frac{8\delta\sqrt{d}}{C_1\rho\zeta' r})}{B\delta}
\label{eq:plugTthres}\\
            &\geq\frac{\delta^3}{C_1'\rho^2} \label{eq:largeeta1}\\
            &=2\mathf,\notag
\end{align}
where \eqref{eq:plugTthres} is due to $T\leq \mathT$ and \eqref{eq:largeeta1} holds by letting
$C_1'=\frac{8C_1^2\log(\frac{8\delta\sqrt{d}}{C_1\rho\zeta' r})}{4\eta C'L}$.
Recall that $B=\tdo(\frac{\sigma^2}{\mathG^2})$ and $\mathG\leq \delta^2/\rho$.
Thus, we already prove that at least one of sequences $\{x_t\}$ and $\{x_t'\}$
escapes the saddle point with high probability, i.e.,
\begin{align}
\exists T\leq \mathT~~, \max\{f(x_0)-f(x_T), f(x_0')-f(x_T')\} \geq 2\mathf,
\end{align}
if their starting points $x_0$ and $x_0'$ satisfying $w_0:=x_0-x_0'=r_0e_1$, where $r_0= \frac{\zeta' r}{\sqrt{d}}$ and $e_1$ denotes the smallest eigenvector direction of Hessian $\hess := \nabla^2 f(\tx)$.
Similar to the classical argument in \citep{jin2017escape}, we know that in the random perturbation ball, the stuck points can only be a short interval in the $e_1$ direction, i.e.,
at least one of two points in the $e_1$ direction will escape the saddle point if their distance is larger than $r_0=\frac{\zeta' r}{\sqrt{d}}$.
Thus, we know that the probability of the starting point $x_0=\tx+\xi$ (where $\xi$ uniformly $\sim \mathbb{B}_0(r)$) located in the stuck region is less than
\begin{align}\label{eq:goodx01}
  \frac{r_0V_{d-1}(r)}{V_d(r)}=
  \frac{r_0\Gamma(\frac{d}{2}+1)}{\sqrt{\pi}r\Gamma(\frac{d}{2}+\frac{1}{2})}
  \leq \frac{r_0}{\sqrt{\pi}r}\big(\frac{d}{2}+1\big)^{1/2}
  \leq \frac{r_0\sqrt{d}}{r}=\zeta',
\end{align}
where $V_d(r)$ denotes the volume of a Euclidean ball with radius $r$ in $d$ dimension,
and the first inequality holds due to Gautschi's inequality.
By a union bound for \eqref{eq:goodx01} and \eqref{eq:largeeta1} (holds with high probability if $x_0$ is not in a stuck region), we know
\begin{align}\label{eq:escape01}
f(x_0)-f(x_T) \geq 2\mathf=\frac{\delta^3}{C_1'\rho^2}
\end{align}
with high probability.
Note that the initial point of this super epoch is $\tx$ before the perturbation (see Line \ref{line:init} of Algorithm \ref{alg:ssrgd}), thus we need to show that the perturbation step $x_0=\tx+\xi$ (where $\xi$ uniformly $\sim \mathbb{B}_0(r)$) does not increase the function value a lot, i.e.,
\begin{align}
f(x_0)&\leq f(\tx) +\inner{\nabla f(\tx)}{x_0-\tx}
                    + \frac{L}{2}\ns{x_0-\tx}  \notag \\
        &\leq f(\tx) +\n{\nabla f(\tx)}\n{x_0-\tx}
                    + \frac{L}{2}\ns{x_0-\tx}  \notag\\
        &\leq f(\tx) +\mathG \cdot r +\frac{L}{2}r^2 \notag\\
        &\leq f(\tx) + \frac{\delta^3}{2C_1'\rho^2} \notag\\
        &= f(\tx) +\mathf, \label{eq:perturbless1}
\end{align}
where the last inequality holds by letting the perturbation radius $r\leq \min\{\frac{\delta^3}{4C_1'\rho^2\mathG}, \sqrt{\frac{\delta^3}{2C_1'\rho^2L}}\}$.

Now we combine with \eqref{eq:escape01} and \eqref{eq:perturbless1} to obtain with high probability
\begin{align}\label{eq:escapehigh1}
f(\tx)-f(x_T)=f(\tx)-f(x_0)+f(x_0)-f(x_T) \geq -\mathf+2\mathf=\frac{\delta^3}{2C_1'\rho^2}.
\end{align}

Thus we have finished the proof for the second situation (around saddle points), i.e., we show that the function value decrease a lot ($\mathf=\frac{\delta^3}{2C_1'\rho^2}$) in a \emph{super epoch} (recall that $T\leq \mathT=\frac{2\log(\frac{8\delta\sqrt{d}}{C_1\rho\zeta' r})}{\eta\delta}$) by adding a random perturbation $\xi \sim \mathbb{B}_0(r)$ at the initial point $\tx$.

\vspace{3mm}
\noindent{{\bf Combing these two situations (large gradients and around saddle points) to prove Theorem \ref{thm:o2}:}}
First, we recall Theorem \ref{thm:o2} here since we want to recall the parameter setting.
\begingroup
\def\thetheorem{\ref{thm:o2}}
\begin{theorem}
Under Assumption \ref{asp:1}, \ref{asp:2} (i.e. \eqref{smoothg2} and \eqref{smoothh2}) and Assumption \ref{asp:var2}, let $\Delta f:= f(x_0)-f^*$, where $x_0$ is the initial point and $f^*$ is the optimal value of $f$. By letting step size $\eta=\tdo(\frac{1}{L})$, batch size $B=\tdo(\frac{\sigma^2}{\mathG^2})=\tdo(\frac{\sigma^2}{\epsilon^2})$, minibatch size $b=\sqrt{B}=\tdo(\frac{\sigma}{\epsilon})$, epoch length $m=b$, perturbation radius $r=\tdo\big(\min(\frac{\delta^3}{\rho^2\epsilon}, \frac{\delta^{3/2}}{\rho\sqrt{L}})\big)$, threshold gradient $\mathG=\epsilon\leq \delta^2/\rho$, threshold function value $\mathf=\tdo(\frac{\delta^3}{\rho^2})$ and super epoch length $\mathT=\tdo(\frac{1}{\eta\delta})$, SSRGD will at least once get to an $(\epsilon,\delta)$-second-order stationary point with high probability using
\begin{equation*}
  \tdo\Big(\frac{L\Delta f\sigma}{\epsilon^3}
+\frac{\rho^2\Delta f\sigma^2}{\epsilon^2\delta^3}
+ \frac{L\rho^2\Delta f\sigma}{\epsilon\delta^4}\Big)
\end{equation*}
stochastic gradients for nonconvex online problem \eqref{eq:online}.
\end{theorem}
\addtocounter{theorem}{-1}
\endgroup

\begin{proofof}{Theorem \ref{thm:o2}}
Now, we prove this theorem by distinguishing the epochs into three types as follows:
\begin{enumerate}
  \item \emph{Type-1 useful epoch}: If at least half of points in this epoch have gradient norm larger than $\mathG$ (Case 2 of Lemma \ref{lem:firstonline});
  \item \emph{Wasted epoch}: If at least half of points in this epoch have gradient norm no larger than $\mathG$ and the starting point of the next epoch has estimated gradient norm larger than $\mathG$ (it means that this epoch does not guarantee decreasing the function value a lot as the large gradients situation, also it cannot connect to the second super epoch situation since the starting point of the next epoch has estimated gradient norm larger than $\mathG$);
  \item \emph{Type-2 useful super epoch}: If at least half of points in this epoch have gradient norm no larger than $\mathG$ and the starting point of the next epoch (here we denote this point as $x_{(s+1)m)}$) has estimated gradient norm no larger than $\mathG$ (i.e., $\n{v_{(s+1)m}}\leq \mathG$) (Case 1 of Lemma \ref{lem:firstonline}), according to Line \ref{line:super} of Algorithm \ref{alg:ssrgd}, we will start a super epoch. So here we denote this epoch along with its following super epoch as a type-2 useful super epoch.
\end{enumerate}
First, it is easy to see that the probability of a wasted epoch happened is less than $2/3$ due to the random stop (see Case 1 of Lemma \ref{lem:firstonline} and Line \ref{line:randomstop} of Algorithm \ref{alg:ssrgd}) and different wasted epoch are independent.
Thus, with high probability, there are at most $\tdo(1)$ wasted epochs happened before a type-1 useful epoch or type-2 useful super epoch.
Now, we use $N_1$ and $N_2$ to denote the number of type-1 useful epochs and type-2 useful super epochs that the algorithm is needed. Recall that $\Delta f:= f(x_0)-f^*$, where $x_0$ is the initial point and $f^*$ is the optimal value of $f$.

For type-1 useful epoch, according to Case 2 of Lemma \ref{lem:firstonline}, we know that the function value decreases at least $\frac{7\eta m\mathG^2}{256}$ with probability at least $1/5$.
Using a standard concentration, we know that with high probability $N_1$ type-1 useful epochs will decrease the function value at least $\frac{7\eta m\mathG^2N_1}{1536}$, note that the function value can decrease at most $\Delta f$.
So $\frac{7\eta m\mathG^2N_1}{1536}\leq \Delta f$, we get $N_1\leq \frac{1536\Delta f}{7\eta m\mathG^2}$.

For type-2 useful super epoch, first we know that the starting point of the super epoch $\tx:=x_{(s+1)m}$ has gradient norm $\n{\nabla f(\tx)}\leq \mathG/2$ and estimated gradient norm $\n{v_{(s+1)m}}\leq \mathG$. Now if $\lambda_{\min}(\nabla^2 f(\tx))\geq -\delta$, then $\tx$ is already a $(\epsilon,\delta)$-second-order stationary point. Otherwise,
$\n{v_{(s+1)m}}\leq \mathG$ and $\lambda_{\min}(\nabla^2 f(\tx)) \leq -\delta$, this is exactly our second situation (around saddle points).
According to \eqref{eq:escapehigh1}, we know that the the function value decrease ($f(\tx)-f(x_T)$) is at least $\mathf=\frac{\delta^3}{2C_1'\rho^2}$ with high probability.
Similar to type-1 useful epoch, we know $N_2\leq \frac{C_1''\rho^2\Delta f}{\delta^3}$ by a union bound (so we change $C_1'$ to $C_1''$, anyway we also have $C_1''=\tdo(1)$).

Now, we are ready to compute the convergence results to finish the proof for Theorem \ref{thm:o2}.
\begin{align}
&N_1(\tdo(1)B+B+mb) +N_2(\tdo(1)B+\big\lceil\frac{\mathT}{m}\big\rceil B+\mathT b) \\
&\leq \tdo\Big(\frac{\Delta f\sigma}{\eta \mathG^2\epsilon}+\frac{\rho^2\Delta f}{\delta^3}(\frac{\sigma^2}{\epsilon^2}+\frac{\sigma}{\eta \delta\epsilon})\Big) \notag\\
& \leq \tdo\Big(\frac{L\Delta f\sigma}{\epsilon^3}
+\frac{\rho^2\Delta f\sigma^2}{\epsilon^2\delta^3}
+ \frac{L\rho^2\Delta f\sigma}{\epsilon\delta^4}\Big)
\end{align}
\end{proofof}

Now, the only remaining thing is to prove Lemma \ref{lem:localonline} and \ref{lem:smallstuckonline}.
We provide these two proofs as follows.

\begingroup
\def\thelemma{\ref{lem:localonline}}
\begin{lemma}[Localization]
Let $\{x_t\}$ denote the sequence by running SSRGD update steps (Line \ref{line:up1}--\ref{line:up2} of Algorithm \ref{alg:ssrgd}) from $x_0$.
Moreover, let the step size $\eta\leq \frac{1}{4C'L}$ and minibatch size $b\geq m$, with probability $1-\zeta$, we have
\begin{align*}
\forall t,~~ \n{x_t-x_0}\leq \sqrt{\frac{4t(f(x_0)-f(x_t))}{5C'L} +\frac{4t^2\eta C^2\sigma^2}{5C'LB}},
\end{align*}
where $C'=O(\log\frac{dt}{\zeta})=\tdo(1)$ and $C=O(\log\frac{dt}{\zeta m})=\tdo(1)$.
\end{lemma}
\addtocounter{lemma}{-1}
\endgroup

\begin{proofof}{Lemma~\ref{lem:localonline}}
First, we assume the variance bound \eqref{eq:highvar1} holds for all $0\leq j\leq t-1$ (this is true with high probability using a union bound by letting $C'=O(\log\frac{dt}{\zeta})$ and $C=O(\log\frac{dt}{\zeta m})$).
Then, according to \eqref{eq:bm21}, we know for any $\tau \leq t$ in some epoch $s$
\begin{align}
f(x_{\tau})&\leq f(x_{sm}) - \frac{\eta}{2}\sum_{j=sm+1}^{\tau}\ns{\nabla f(x_{j-1})}
        - \big(\frac{1}{2\eta}- \frac{L}{2}-\eta C'^2L^2\big)\sum_{j=sm+1}^{\tau}\ns{x_j-x_{j-1}} \notag\\
        &\qquad + \frac{(\tau-sm)\eta C^2\sigma^2}{B} \notag\\
    &\leq f(x_{sm}) - \big(\frac{1}{2\eta}- \frac{L}{2}-\eta C'^2L^2\big)\sum_{j=sm+1}^{\tau}\ns{x_j-x_{j-1}}+ \frac{(\tau-sm)\eta C^2\sigma^2}{B} \notag\\
    &\leq f(x_{sm}) - \frac{5C'L}{4}\sum_{j=sm+1}^{\tau}\ns{x_j-x_{j-1}}+ \frac{(\tau-sm)\eta C^2\sigma^2}{B}, \label{eq:dist1}
\end{align}
where the last inequality holds since the step size $\eta\leq \frac{1}{4C'L}$ and assuming $C'\geq 1$.
Now, we sum up \eqref{eq:dist1} for all epochs before iteration $t$,
\begin{align*}
f(x_{t})
    &\leq f(x_{0}) - \frac{5C'L}{4}\sum_{j=1}^{t}\ns{x_j-x_{j-1}}+ \frac{t\eta C^2\sigma^2}{B}.
\end{align*}
Then, the proof is finished as
\begin{align*}
   \n{x_t-x_0}\leq \sum_{j=1}^t\n{x_j-x_{j-1}}\leq \sqrt{t\sum_{j=1}^{t}\ns{x_j-x_{j-1}}}
    \leq \sqrt{\frac{4t(f(x_0)-f(x_t))}{5C'L} +\frac{4t^2\eta C^2\sigma^2}{5C'LB}}.
\end{align*}
\end{proofof}

\begingroup
\def\thelemma{\ref{lem:smallstuckonline}}
\begin{lemma}[Small Stuck Region]
If the initial point $\tx$ satisfies $-\gamma:=\lambda_{\min}(\nabla^2 f(\tx))\leq -\delta$,
then let $\{x_t\}$ and $\{x_t'\}$ be two coupled sequences by running SSRGD update steps (Line \ref{line:up1}--\ref{line:up2} of Algorithm \ref{alg:ssrgd}) with the same choice of batches and minibatches (i.e., $I_B$'s (see \eqref{eq:batch} and Line \ref{line:up1}) and $I_b$'s (see Line \ref{line:up2})) from $x_0$ and $x_0'$ with $w_0:=x_0-x_0'=r_0e_1$, where $x_0\in\mathbb{B}_{\tx}(r)$, $x_0'\in\mathbb{B}_{\tx}(r)$ , $r_0=\frac{\zeta' r}{\sqrt{d}}$ and $e_1$ denotes the smallest eigenvector direction of Hessian $\nabla^2 f(\tx)$.
Moreover, let the super epoch length $\mathT=\frac{2\log(\frac{8\delta\sqrt{d}}{C_1\rho\zeta' r})}{\eta\delta}=\tdo(\frac{1}{\eta\delta})$, the step size $\eta\leq \min\big(\frac{1}{16\log(\frac{8\delta\sqrt{d}}{C_1\rho\zeta' r})L}, \frac{1}{8C_2L\log \mathT}\big)=\tdo(\frac{1}{L})$, minibatch size $b\geq m$, batch size $B=\tdo(\frac{\sigma^2}{\mathG^2})$ and
the perturbation radius $r\leq \frac{\delta}{C_1 \rho}$, then with probability $1-\zeta$, we have
\begin{align*}
\exists T\leq \mathT,~~ \max\{\n{x_T-x_0}, \n{x_T'-x_0'}\}\geq \frac{\delta}{C_1\rho},
\end{align*}
where $C_1\geq \frac{20C_2}{\eta L}$, $C_2=O(\log\frac{d \mathT}{\zeta})=\tdo(1)$ and $C_2'=O(\log\frac{d \mathT}{\zeta m})=\tdo(1)$.
\end{lemma}
\addtocounter{lemma}{-1}
\endgroup

\begin{proofof}{Lemma~\ref{lem:smallstuckonline}}
We prove this lemma by contradiction. Assume the contrary,
\begin{align}\label{eq:distbound1}
\forall t\leq \mathT~~, \n{x_t-x_0} \leq \frac{\delta}{C_1\rho} \mathrm{~~and~~} \n{x_t'-x_0'} \leq \frac{\delta}{C_1\rho}
\end{align}
We will show that the distance between these two coupled sequences $w_t:=x_t-x_t'$ will grow exponentially since they have a gap in the $e_1$ direction at the beginning, i.e., $w_0:=x_0-x_0'=r_0e_1$, where $r_0=\frac{\zeta' r}{\sqrt{d}}$ and $e_1$ denotes the smallest eigenvector direction of Hessian $\hess := \nabla^2 f(\tx)$.
However, $\n{w_t}=\n{x_t-x_t'}\leq \n{x_t-x_0}+\n{x_0-\tx}+\n{x_t'-x_0'}+\n{x_0'-\tx}\leq 2r+2\frac{\delta}{C_1\rho}$ according to \eqref{eq:distbound1} and the perturbation radius $r$.
It is not hard to see that the exponential increase will break this upper bound, thus we get a contradiction.

In the following, we prove the exponential increase of $w_t$ by induction.
First, we need the expression of $w_t$ (recall that $x_t=x_{t-1}-\eta v_{t-1}$ (see Line \ref{line:update} of Algorithm \ref{alg:ssrgd})):
\begin{align}
w_t&=w_{t-1}-\eta(v_{t-1}-v_{t-1}') \notag\\
&=w_{t-1}-\eta\big(\nabla f(x_{t-1})- \nabla f(x_{t-1}')
        +v_{t-1}-\nabla f(x_{t-1})-v_{t-1}'+\nabla f(x_{t-1}')\big) \notag\\
&=w_{t-1}-\eta\Big(\int_0^1\nabla^2 f(x_{t-1}' + \theta(x_{t-1}-x_{t-1}'))d\theta(x_{t-1}-x_{t-1}')
        +v_{t-1}-\nabla f(x_{t-1})-v_{t-1}'+\nabla f(x_{t-1}')\Big) \notag\\
&=w_{t-1}-\eta\Big((\hess +  \Delta_{t-1})w_{t-1}
        +v_{t-1}-\nabla f(x_{t-1})-v_{t-1}'+\nabla f(x_{t-1}')\Big) \notag\\
&=(I-\eta \hess)w_{t-1}-\eta(\Delta_{t-1}w_{t-1}+y_{t-1}) \notag\\
&=(I-\eta \hess)^{t}w_0-\eta\sum_{\tau=0}^{t-1}(I-\eta \hess)^{t-1-\tau}(\Delta_\tau w_\tau+y_\tau) \label{eq:expw1}
\end{align}
where $\Delta_{\tau} :=\int_0^1(\nabla^2 f(x_\tau'+\theta(x_\tau-x_\tau'))-\hess)d\theta$
and $y_\tau :=v_{\tau}-\nabla f(x_{\tau})-v_{\tau}'+\nabla f(x_{\tau}')$.
Note that the first term of \eqref{eq:expw1} is in the $e_1$ direction and is exponential with respect to $t$, i.e., $(1+\eta\gamma)^t r_0 e_1$, where $-\gamma:=\lambda_{\min}(\hess)=\lambda_{\min}(\nabla^2 f(\tx))\leq -\delta$.
To prove the exponential increase of $w_t$, it is sufficient to show that the first term of \eqref{eq:expw1} will dominate the second term.
We inductively prove the following two bounds
\begin{enumerate}
  \item $\frac{1}{2}(\base)^t r_0\leq\n{w_t}\leq\frac{3}{2}(\base)^t r_0$
  \item $\n{y_t}\leq 2\eta\gamma L(\base)^t r_0$
\end{enumerate}
First, check the base case $t=0$, $\n{w_0}=\n{r_0 e_1}=r_0$ holds for Bound 1. However, for Bound 2, we use Bernstein inequality (Proposition \ref{prop:bernstein}) to show that
$\n{y_0}= \n{v_{0}-\nabla f(x_0)-v_{0}'+\nabla f(x_{0}')}\leq \eta\gamma L r_0$.
According to \eqref{eq:batch}, we know that $v_{0}= \frac{1}{B}\sum_{j\in I_B}\nabla f_j(x_{0})$ and $v_{0}'= \frac{1}{B}\sum_{j\in I_B}\nabla f_j(x_{0}')$ (recall that these two coupled sequence $\{x_t\}$ and $\{x_t'\}$ use the same choice of batches and minibatches (i.e., $I_B$'s and $I_b$'s).
Now, we have
\begin{align}
y_0&=v_{0}-\nabla f(x_0)-v_{0}'+\nabla f(x_{0}') \notag\\
  &=\frac{1}{B}\sum_{j\in I_B} \nabla f_j(x_{0})-\nabla f(x_0)
        -\frac{1}{B}\sum_{j\in I_B}\nabla f_j(x_{0}')+\nabla f(x_{0}')\notag\\
 &=\frac{1}{B}\sum_{j\in I_B}\Big(\nabla f_j(x_{0})-\nabla f_j(x_{0}')
        -(\nabla f(x_0) -\nabla f(x_{0}'))\Big). \label{eq:base}
\end{align}
We first bound each individual term of \eqref{eq:base}:
\begin{align}
\|\nabla f_j(x_{0})-\nabla f_j(x_{0}')
        -(\nabla f(x_0) -\nabla f(x_{0}'))\|\leq 2L\|x_{0}-x_{0}'\|=2L\n{w_0}=2Lr_0,\label{eq:b1x}
\end{align}
where the inequality holds due to the gradient Lipschitz Assumption \ref{asp:1}.
Then, consider the variance term of \eqref{eq:base}:
\begin{align}
  &\sum_{j\in I_B}\E[\ns{\nabla f_j(x_{0})-\nabla f_j(x_{0}')
        -(\nabla f(x_0) -\nabla f(x_{0}'))}] \notag\\
          &\leq \sum_{j\in I_B}\E[\ns{\nabla f_j(x_{0})-\nabla f_j(x_{0}')}] \notag\\
          &\leq BL^2\ns{x_{0}-x_{0}'} \notag\\
          &=BL^2\ns{w_0}=BL^2r_0^2, \label{eq:b2x}
\end{align}
where the first inequality uses the fact $\E[\ns{x-\E x}]\leq \E[\ns{x}]$, and the last inequality uses the gradient Lipschitz Assumption \ref{asp:1}.
According to \eqref{eq:b1x} and \eqref{eq:b2x}, we can bound $y_0$ by Bernstein inequality (Proposition \ref{prop:bernstein}) as
\begin{align*}
\pr\Big\{\big\|y_0\big\|\geq \frac{\alpha}{B} \Big\} &\leq (d+1) \exp\Big(\frac{-\alpha^2/2}{\sigma^2+R\alpha/3}\Big) \notag \\
    & = (d+1) \exp\Big(\frac{-\alpha^2/2}{BL^2r_0^2+ 2Lr_0\alpha/3}\Big)
    \notag \\
    & = \zeta,
\end{align*}
where the last equality holds by letting $\alpha=C_5L\sqrt{B}r_0$, where $C_5=O(\log\frac{d}{\zeta})$. Note that we can further relax the parameter $C_5$ to $C_2'=O(\log\frac{d \mathT}{\zeta m})=\tdo(1)$
for making sure the above arguments hold with probability $1-\zeta$ for all epoch starting points $y_{sm}$ with $sm\leq \mathT$.
Thus, we have with probability $1-\zeta$,
\begin{align}
    \|y_0\| \leq  \frac{C_2'Lr_0}{\sqrt{B}} \leq \eta\gamma L r_0,
\label{eq:basey}
\end{align}
where the last inequality holds
due to $B=\tdo(\frac{\sigma^2}{\mathG^2})$
(recall that $-\gamma:=\lambda_{\min}(\hess)=\lambda_{\min}(\nabla^2 f(\tx))\leq -\delta$ and $\mathG\leq \delta^2/\rho$).

Now, we know that Bound 1 and Bound 2 hold for the base case $t=0$ with high probability. Assume they hold for all $\tau\leq t-1$, we now prove they hold for $t$ one by one.
For Bound 1, it is enough to show the second term of \eqref{eq:expw1} is dominated by half of the first term.
\begin{align}
\n{\eta\sum_{\tau=0}^{t-1}(I-\eta \hess)^{t-1-\tau}(\Delta_\tau w_\tau)}
&\leq \eta\sum_{\tau=0}^{t-1}(\base)^{t-1-\tau}\n{\Delta_\tau}\n{w_\tau} \notag\\
&\leq \frac{3}{2}\eta(\base)^{t-1}r_0\sum_{\tau=0}^{t-1}\n{\Delta_\tau} \label{eq:0x}\\
&\leq \frac{3}{2}\eta(\base)^{t-1}r_0\sum_{\tau=0}^{t-1}\rho D_\tau^x \label{eq:1x}\\
&\leq \frac{3}{2}\eta(\base)^{t-1}r_0t\rho \big(\Dtop\big)\label{eq:2x}\\
&\leq \frac{3}{C_1}\eta\delta t(\base)^{t-1}r_0\label{eq:3x}\\
&\leq \frac{6\log(\frac{8\delta\sqrt{d}}{C_1\rho\zeta' r})}{C_1}(\base)^{t-1}r_0 \label{eq:4x}\\
&\leq \frac{1}{4}(\base)^{t}r_0, \label{eq:5x}
\end{align}
where \eqref{eq:0x} uses the induction for $w_\tau$ with $\tau\leq t-1$,
\eqref{eq:1x} uses the definition $D_\tau^x:=\max\{\n{x_\tau-\tx},\n{x_\tau'-\tx}\}$, \eqref{eq:2x} follows from $\n{x_t-\tx}\leq\n{x_t-x_0}+\n{x_0-\tx}=\Dtop$ due to  \eqref{eq:distbound1} and the perturbation radius $r$,
\eqref{eq:3x} holds by letting the perturbation radius $r\leq \frac{\delta}{C_1\rho}$,
\eqref{eq:4x} holds since $t\leq\mathT=\frac{2\log(\frac{8\delta\sqrt{d}}{C_1\rho\zeta' r})}{\eta\delta}$,
and \eqref{eq:5x} holds by letting $C_1\geq 24\log(\frac{8\delta\sqrt{d}}{\rho\zeta' r})$.

\begin{align}
\n{\eta\sum_{\tau=0}^{t-1}(I-\eta \hess)^{t-1-\tau}y_\tau}
&\leq \eta\sum_{\tau=0}^{t-1}(\base)^{t-1-\tau}\n{y_\tau} \notag\\
&\leq \eta\sum_{\tau=0}^{t-1}(\base)^{t-1-\tau} 2\eta\gamma L(\base)^\tau r_0 \label{eq:10x}\\
&=2\eta \eta\gamma Lt(\base)^{t-1} r_0 \notag\\
&\leq 2\eta \eta\gamma L \frac{2\log(\frac{8\delta\sqrt{d}}{C_1\rho\zeta' r})}{\eta\delta} (\base)^{t-1} r_0 \label{eq:11x}\\
&\leq 4\eta \log(\frac{8\delta\sqrt{d}}{C_1\rho\zeta' r})L(\base)^{t-1} r_0 \label{eq:12x}\\
&\leq \frac{1}{4}(\base)^{t}r_0, \label{eq:13x}
\end{align}
where \eqref{eq:10x} uses the induction for $y_\tau$ with $\tau\leq t-1$,
\eqref{eq:11x} holds since $t\leq\mathT=\frac{2\log(\frac{8\delta\sqrt{d}}{C_1\rho\zeta' r})}{\eta\delta}$,
\eqref{eq:12x} holds $\gamma\geq \delta$ (recall $-\gamma:=\lambda_{\min}(\hess)=\lambda_{\min}(\nabla^2 f(\tx))\leq -\delta$),
and \eqref{eq:13x} holds by letting $\eta\leq \frac{1}{16 \log(\frac{8\delta\sqrt{d}}{C_1\rho\zeta' r})L}$.

Combining \eqref{eq:5x} and \eqref{eq:13x}, we proved the second term of \eqref{eq:expw1} is dominated by half of the first term.
Note that the first term of \eqref{eq:expw1} is $\n{(I-\eta \hess)^{t}w_0}=(1+\eta\gamma)^t r_0$. Thus, we have
\begin{align}\label{eq:wtx}
\frac{1}{2}(\base)^t r_0\leq\n{w_t}\leq\frac{3}{2}(\base)^t r_0
\end{align}

Now, the remaining thing is to prove the second bound $\n{y_t}\leq \eta\gamma L(\base)^t r_0$. First, we write the concrete expression of $y_t$:
\begin{align}
y_t &= v_{t}-\nabla f(x_{t})-v_{t}'+\nabla f(x_{t}') \notag \\
    &= \frac{1}{b}\sum_{i\in I_b}\big(\nabla f_i(x_{t})-\nabla f_i(x_{t-1})\big)
        + v_{t-1}-\nabla f(x_{t})\notag\\
        &\qquad
        -\frac{1}{b}\sum_{i\in I_b}\big(\nabla f_i(x_{t}')-\nabla f_i(x_{t-1}')\big)
        - v_{t-1}'+\nabla f(x_{t}')  \label{eq:30x} \\
    &= \frac{1}{b}\sum_{i\in I_b}\big(\nabla f_i(x_{t})-\nabla f_i(x_{t-1})\big)
        + \nabla f(x_{t-1})-\nabla f(x_{t})  \notag\\
        &\qquad -\frac{1}{b}\sum_{i\in I_b}\big(\nabla f_i(x_{t}')-\nabla f_i(x_{t-1}')\big)
        - \nabla f(x_{t-1}') +\nabla f(x_{t}') \notag\\
        &\qquad
        +v_{t-1}- \nabla f(x_{t-1})
        -v_{t-1}' + \nabla f(x_{t-1}') \notag\\
    &= \frac{1}{b}\sum_{i\in I_b}\big(\nabla f_i(x_{t})-\nabla f_i(x_{t}')
        -\nabla f_i(x_{t-1})+\nabla f_i(x_{t-1}')\big)\notag\\
        &\qquad
        -\big(\nabla f(x_{t})-\nabla f(x_{t}') - \nabla f(x_{t-1}) + \nabla f(x_{t-1}')\big)
        + y_{t-1}, \notag
\end{align}
where \eqref{eq:30x} is due to the definition of the estimator $v_t$ (see Line \ref{line:up2} of Algorithm \ref{alg:ssrgd}).
We further define the difference $z_t:=y_t-y_{t-1}$.
It is not hard to verify that $\{y_t\}$ is a martingale sequence and $\{z_t\}$ is the associated martingale difference sequence.
We will apply the Azuma-Hoeffding inequalities to get an upper bound for $\n{y_t}$
and then we prove $\n{y_t}\leq 2\eta\gamma L(\base)^t r_0$ based on that upper bound.
In order to apply the Azuma-Hoeffding inequalities for martingale sequence $\n{y_t}$, we first need to bound the difference sequence $\{z_t\}$.
We use the Bernstein inequality to bound the differences as follows.
\begin{align}
  z_t&=y_t-y_{t-1} \notag\\
& = \frac{1}{b}\sum_{i\in I_b}\big(\nabla f_i(x_{t})-\nabla f_i(x_{t}')
        -\nabla f_i(x_{t-1})+\nabla f_i(x_{t-1}')\big)\notag\\
        &\qquad
        -\big(\nabla f(x_{t})-\nabla f(x_{t}')
        - \nabla f(x_{t-1}) + \nabla f(x_{t-1}')\big)\notag\\
&=\frac{1}{b}\sum_{i\in I_b}\Big(\big(\nabla f_i(x_{t})-\nabla f_i(x_{t}')\big)
        -\big(\nabla f_i(x_{t-1})-\nabla f_i(x_{t-1}')\big)\notag\\
        &\qquad\qquad\quad
        -\big(\nabla f(x_{t})-\nabla f(x_{t}') \big)
        + \big(\nabla f(x_{t-1}) - \nabla f(x_{t-1}')\big)\Big). \label{eq:zk2x}
\end{align}
We define $u_i:=\big(\nabla f_i(x_{t})-\nabla f_i(x_{t}')\big)
        -\big(\nabla f_i(x_{t-1})-\nabla f_i(x_{t-1}')\big)
        -\big(\nabla f(x_{t})-\nabla f(x_{t}') \big)
        + \big(\nabla f(x_{t-1}) - \nabla f(x_{t-1}')\big)$,
and then we have
\begin{align}
\|u_i\|&=\|\big(\nabla f_i(x_{t})-\nabla f_i(x_{t}')\big)
        -\big(\nabla f_i(x_{t-1})-\nabla f_i(x_{t-1}')\big)
        -\big(\nabla f(x_{t})-\nabla f(x_{t}') \big)
        + \big(\nabla f(x_{t-1}) - \nabla f(x_{t-1}')\big)\| \notag\\
   &\leq \Big\|\int_0^1\nabla^2 f_i(x_{t}'
                      + \theta(x_{t}-x_{t}'))d\theta(x_{t}-x_{t}')
                    -\int_0^1\nabla^2 f_i(x_{t-1}'
                      + \theta(x_{t-1}-x_{t-1}'))d\theta(x_{t-1}-x_{t-1}')\notag\\
        &\qquad
                -\int_0^1\nabla^2 f(x_{t}'
                    +\theta(x_{t}-x_{t}'))d\theta(x_{t}-x_{t}')
                +\int_0^1\nabla^2 f(x_{t-1}'
                    +\theta(x_{t-1}-x_{t-1}'))d\theta(x_{t-1}-x_{t-1}')\Big\| \notag\\
   &=\n{\hess_iw_t+\Delta_{t}^i w_t - (\hess_iw_{t-1}+\Delta_{t-1}^i w_{t-1})
                -(\hess w_t+\Delta_{t} w_t)+(\hess w_{t-1}+\Delta_{t-1} w_{t-1})}    \label{eq:40x} \\
    &\leq \n{(\hess_i-\hess)(w_t-w_{t-1})}
            +\n{(\Delta_{t}^i -\Delta_{t}) w_t-(\Delta_{t-1}^i-\Delta_{t-1}) w_{t-1}} \notag\\
   &\leq 2L\|w_t-w_{t-1}\|+2\rho D_t^x\n{w_t}+2\rho D_{t-1}^x\n{w_{t-1}}, \label{eq:b12x}
\end{align}
where \eqref{eq:40x} holds since we define $\Delta_{t}:=\int_0^1(\nabla^2 f(x_t'+\theta(x_t-x_t'))-\hess)d\theta$ and $\Delta_{t}^i:=\int_0^1(\nabla^2 f_i(x_t'+\theta(x_t-x_t'))-\hess_i)d\theta$,
and the last inequality holds due to the gradient Lipschitz Assumption \ref{asp:1} and Hessian Lipschitz Assumption \ref{asp:2} (recall $D_t^x:=\max\{\n{x_t-\tx},\n{x_t'-\tx}\}$).
Then, consider the variance term
\begin{align}
  &\sum_{i\in I_b}\E[\|u_i\|^2] \notag\\
          &\leq\sum_{i\in I_b}\E[\ns{\big(\nabla f_i(x_{t})-\nabla f_i(x_{t}')\big)
                    -\big(\nabla f_i(x_{t-1})-\nabla f_i(x_{t-1}')\big)}] \notag\\
          &=\sum_{i\in I_b}\E[\ns{\hess_iw_t+\Delta_{t}^i w_t
                                - (\hess_iw_{t-1}+\Delta_{t-1}^i w_{t-1})}] \notag\\
          &\leq b(L\|w_t-w_{t-1}\|+\rho D_t^x\n{w_t}+\rho D_{t-1}^x\n{w_{t-1}})^2, \label{eq:b22x}
\end{align}
where the first inequality uses the fact $\E[\ns{x-\E x}]\leq \E[\ns{x}]$, and the last inequality uses the gradient Lipschitz Assumption \ref{asp:1} and Hessian Lipschitz Assumption \ref{asp:2}.
According to \eqref{eq:b12x} and \eqref{eq:b22x}, we can bound the difference $z_k$ by Bernstein inequality (Proposition \ref{prop:bernstein}) as (where $R=2L\|w_t-w_{t-1}\|+2\rho D_t^x\n{w_t}+2\rho D_{t-1}^x\n{w_{t-1}}$ and $\sigma^2=b(L\|w_t-w_{t-1}\|+\rho D_t^x\n{w_t}+\rho D_{t-1}^x\n{w_{t-1}})^2$)
\begin{align*}
\pr\Big\{\big\|z_t\big\|\geq \frac{\alpha}{b} \Big\} &\leq (d+1) \exp\Big(\frac{-\alpha^2/2}{\sigma^2+R\alpha/3}\Big)= \zeta_k,\\
\end{align*}
where the last equality holds by letting $\alpha=C_4\sqrt{b}(L\|w_t-w_{t-1}\|+\rho D_t^x\n{w_t}+\rho D_{t-1}^x\n{w_{t-1}})$, where $C_4=O(\log\frac{d}{\zeta_k})=\tdo(1)$.

Now, we have a high probability bound for the difference sequence $\{z_k\}$, i.e.,
\begin{align}
    \|z_k\| \leq  c_k= \frac{C_4(L\|w_t-w_{t-1}\|+\rho D_t^x\n{w_t}+\rho D_{t-1}^x\n{w_{t-1}})}{\sqrt{b}} \quad \mathrm{~with~probability~} 1-\zeta_k.
\end{align}

Now, we are ready to get an upper bound for $y_t$ by using the martingale Azuma-Hoeffding inequality.
Note that we only need to focus on the current epoch that contains the iteration $t$ since the martingale sequence $\{y_t\}$ starts with a new point $y_{sm}$ for each epoch $s$ due to the estimator $v_{sm}$. Also note that the starting point $y_{sm}$ can be bounded with the same upper bound \eqref{eq:basey} for all epoch $s$.
Let $s$ denote the current epoch, i.e, iterations from $sm+1$ to current $t$, where $t$ is no larger than $(s+1)m$.
According to Azuma-Hoeffding inequality (Proposition \ref{prop:azumahigh}) and letting $\zeta_k = \zeta/m$, we have
\begin{align*}
\pr\Big\{\big\|y_{t}-y_{sm}\big\|\geq \beta \Big\} &\leq (d+1) \exp\Big(\frac{-\beta^2}{8\sum_{k=sm+1}^{t} c_k^2}\Big)+\zeta \notag \\
    & = 2\zeta,
\end{align*}
where the last equality is due to $\beta=\sqrt{8\sum_{k=sm+1}^{t} c_k^2\log\frac{d}{\zeta}}
=\frac{C_3\sqrt{\sum_{k=sm+1}^{t}(L\|w_t-w_{t-1}\|+\rho D_t^x\n{w_t}+\rho D_{t-1}^x\n{w_{t-1}})^2}}{\sqrt{b}}$, where $C_3=O(C_4\sqrt{\log\frac{d}{\zeta}})=\tdo(1)$.
Recall that $y_k:=v_{k}-\nabla f(x_{k})-v_{k}'+\nabla f(x_{k}')$ and at the beginning point of this epoch $y_{sm}=\n{v_{sm}-\nabla f(x_{sm})-v_{sm}'+\nabla f(x_{sm}')} \leq \eta\gamma L r_0$ with probability $1-\zeta$ (see \eqref{eq:basey}).
Combining with \eqref{eq:basey} and using a union bound, we have

\begin{align}\label{eq:highvar2xon}
\n{y_{t}}\leq \beta +\n{y_{sm}} \leq \frac{C_3\sqrt{\sum_{k=sm+1}^{t}(L\|w_t-w_{t-1}\|+\rho D_t^x\n{w_t}+\rho D_{t-1}^x\n{w_{t-1}})^2}}{\sqrt{b}} +\eta\gamma L r_0
\end{align}
with probability $1-3\zeta$, where $t$ belongs to $[sm+1,(s+1)m]$.
Note that we can further relax the parameter $C_3$ in \eqref{eq:highvar2xon} to $C_2=O(\log\frac{d \mathT}{\zeta})$ (see \eqref{eq:highvar2on})
for making sure the above arguments hold with probability $1-\zeta$ for all $t\leq \mathT$ by using a union bound for $\zeta_t$'s:

\begin{align}\label{eq:highvar2on}
\n{y_{t}}\leq \frac{C_2\sqrt{\sum_{k=sm+1}^{t}(L\|w_t-w_{t-1}\|+\rho D_t^x\n{w_t}+\rho D_{t-1}^x\n{w_{t-1}})^2}}{\sqrt{b}} +\eta\gamma L r_0,
\end{align}
where $t$ belongs to $[sm+1,(s+1)m]$.

\vspace{2mm}
Now, we will show how to bound the right-hand-side of \eqref{eq:highvar2on} to finish the proof, i.e., prove the remaining second bound $\n{y_t}\leq 2\eta\gamma L(\base)^t r_0$.

First, we show that the last two terms in the first term of right-hand-side of \eqref{eq:highvar2on} can be bounded as

\begin{align}
\rho D_t^x\n{w_t}+\rho D_{t-1}^x\n{w_{t-1}} &\leq
\rho\big(\Dtop\big)\frac{3}{2}(\base)^t r_0 + \rho\big(\Dtop\big)\frac{3}{2}(\base)^{t-1} r_0 \notag\\
& \leq 3\rho\big(\Dtop\big)(\base)^t r_0 \notag\\
& \leq \frac{6\delta}{C_1}(\base)^t r_0, \label{eq:50x}
\end{align}
where the first inequality follows from the induction of $\n{w_{t-1}}\leq\frac{3}{2}(\base)^{t-1} r_0$ and the already proved
$\n{w_t}\leq\frac{3}{2}(\base)^t r_0$ in \eqref{eq:wtx},
and the last inequality holds by letting the perturbation radius $r\leq \frac{\delta}{C_1 \rho}$.

Now, we show that the first term in \eqref{eq:highvar2on} can be bounded as

\begin{align}
L\|w_t-w_{t-1}\| &=
L\big\|-\eta \hess(I-\eta \hess)^{t-1}w_0
        -\eta\sum_{\tau=0}^{t-2}\eta \hess(I-\eta \hess)^{t-2-\tau}(\Delta_\tau w_\tau+y_\tau)
        +\eta(\Delta_{t-1} w_{t-1}+y_{t-1}) \big\|
\notag\\
& \leq L\eta \gamma(\base)^{t-1}r_0
        + L\big\|\eta\sum_{\tau=0}^{t-2}
                \eta \hess(I-\eta \hess)^{t-2-\tau}(\Delta_\tau w_\tau+y_\tau) \big\|
        + L\|\eta(\Delta_{t-1} w_{t-1}+y_{t-1})\| \notag\\
& \leq L\eta \gamma(\base)^{t-1}r_0
        + L\eta\big\|\sum_{\tau=0}^{t-2}
                \eta \hess(I-\eta \hess)^{t-2-\tau}\big\|
                \max_{0\leq k\leq t-2}\n{\Delta_k w_k+y_k} \notag\\
        &\qquad
        + L\eta\rho\big(\Dtop\big)\| w_{t-1}\|+L\eta\|y_{t-1}\| \label{eq:60x}\\
& \leq L\eta \gamma(\base)^{t-1}r_0
        + L\eta\sum_{\tau=0}^{t-2}\frac{1}{t-1-\tau}
                \max_{0\leq k\leq t-2}\n{\Delta_k w_k+y_k} \notag\\
        &\qquad
        + L\eta\rho\big(\Dtop\big)\| w_{t-1}\|+L\eta\|y_{t-1}\| \label{eq:61x}\\
& \leq L\eta \gamma(\base)^{t-1}r_0
        + L\eta\log t
                \max_{0\leq k\leq t-2}\n{\Delta_k w_k+y_k} \notag\\
        &\qquad
        + L\eta\rho\big(\Dtop\big)\| w_{t-1}\|+L\eta\|y_{t-1}\| \notag\\
& \leq L\eta \gamma(\base)^{t-1}r_0
        + L\eta\log t
                \max_{0\leq k\leq t-2}\n{\Delta_k w_k+y_k} \notag\\
        &\qquad
        + L\eta\rho\big(\Dtop\big)\frac{3}{2}(\base)^{t-1} r_0
        +2L\eta \eta\gamma L(\base)^{t-1} r_0 \label{eq:62x}\\
& \leq L\eta \gamma(\base)^{t-1}r_0
        + L\eta\log t
                \Big(\rho\big(\Dtop\big)\frac{3}{2}(\base)^{t-2} r_0
                    +2\eta\gamma L(\base)^{t-2} r_0\Big) \notag\\
        &\qquad
        + L\eta\rho\big(\Dtop\big)\frac{3}{2}(\base)^{t-1} r_0
        +2L\eta \eta\gamma L(\base)^{t-1} r_0 \label{eq:63x}\\
& \leq L\eta \gamma(\base)^{t-1}r_0
        + L\eta\log t
                \Big(\frac{3\delta}{C_1}(\base)^{t-2} r_0
                    +2\eta\gamma L(\base)^{t-2} r_0\Big) \notag\\
        &\qquad
        +\frac{3 L\eta\delta}{C_1}(\base)^{t-1} r_0
        +2L\eta \eta\gamma L(\base)^{t-1} r_0 \label{eq:64x}\\
& \leq  \Big(\frac{4}{C_1}\log t +4L\eta\log t\Big)\eta\gamma L(\base)^t r_0, \label{eq:65x}
\end{align}
where the first equality follows from \eqref{eq:expw1},
\eqref{eq:60x} holds from the following \eqref{eq:66x},
\begin{align}\label{eq:66x}
  \n{\Delta_{t}}\leq \rho D_t^x \leq \rho\big(\Dtop\big),
\end{align}
where \eqref{eq:66x} holds due to Hessian Lipschitz Assumption \ref{asp:2}, \eqref{eq:distbound1} and the perturbation radius $r$ (recall that $\Delta_{t}:=\int_0^1(\nabla^2 f(x_t'+\theta(x_t-x_t'))-\hess)d\theta$, $\hess:=\nabla^2 f(\tx)$ and $D_t^x:=\max\{\n{x_t-\tx},\n{x_t'-\tx}\}$),
\eqref{eq:61x} holds due to $\n{\eta \hess(I-\eta \hess)^{t}}\leq \frac{1}{t+1}$,
\eqref{eq:62x} holds by plugging the induction $\n{w_{t-1}}\leq\frac{3}{2}(\base)^{t-1} r_0$
and $\n{y_{t-1}}\leq 2\eta\gamma L(\base)^{t-1} r_0$,
\eqref{eq:63x} follows from \eqref{eq:66x}, the induction
$\n{w_{k}}\leq\frac{3}{2}(\base)^{k} r_0$ and
$\n{y_{k}}\leq 2\eta\gamma L(\base)^{k} r_0$ (hold for all $k\leq t-1$),
\eqref{eq:64x} holds by letting the perturbation radius $r\leq \frac{\delta}{C_1 \rho}$,
and the last inequality holds due to $\gamma\geq \delta$ (recall $-\gamma:=\lambda_{\min}(\hess)=\lambda_{\min}(\nabla^2 f(\tx))\leq -\delta$).

By plugging \eqref{eq:50x} and \eqref{eq:65x} into \eqref{eq:highvar2on}, we have
\begin{align}
\n{y_{t}}
&\leq C_2\left(\frac{6\delta}{C_1}(\base)^t r_0+\Big(\frac{4}{C_1}\log t+4L\eta\log t\Big)\eta\gamma L(\base)^t r_0\right)+\eta\gamma L r_0 \notag\\
&\leq C_2\Big(\frac{6}{C_1\eta L}+\frac{4}{C_1}\log t+4L\eta\log t\Big)\eta\gamma L(\base)^t r_0 +\eta\gamma L r_0 \notag\\
&\leq 2\eta\gamma L(\base)^t r_0, \label{eq:lastx}
\end{align}
where the second inequality holds due to $\gamma\geq \delta$,
and the last inequality holds by letting
$C_1\geq \frac{20C_2}{\eta L}$
and $\eta \leq \frac{1}{8C_2 L\log t}$.
Recall that $C_2=O(\log\frac{d \mathT}{\zeta})$ is enough to let the arguments
in this proof hold with probability $1-\zeta$ for all $t\leq \mathT$.

From \eqref{eq:wtx} and \eqref{eq:lastx}, we know that the two induction bounds hold for $t$.
We recall the first induction bound here:
\begin{enumerate}
  \item $\frac{1}{2}(\base)^t r_0\leq\n{w_t}\leq\frac{3}{2}(\base)^t r_0$
\end{enumerate}
Thus, we know that $\n{w_t} \geq \frac{1}{2}(\base)^t r_0=\frac{1}{2}(\base)^t\frac{\zeta' r}{\sqrt{d}}$. However, $\n{w_t}:=\n{x_t-x_t'}\leq \n{x_t-x_0}+\n{x_0-\tx}+\n{x_t'-x_0'}+\n{x_0'-\tx}\leq 2r+2\frac{\delta}{C_1\rho}\leq \frac{4\delta}{C_1\rho}$ according to \eqref{eq:distbound1} and the perturbation radius $r$.
The last inequality is due to the perturbation radius $r\leq \frac{\delta}{C_1 \rho}$ (we already used this condition in the previous arguments).
This will give a contradiction for \eqref{eq:distbound1} if $\frac{1}{2}(\base)^t\frac{\zeta' r}{\sqrt{d}}\geq \frac{4\delta}{C_1\rho}$ and it will happen if $t\geq \frac{2\log(\frac{8\delta\sqrt{d}}{C_1\rho\zeta' r})}{\eta\delta}$.

So the proof of this lemma is finished by contradiction if we let $\mathT:=\frac{2\log(\frac{8\delta\sqrt{d}}{C_1\rho\zeta' r})}{\eta\delta}$, i.e., we have
\begin{align*}
\exists T\leq \mathT,~~ \max\{\n{x_T-x_0}, \n{x_T'-x_0'}\}\geq \frac{\delta}{C_1\rho}.
\end{align*}
\end{proofof}

\end{document}